%% file: main-submitted.tex
\documentclass[authoryear]{elsarticle}

\usepackage{changepage}
\usepackage{amsmath}
\usepackage{amssymb}
\usepackage{titlesec}
\usepackage{tabularx}
\usepackage{siunitx}
\usepackage{booktabs}
\usepackage{caption}
\usepackage{mathtools}
\usepackage[hyphens]{url}
\usepackage{ragged2e}
\usepackage{pgfplotstable}
\usepackage{longtable}
\usepackage{subcaption}
\usepackage{chngcntr}
\usepackage[margin=3cm]{geometry}
\usepackage{hyperref}
\usepackage{tikz}
\hypersetup{
    colorlinks=true,
    linkcolor=blue,
    filecolor=blue,      
    urlcolor=blue,
    citecolor=blue,
    breaklinks=true
}

\makeatletter
\def\@author#1{\g@addto@macro\elsauthors{\normalsize%
		\def\baselinestretch{1}%
		\upshape\authorsep#1\unskip\textsuperscript{%
			\ifx\@fnmark\@empty\else\unskip\sep\@fnmark\let\sep=,\fi
			\ifx\@corref\@empty\else\unskip\sep\@corref\let\sep=,\fi
		}%
		\def\authorsep{\unskip,\space}%
		\global\let\@fnmark\@empty
		\global\let\@corref\@empty  
		\global\let\sep\@empty}%
	\@eadauthor={#1}
}
\makeatother

\pgfplotsset{compat=1.14}

\titleformat{\paragraph}
{\normalfont\normalsize\itshape}{\theparagraph}{1em}{}
\titlespacing*{\paragraph}
{0pt}{3.25ex plus 1ex minus .2ex}{1.5ex plus .2ex}

\mathchardef\mhyphen="2D

\newcommand\copyrighttext{%
	\footnotesize $\copyright$ 2020. This manuscript version is made available under the CC-BY-NC-ND 4.0 license \url{http://creativecommons.org/licenses/by-nc-nd/4.0/}}
\newcommand\copyrightnotice{%
	\begin{tikzpicture}[remember picture,overlay]
	\node[anchor=south,yshift=10pt] at (current page.south) {\fbox{\parbox{\dimexpr\textwidth-\fboxsep-\fboxrule\relax}{\copyrighttext}}};
	\end{tikzpicture}%
}
\begin{document}
	
\begin{frontmatter}

\title{Recurrent Neural Networks for Time Series Forecasting: Current Status 
and Future Directions}

\author{Hansika Hewamalage\corref{cor1}}
\author{Christoph Bergmeir}
\cortext[cor1]{Corresponding author. Postal Address: Faculty of Information Technology, P.O. Box 63 Monash University, Victoria 3800, Australia. E-mail address: hansika.hewamalage@monash.edu}
\author{Kasun Bandara}
\address{Faculty of Information Technology, Monash University, Melbourne, Australia.}

\begin{abstract}
Recurrent Neural Networks (RNN) have become competitive forecasting methods, as most notably shown in the winning method of the recent M4 competition. 
However, established statistical models such as ETS and ARIMA gain their popularity not only from their high accuracy, but they are also suitable for non-expert users as they are robust, efficient, and automatic. In these areas, RNNs have still a long way to go. We present an extensive empirical study and an open-source software framework of existing RNN architectures for forecasting, that allow us to develop guidelines and best practices for their use.
For example, we conclude that RNNs are capable of modelling seasonality directly if the series in the dataset possess homogeneous seasonal patterns, otherwise we recommend a deseasonalization step.
Comparisons against ETS and ARIMA demonstrate that the implemented (semi-)automatic RNN models are no silver bullets, but they are competitive alternatives in many situations.

\end{abstract}


\end{frontmatter}

\copyrightnotice

\input{sections/introduction}
\input{sections/background}
\input{sections/methodology}
\input{sections/experimental_framework}

\input{sections/results}
\input{sections/conclusion}

\input{sections/future_directions}

\clearpage

\section*{Acknowledgment}

This research was supported by the Australian Research Council under grant DE190100045, Facebook Statistics for Improving Insights and Decisions research award, Monash University Graduate Research funding and MASSIVE - High performance computing facility, Australia.


\bibliographystyle{elsarticle-harv}
\bibliography{references}

\end{document}

%% file: sections/introduction.tex
\section{Introduction}
\label{sec:introduction}

The forecasting field in the past has been characterised by practitioners on the one hand discarding Neural Networks (NN) as not being competitive, and on the other hand NN enthusiasts presenting many complex novel NN architectures, mostly without convincing empirical evaluations against simpler univariate statistical methods. In particular, this notion was supported by many time series forecasting competitions such as the M3, NN3 and NN5 competitions~\citep{Makridakis2018-nt, Crone2011-vq, Crone2008-ye}. Consequently, NNs were labelled as not suitable for forecasting~\citep{hyndman2018blog}. 

There is a number of possible reasons for the underperformance of NNs in the past, one being that individual time series themselves usually are too short to be modeled using complex approaches. Another possibility may be that the time series' characteristics have changed over time so that even long time series may not contain enough relevant data to fit a complex model. Thus, to model sequences by complex approaches, it is essential that they have adequate length as well as that they are generated from a comparatively stable system. Also, NNs are further criticized for their black-box nature~\citep{Makridakis2018-nt}. Thus, forecasting practitioners traditionally have often opted for more straightforward statistical techniques.

However, we are now living in the Big Data era. Companies have gathered plethora of data over the years, which contain important information about their business patterns. Big Data in the context of time series does not necessarily mean that the individual time series contain lots of data. Rather, it typically means that there are many related time series from the same domain. In such a context, univariate forecasting techniques that consider individual time series in isolation, may fail to produce reliable forecasts. They become inappropriate for the Big Data context where a single model could learn simultaneously from many similar time series. On the other hand, more complex models such as NNs benefit most from the availability of massive amounts of data.

Therefore, researchers are now looking successfully into the possibility of applying NNs as substitutes to many other machine learning and statistical techniques.
Most notably, in the recent M4 competition, a Recurrent Neural Network (RNN) was able to achieve impressive performance and win the competition \citep{smyl2020esrnn}. Other examples for successful new developments in the field are novel architectures such as DeepAR, Multi-Quantile Recurrent Neural Network (MQRNN), Spline Quantile Function RNNs and Deep State Space Models for probabilistic forecasting~\citep{Flunkert2017-wp, Wen2017-xz, spline_quantile, deep_state_space_rangapuram}. Though there is an abundance of Machine Learning and Neural Network forecasting methods in the literature, as \cite{Makridakis2018-nt} point out, the methods are typically not evaluated rigorously against statistical benchmarks and would usually perform worse than those. This finding of \cite{Makridakis2018-nt} was arguably one of the main drivers for organising the M4 competition~\citep{Makridakis2018-fc}. Furthermore, often the datasets used or the code implementations for the NN related forecasting research are not made publicly available, which poses an issue for reproducibility of the claimed performance~\citep{Makridakis2018-nt}. A lack of released code implementations also makes it difficult for the forecasting community to adapt such research work to their practical forecasting objectives. 

In contrast, popular statistical models such as ETS and ARIMA that have traditionally supported forecasting in a univariate context gain their popularity not only from their high accuracy. They also have the advantages of being relatively simple, robust, efficient, and automatic, so that they can be used by non-expert users.  
For example, the \texttt{forecast}~\citep{Hyndman2008-gt} package in the \texttt{R} programming language~\citep{r_language} implements a number of statistical techniques related to forecasting such as ARIMA, ETS, Seasonal and Trend Decomposition using Loess (STL Decomposition) in a single cohesive software package. This package still outshines many other forecasting packages later developed, mainly due to its simplicity, accuracy, robustness and ease of use. 

Consequently, many users of traditional univariate techniques will not have the expertise to develop and adapt complex RNN models. They will want to apply competitive, yet easy to use models that can replace the univariate models they currently use in production. Therefore, regardless of the recent successes of RNNs in forecasting, they may still be reluctant to try RNNs as an alternative since they may not have the expert knowledge to use the RNNs adequately and achieve satisfactory accuracy. This is also directly related to the recently emerged dispute in the forecasting community around whether `off-the-shelf' deep learning techniques are able to outperform classical benchmarks. Furthermore, besides the abovementioned intuitions around short isolated series versus large time series databases, no established guidelines exist as to when traditional statistical methods will outperform RNNs, and which particular RNN architecture should be used over another or how their parameters should be tuned to fit a practical forecasting context. 
Although the results from the M4 forecasting competition have clearly shown the potential of RNNs, still it remains unclear how competitive RNNs can be in practice in an automated standard approach, without extensive expert input as in a competition context. 
Therefore, it is evident that the forecasting community would benefit from standard software implementations, as well as guidelines and extensive experimental comparisons of the performance of traditional forecasting methods and the different RNN architectures available. 

Recently, few works have addressed the development of standard software in the area. A package has been developed by Tensorflow in Python, for structural time series modelling using the Tensorflow Probability library~\citep{tensorflow_probability}. This package provides support for generating probabilistic forecasts by modelling a time series as a sum of several structural components such as seasonality, local linear trends, and external variables.  
GluonTS, a Python based open-source forecasting library recently introduced by \citet{gluon_ts} is specifically designed for the purpose of supporting forecasting researchers with easy experimentation using deep neural networks.

Our work also presents a standard software framework, but focused on RNNs and supported by a
review of the literature and implemented techniques, as well as an extensive empirical study. In particular, our study has four main contributions.
First, we offer a systematic overview and categorization of the relevant literature. 
Secondly, we perform a rigorous empirical evaluation of a number of the most popular RNN architectures for forecasting on several publicly available datasets for pure univariate forecasting problems (i.e., without exogenous variables). The implemented models are standard RNN architectures with adequate preprocessing without considering special and sophisticated network types such as in \citet{smyl2020esrnn}'s work involving RNNs. This is mainly because our motivation in this study is to evaluate how competitive are the off-the-shelf RNN techniques for forecasting against traditional univariate techniques and thus provide insights to non-expert forecasting practitioners to start using RNNs. We compare the performance of the involved models against two state-of-the-art statistical forecasting benchmarks, namely the implementations of ETS and ARIMA models from the \texttt{forecast}~\citep{Hyndman2008-gt} package. We stick to single seasonality forecasting problems to be able to compare with those two benchmarks. Although the state of the art in forecasting can now be seen as the methods of the M4 competition winners, most notably \citet{smyl2020esrnn} and \citet{fforma}, we do not present comparisons against those methods as they are not automated in a way that they would be straightforwardly applicable to other datasets, and the idea of our research is to tune RNNs to replace the fully automatic univariate benchmarks that are being used heavily by practitioners for everyday forecasting activities outside of competition environments.
Thirdly, based on the experiments we offer conclusions as a best practices guideline to tune the networks in general. We introduce guidelines at every step of the RNN modelling from the initial preprocessing of the data to tuning the hyperparameters of the models. The methodologies are generic in that they are not tied to any specific domain. Thus, the main objective of our study is to make this work reproducible by other practitioners for their practical forecasting tasks.
Finally, all implementations are publicly available as a cohesive open-source software framework\footnote{Available at: \url{https://github.com/HansikaPH/time-series-forecasting}.}.

The rest of the paper is structured as follows. We first give a comprehensive background study of all related concepts in Section \ref{sec:background_study}, including the traditional univariate forecasting techniques and different NN architectures for forecasting mentioned in the literature. 
Section \ref{sec:methodology} presents the details of the methodology employed, including the RNN architectures implemented and the corresponding data preprocessing techniques.
In Section \ref{sec:experimental_framework}, we explain the experimental framework used for the study with a description of the used datasets, the training, validation and testing specifics and the comparison benchmarks used. In Section \ref{sec:analysis_of_results}, we present a critical analysis of the results followed by the conclusions in Section \ref{sec:conclusion}, and future directions in Section~\ref{sec:future_directions}.

%% file: sections/background.tex
\section{Background Study}
\label{sec:background_study}

This section details the literature related to Univariate Forecasting, Traditional Univariate Forecasting Techniques, Artificial Neural Networks (ANN) and leveraging cross-series information when using ANNs.

\subsection{Univariate Forecasting}
A purely univariate forecasting problem refers to predicting future values of a time series based on its own past values. That is, there is only one time dependent variable. 
Given the target series $X = \{x_1, x_2, x_3, ..., x_t, ..., x_T\}$ 
the problem of univariate forecasting can be formulated as follows:

\begin{equation}
\label{eqn:univariate_forecasting}
\{x_{T+1}, ..., x_{T+H}\} = F(x_1, x_2, ..., x_T) + \epsilon
\end{equation}

Here, $F$ is the function approximated by the model developed for the problem with the variable $X$. The function predicts the values of the series for the future time steps from $T+1$ to $T+H$, where $H$ is the intended forecasting horizon. $\epsilon$ denotes the error associated with the function approximation $F$.

\subsection{Traditional Univariate Forecasting Techniques}
Time series forecasting has been traditionally a research topic in Statistics and Econometrics, from simple methods such as Seasonal Na\"ive and Simple Exponential Smoothing, to more complex ones such as ETS~\citep{hyndman_2008} and ARIMA~\citep{Box1990-km}.
In general, traditional univariate methods were on top in comparison to other computational intelligence methods at many forecasting competitions including NN3, NN5 and M3~\citep{Crone2011-vq, Makridakis2000-kk}. The benefit of the traditional univariate methods is that they work well when the volume of the data is minimal~\citep{Bandara2017-gb}. The number of parameters to be determined in these techniques is quite low compared to other complex machine learning techniques. 
However, such traditional univariate techniques introduced thus far lack few key requirements involved with complex forecasting tasks. Since one model is built per each series, a frequent retraining is required which is compute intensive especially in the case of massive time series databases. 
Also, these univariate techniques are not meant for exploiting cross-series information using global models since they take into account only the features and patterns inherent in a single time series at a time. This is not an issue if the individual time series are long enough having many data points so that the models become capable of capturing the sequential patterns. However in practice this is usually not the case. On the other hand, learning from many time series can be effectively used to address such problems of limited data availability in the single series. 

\subsection{Artificial Neural Networks}
With the ever increasing availability of data, ANNs have become a dominant and popular technique for machine learning tasks in the recent past. A Feed Forward Neural Network (FFNN) is the most basic type of ANN. It has only forward connections in between the neurons contrary to the RNNs which have feedback loops. There are a number of works where ANNs are used for forecasting. \citet{Zhang1998-vz} provide a comprehensive summary of such work in their paper. According to these authors, ANNs possess various appealing attributes which make them good candidates for forecasting against the aforementioned statistical techniques. First, ANNs can model any form of unknown relationship in the data with minimum a-priori assumptions. Second, ANNs can generalize and transfer the learned relationships to unseen data. Third, ANNs are universal approximators, meaning that they are capable of modelling any form of relationship in the data, especially non-linear relationships~\citep{Hornik1989-dd}. The range of functions modelled by an ANN is much higher than the range covered by the statistical techniques.

The studies of \citet{Tang1991-jz} to compare ANNs against the Box-Jenkins methodology for forecasting establish that ANNs are comparatively better for forecasting problems with long forecasting horizons. \citet{Claveria2014-qv} derive the same observations with respect to a tourism demand forecasting problem. However, determining the best network structure and the training procedure for a given problem are crucial to tune ANNs for maximal accuracy. An equally critical decision is the selection of the input variables for the modelling~\citep{Zhang2007-jk}. ANNs have been applied for forecasting in a multitude of domains over the years. \citet{Mandal2006-gq} use an ANN for electric load forecasting along with Euclidean Norm applied on weather information of both the training data as well as the forecasting duration to derive similarity between data points in the training data and the forecasts. This technique is called the similar days approach. Efforts have also been made to minimize the manual intervention in the NN modelling process to make it automated. To this end, \citet{Yan2012-ar} propose a new form of ANN named as the Generalized Regression Neural Network (GRNN) which is a special form of a Radial Basis Function (RBF) Network. It requires the estimation of just one design parameter, the Spread Factor which decides the width of the RBF and consequently how much of the training samples contribute to the output. A hybrid approach to forecasting is proposed by \citet{Zhang2003-de} combining an ARIMA model with an ANN and thus leveraging the strengths of both models. The ARIMA model is used to model the linear component of the time series while the ANN can then model the residual which corresponds to the non-linear part. This approach is closely related to boosting, commonly used as an ensembling technique to reduce bias in predictions. The idea has been inspired from the common concept that a combination of several models can often outperform the individual models in isolation. \citet{Zhang2001-ak} and \citet{Rahman2016-sf} have also worked using ANNs for forecasting, specifically with ensembles. 

The most common way to feed time series data into an ANN, specifically an FFNN is to break the whole sequence into consecutive input windows and then get the FFNN to predict the window or the single data point immediately following the input window. Yet, FFNNs ignore the temporal order within the input windows and every new input is considered in isolation~\citep{Bianchi2017-er}. No state is carried forward from the previous inputs to the future time steps. This is where RNNs come into play; a specialized NN developed for modelling data with a time dimension. 

\subsubsection{Recurrent Neural Networks for Forecasting}
\label{sec:rnn_architectures_review}

RNNs are the most commonly used NN architecture for sequence prediction problems. They have particularly gained popularity in the domain of natural language processing. Similar to ANNs, RNNs are universal approximators~\citep{Schafer2006-ln} as well. However, unlike ANNs, the feedback loops of the recurrent cells inherently address the temporal order as well as the temporal dependencies of the sequences~\citep{Schafer2006-ln}.

Every RNN is a combination of a number of RNN units. The most popular RNN units commonly used for sequence modelling tasks are the Elman RNN cell, Long Short-Term Memory (LSTM) cell and the Gated Recurrent Unit (GRU)~\citep{Elman1990-by, Hochreiter1997-vh, Cho2014-vi}. Apart from them, other variants have been introduced such as Depth Gated LSTM, Clockwork RNN, Stochastic Recurrent Networks and Bidirectional RNN~\citep{Yao2015-re, Koutnik2014-sp, Bayer2014-pr, Schuster1997-zq}. Nevertheless, these latter RNN units have hardly been used in the forecasting literature. They were mostly designed with language modelling tasks in mind.

\citet{Jozefowicz2015-fz} perform an empirical evaluation of different RNN cells. In particular, those authors aim to find an RNN unit that performs better than the LSTM and GRU in three defined tasks namely, arithmetic computations, XML modelling where the network has to predict the next character in a sequence of XML data and a language modelling task using the Penn TreeBank dataset. They do not cover time series forecasting in particular. 
The work by \citet{Bianchi2017-er} specifically targets the problem of short-term load forecasting. Their experiments systematically test the performance of all three of the popular recurrent cells, Elman RNN (ERNN) cell, LSTM cell and the GRU cell, and compare those with Echo State Networks (ESN) and the Non-linear Autoregressive with eXogenous (NARX) inputs Network. The experiments are performed on both synthetic time series as well as real-world time series. From the results, the authors conclude that both LSTM and GRU demonstrate similar performance in the chosen datasets. In essence, it is hard to differentiate which one is better in which scenario. Additionally, the ERNN shows a comparable performance to the gated RNN units, and it is faster to train. The authors argue that gated RNN units can potentially outperform ERNNs in language modelling tasks where the temporal dependencies can be highly non-linear and abrupt. Furthermore, the authors state that gradient-based RNN units (ERNN, LSTM, GRU) are relatively slow in terms of training time due to the time-consuming backpropagation through time procedure.

A recurrent unit can constitute an RNN in various types of architectures. A number of different RNN architectures for forecasting can be found in the literature. Although most commonly used for natural language processing tasks, these architectures are used in different time series forecasting tasks as well. The stacked architecture can be claimed as the most commonly used architecture for forecasting with RNNs. \citet{Bandara2017-gb} employ the Stacked model in their work on using a clustering approach for grouping related time series in forecasting. Due to the vanishing gradient problem existent in the vanilla RNN cells, LSTM cells (with peephole connections) are used instead. The method includes several essential data preprocessing steps and a clustering phase of the related time series to exploit cross-series information, and those authors demonstrate significant results on both the CIF 2016 as well as the NN5 forecasting competition datasets. In fact, it is the same architecture used by \citet{Smyl_undated-zz} to win the CIF 2016 forecasting competition. In the competition, Smyl has developed two global models, one for the time series with forecasting horizon 12 and the other one for the time series with horizon 6. The work by \citet{Smyl2016-rf} closely follows the aforementioned stacked architecture with a slight modification known as the skip connections. Skip connections allow layers far below in the stack to directly pass information to layers well above and thus minimize the vanishing gradient effect. With the skip connections added, the architecture is denoted as the ResNet architecture which is adapted from the work of \citet{He2015-yn} originally for image recognition.   

Another RNN architecture popular in neural machine translation tasks is the Sequence to Sequence (S2S) architecture introduced by \citet{Sutskever2014-hx}. The overall model has two components, the encoder and the decoder which both act as two RNN networks on their own. \citet{Peng2018-ea} apply a S2S architecture for host load prediction using GRU cells as the RNN unit in the network. Those authors test their approach using two datasets, the Google clusters dataset and the Dinda dataset from traditional Unix systems. The approach is compared against two other state-of-the-art RNN models, an LSTM-based network and the ESN. According to the results, the GRU based S2S model manages to outperform the other models in both the datasets. The DeepAR model for probabilistic forecasting, developed by \citet{Flunkert2017-wp} for Amazon also uses a S2S architecture for prediction. The authors of that work use the same architecture for both the encoder and the decoder components, although in practice the two components usually differ. Therefore, the weights of both the encoder and the decoder are the same. \citet{Wen2017-xz} develop a probabilistic forecasting model using a slightly modified version of a S2S network. As teacher-signal enforcing of the decoder using autoregressive connections generally leads to error accumulation throughout the prediction horizon, those authors use a combination of two Multi Layer Perceptrons (MLP) for the decoder; a global MLP which encapsulates the encoder outputs and the inputs of the future time steps and a local MLP which acts upon each specific time step of the prediction horizon to generate the quantiles for that point. Since the decoder does not use any autoregressive connections, the technique is called the Direct Multi Horizon Strategy.

More recently, S2S models aka autoencoders are used in forecasting to extract time series features followed by another step to generate the actual predictions. 
\citet{Zhu2017-yb} use an autoencoder to address the issue of uncertainty, specifically in the form of model misspecification. The models that are fitted using the training data may not be the optimal models for the test data, when the training data patterns are different from the test data. The autoencoder can alleviate this by extracting the time series features during training and calculating the difference between these features and the features of the series encountered during testing. This gives the model the intuition of how different the training and test data partitions are. 
\citet{Laptev2017-xa} incorporate an autoencoder to train a global model across many heterogeneous time series. The autoencoder in this context is expected to extract the features of each series which are later concatenated to the input window for forecasting by an LSTM model. Likewise, apart from direct prediction, the S2S model is also used in intermediate feature extraction steps prior to the actual forecasting.

\citet{Bahdanau2014-bj} and \citet{Luong2015-lp} present two different variants of the S2S model, with attention mechanisms. The fundamental idea behind these attention mechanisms is to overcome the problem in S2S models that they encode all the information in the whole time series to just a single vector and then decode this vector to generate outputs. Embedding all the information in a fixed-size vector can result in information loss~\citep{Qin2017-lt}. Hence, the attention model tries to overcome this by identifying important points in the time series to pay attention to. All the points in the time series are assigned weights which are computed for each output time step and the more important points are assigned higher weights than the less important ones. This method has been empirically proven to be heavily successful in various neural machine translation tasks where the translation of each word from one language to another requires specific attention on particular words in the source language sentence. 

Since time series possess seasonality components, attention weights as described above can be used in a forecasting context. For instance, if a particular monthly time series has a yearly seasonality, predicting the value for the next immediate month benefits more from the value of the exact same month of the previous year. This is analogous to assigning more weight to the value of the sequence exactly 12 months ago. The work of \citet{Suilin_undated-hh} done for the Kaggle challenge of Wikipedia Web Traffic Forecasting encompasses this idea~\citep{kaggle}. However, that author's work does not use the traditional \citet{Bahdanau2014-bj} or \citet{Luong2015-lp} attention since those techniques require re-computation of the attention weights for the whole sequence at each time step, which is computationally expensive. There are two variants in the attention mechanism proposed by \citet{Suilin_undated-hh}. In the first method, the encoder outputs that correspond to important points identified before are directly fed as inputs to the respective time steps in the decoder. In the second method, the important points concept is relaxed to take a weighted average of those important points along with their two neighbouring points. This scheme helps to account for the noise in the time series as well as to cater for the different lengths of the months and the leap years. This latter attention has been further extended to form a type of hierarchical attention using an interconnected hierarchy of 1D convolution layers corresponding to weighted averaging the neighbouring points and max pooling layers in between. Despite using the same weights for all the forecasting steps at the decoder, \citet{Suilin_undated-hh} claims that this reduces the error significantly. 

\citet{Cinar2017-nt} use a more complex attention scheme to treat the periods in the time series. The idea of those authors is that the classical attention mechanism proposed by \citet{Bahdanau2014-bj} is intended for machine translation and therefore it does not attend to the seasonal periods in time series in a forecasting condition. Hence, they propose an extension to the Bahdanau-style attention and name it as Position-based Content Attention Mechanism. An extra term ($\pi^{(1)}$) is used in the attention equations to embed the importance of each time step in the time series to calculate the outputs for the forecast horizon. Based on whether a particular time step in the history corresponds to a pseudo-period of the output step or not, the modified equation can be used to either augment or diminish the effect of the hidden state. The vector $\pi^{(1)}$ is expected to be trained along with the other parameters of the network. Those authors perform experiments using their proposed technique over six datasets, both univariate and multivariate. They further compare the resulting error with the traditional attention mechanism as well as ARIMA and Random Forest. The results imply that the proposed variant of the attention mechanism is promising since it is able to surpass the other models in terms of accuracy for five of the tested six datasets. Furthermore, the plots of the attention weights of the classical attention scheme and the proposed variant denote that by introducing this variant, the pseudo-periods get assigned more weight whereas in the na\"ive attention, the weights are increasing towards the time steps closer to the forecast origin. 

Another work by \citet{Qin2017-lt} proposes a dual-stage attention mechanism (DA-RNN) for multivariate forecasting problems. In the first stage of the attention which is the input attention, different weights are assigned to different driving series based on their significance to contribute towards forecasting at each time step. This is done prior to feeding input to the encoder component of the S2S network. The input received by each time step of the encoder is a set of values from different exogenous driving series whose influence is sufficiently increased or decreased. The weights pertaining to each driving series' values are derived using another MLP which is trained along with the S2S network. In the second stage of attention which is the temporal attention, a usual Bahdanau-style attention scheme is employed to assign weights to the encoder outputs at each prediction step. In terms of the experiments, the model is tested using two datasets. An extensive comparison is performed among many models including ARIMA, NARX RNN, Encoder Decoder, Attention RNN, Input Attention RNN and the Dual Stage Attention RNN. The Dual Stage Attention model is able to outperform all the other models in both the datasets. Further experiments using noisy time series reveal that the Dual Stage Attention Model is comparatively robust to noise.

More recently, \citet{Liang2018-bd} have developed a multi-level attention network named GeoMAN for time series forecasting of Geo-sensory data. According to those authors, the significance of that model in comparison to the DA-RNN model is that this model explicitly handles the special characteristics inherent in Geo-sensory data such as the spatio-temporal correlation. While the DA-RNN has a single input attention mechanism to differentiate between different driving series, the GeoMAN has two levels of spatial attention; the local spatial attention which highlights different local time series produced by the same sensor based on their importance to the target series and the global spatial attention which does the same for the time series produced by different surrounding sensors globally. The concatenation of the context vectors produced by these two attention mechanisms is then fed to a third temporal attention mechanism used on the decoder. Those authors emphasize the importance of the two-fold spatial attention mechanism as opposed to one input attention as in the work by \citet{Qin2017-lt} since the latter treats both local and global time series as equal in its attention technique. The underlying attention scheme used for all three attention steps follows a Bahdanau-style additive scheme. Empirical evidence on two Geo-sensory datasets have proven the proposed GeoMAN model to outperform other state-of-the-art techniques such as S2S models, ARIMA, LSTM, and DA-RNN. 

Apart from using individual RNN models, several researchers have also experimented using ensembles of RNN models for forecasting. Ensembles are meant for combining multiple weak learners together to generate a more robust prediction and thus reduce the individual training time of each base RNN. \citet{Smyl2017-ensemble} segments the problem into two parts; developing a group of specialized RNN models and ensembling them to derive a combined prediction. That author's methodology follows the idea that general clustering based on a standard metric for developing ensembles does not offer the best performance in the context of forecasting. Therefore this approach randomly assigns time series of the dataset to one of the RNNs in the pool for training at the first epoch. Once all networks are trained, every time series in the dataset is assigned to a specific $N$ number of best networks which give the minimum error after the training in that epoch. RNNs are trained in this manner iteratively using the newly allocated series until the validation error grows or the number of epochs finishes. The ensembling includes using another network to determine which RNN in the pool should make the forecasts for a given series or feeding the forecasts of each network to another network to make the final predictions or using another network to determine the weights to be assigned to the forecasts of each RNN in the pool. Despite the complexity associated with these techniques, that author claims that none of them work well. Simple techniques such as average, weighted average (weight based on training loss of each network or the rate of becoming a best network) of all the networks or the weighted average of the $N$ best networks work well on the other hand. This methodology demonstrates encouraging results on the monthly series of the M3 competition dataset. The RNNs used for the pool comprise of a stacked LSTM with skip connections. This technique is also used in the winning solution of \citet{smyl2020esrnn} at the M4 competition which proves that this approach works well. 

\citet{Krstanovic2017-xb} suggest that the higher the diversity of the base learners, the better the accuracy of the final ensemble. Those authors develop an ensemble using the method known as Stacking where a number of base LSTM learners are combined using a meta-learner whose inputs are the outputs of the base learners. The outputs of the meta-learner correspond to the final forecasts expected. The dissimilarity of the base learners is enforced by using a range of values for the hyperparameters such as dropout rate, number of hidden layers, number of nodes for the layers and the learning rate. For the meta-learner those authors use Ridge Regression, eXtreme Gradient Boosting (XGBoost) and Random Forest. The method is compared with other popular techniques such as LSTM, Ensemble via mean forecast, Moving Average, ARIMA and XGBoost on four different datasets. The suggested ensemble with Stacking in general performs best.  

A modified boosting algorithm for RNNs in the time series forecasting context is proposed by \citet{Assaad2008-kt} based on the AdaBoost algorithm. The novelty aspect of this approach is that the whole training set is considered for model training at each iteration by using a specialized parameter k which controls the weight exerted on each training series based on its error in the previous iteration. Therefore, in every iteration, the series that got a high error in the previous iteration are assigned more weight. A value of 0 for k, assigns equal weights for all the time series in the training set. Furthermore, the base models in this approach are merged together using a weighted median which is more robust when encountered with outliers as opposed to weighted mean. Experiments using two time series datasets show promising performance of this ensemble model on both single-step ahead and multi-step ahead forecasting problems, compared to other models in the literature such as MLP or Threshold Autoregressive (TAR) Model.

\subsection{Leveraging Cross-Series Information}
\label{sec:cross_series_lit_review}

Exploiting cross-series information in forecasting is an idea that gets increased attention lately, especially in the aftermath of the M4 competition. The idea is that instead of developing one model per each time series in the dataset, a model is developed by exploiting information from many time series simultaneously. In the literature such models are often referred to as global models whereas univariate models which build a model per every series are known as local models~\citep{Januschowski2020ijf}. However, the application of global models to a set of time series does not indicate any interdependence between them with respect to the forecasts. Rather, it means that the parameters are estimated globally for all the time series available~\citep{Januschowski2020ijf}. The former is basically the scenario of multivariate forecasting where the value of one time series is driven by other external time varying variables. Such relationships are directly modelled in the forecast equations of the method. Yet, a global model trained across series usually works independently on the individual series in a univariate manner when producing the forecasts. 

In modern forecasting problems, often the requirement is to produce forecasts for many time series which may have similar patterns, as opposed to forecasting just one time series. One common example in the domain of retail is to produce forecasts for many similar products. In such scenarios, global models can demonstrate their true potential by learning across series to incorporate more information, in comparison to the local methods. \citet{pooled_regression} use a similar idea in their work for demand forecasting of stock-keeping units (SKUs). In particular, for SKUs with limited or no promotional history associated with it, the coefficients of the regression model are calculated by pooling across many SKUs. However, the regression model used can capture only linear relationships and it does not maintain any internal state per each time series.

In recent literature, researchers have used the idea of developing global models in the context of deep neural networks. For RNNs, this means that the weights are calculated globally, yet the state is maintained per each time series. The winning solution by \citet{smyl2020esrnn} at the M4 forecasting competition uses the global model concept with local parameters as well, to cater for individual requirements of different series. \citet{Bandara2017-gb} do this by clustering groups of related time series. A global model is developed per each cluster. \citet{Flunkert2017-wp} also apply the idea of cross-series information in their DeepAR model for probabilistic forecasting. More recently, \citet{Wen2017-xz, deep_state_space_rangapuram, deep_factors_wang} and \citet{n_beats} also employ the cross series learning concept in their work using deep neural networks for forecasting. \citet{bandara_ecommerce} develop global models for forecasting in an E-commerce environment by considering sales demand patterns of similar products. 

%% file: sections/methodology.tex
\section{Methodology}
\label{sec:methodology}

In this section, we describe the details of the methodology employed in our comprehensive experimental study. We present the different recurrent unit types, the RNN architectures as well as the learning algorithms that we implement and compare in our work. 

\subsection{Recurrent Neural Networks}

Our work implements a number of RNN architectures along with different RNN units. These are explained in detail next.

\subsubsection{Recurrent Units}
\label{sec:recurrent_units_review}

Out of the different RNN units mentioned in the literature, we select the following three types of recurrent units to constitute the layers of the RNNs in our experiments.

\begin{itemize}
	\item Elman Recurrent Unit
	\item Gated Recurrent Unit
	\item Long Short-Term Memory with Peephole Connections
\end{itemize}
The base recurrent unit is introduced by \citet{Elman1990-by}. The structure of the basic ERNN cell is as shown in Figure \ref{fig:elman_rnn}. 

\begin{center}
	\captionsetup{type=figure}
	\includegraphics[scale=0.5]{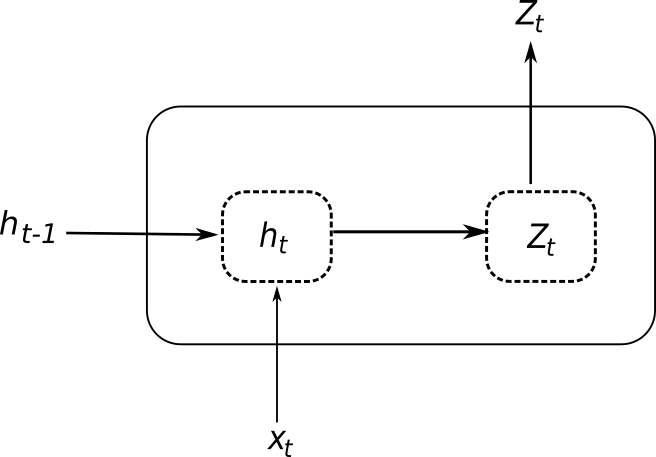}
	\captionof{figure}{Elman Recurrent Unit}
	\label{fig:elman_rnn}
\end{center}

\begin{subequations}
	\begin{gather}
	h_{t} = \sigma(W_{i}\cdot h_{t - 1} + V_{i}\cdot x_{t} + b_{i}) \label{eqn:basic_rnn_state}\\
	z_{t} = tanh(W_{o}\cdot h_{t} + b_{o}) \label{eqn:basic_rnn_output}
	\end{gather}
\end{subequations}

In Equations \ref{eqn:basic_rnn_state} and \ref{eqn:basic_rnn_output}, $h_{t} \in \mathbb{R}^{d}$ denotes the hidden state of the RNN cell (d being the cell dimension). This is the only form of memory in the ERNN cell. $x_{t} \in \mathbb{R}^{m}$ (m being the size of the input) and $z_{t} \in \mathbb{R}^{d}$ denote the input and output of the cell at time step $t$. $W_{i} \in \mathbb{R}^{d \times d}$ and $V_{i} \in \mathbb{R}^{d \times d}$ denote the weight matrices whereas $b_{i} \in \mathbb{R}^{d}$ denotes the bias vector for the hidden state. Likewise, $W_{o} \in \mathbb{R}^{d \times d}$ and $b_{o} \in \mathbb{R}^{d}$ signify the weight matrix and the bias vector of the cell output. The current hidden state depends on the hidden state of the previous time step as well as the current input. This is supported with the feedback loops in the RNN cell connecting its current state to the next state. These connections are of extreme importance to consider past information in updating the current cell state. In the experiments, we use the sigmoid function (indicated by $\sigma$) as the activation of the hidden state and the hyperbolic tangent function (indicated by tanh) as the activation of the output.

The ERNN Cell suffers from the well known vanishing gradient and exploding gradient problems over very long sequences. This implies that the simple RNN cells are not capable of carrying long term dependencies to the future. When the sequences are quite long, the backpropagated gradients tend to diminish (vanish) and consequently the weights do not get updated adequately. On the other hand, when the gradients are huge, they may burst (explode) over long sequences resulting in unstable weight matrices. Both these issues are ensued from the gradients being intractable and hinder the ability of the RNN cells to capture long term dependencies.

Over the years, several other variations have been introduced to this base recurrent unit addressing its shortcomings. The LSTM cell introduced by \citet{Hochreiter1997-vh} is perhaps the most popular cell for natural language processing tasks due to its capability to capture long-term dependencies in the sequence while alleviating gradient vanishing issues. The structure of the LSTM Cell is illustrated in Figure \ref{fig:lstm_cell} 

\begin{center}
	\captionsetup{type=figure}
	\includegraphics[scale=0.3]{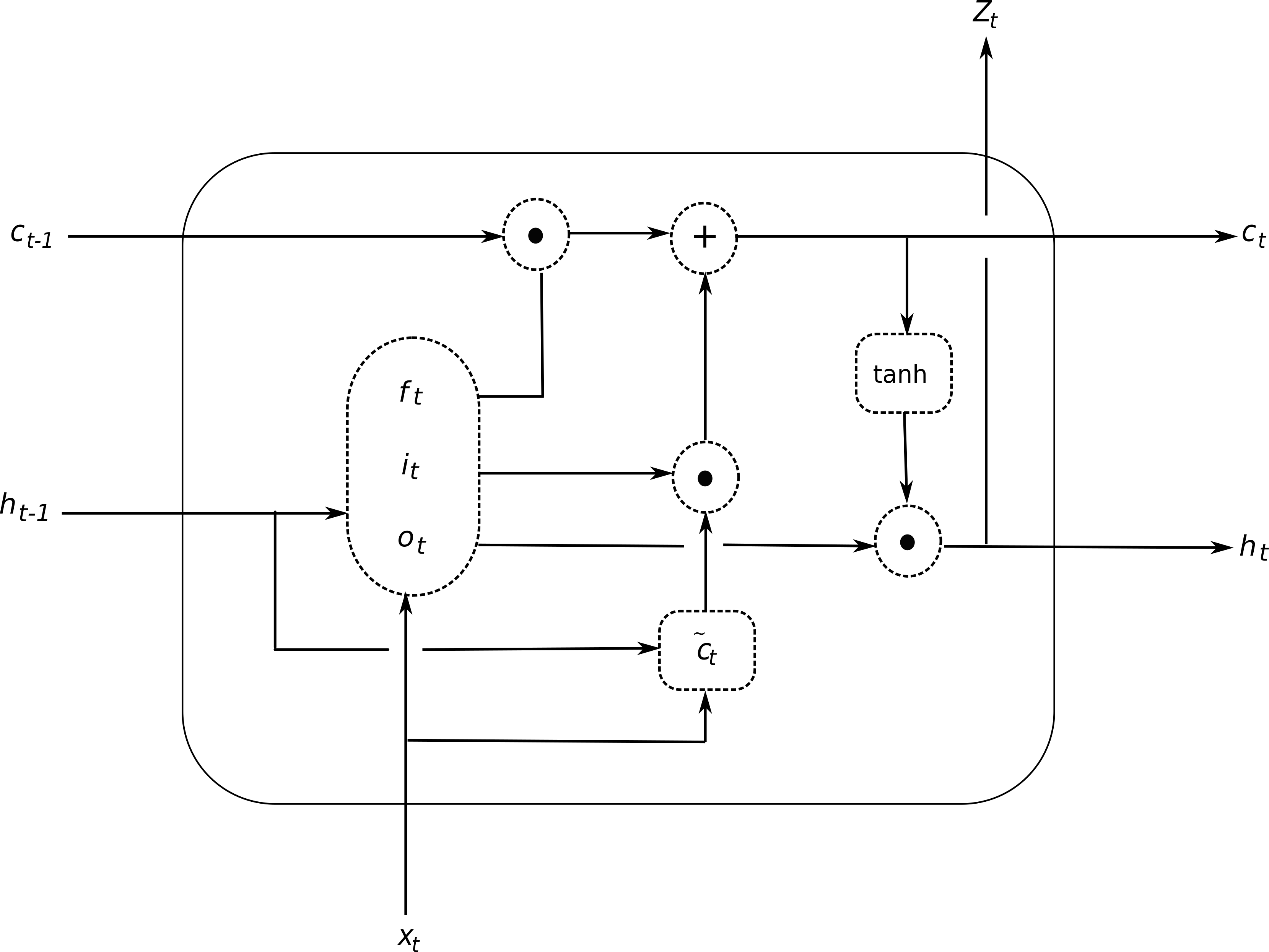}
	\captionof{figure}{Basic Long Short-Term Memory Unit}
	\label{fig:lstm_cell}
\end{center}

\begin{subequations}
	\begin{gather}
	i_{t} = \sigma(W_{i} \cdot h_{t - 1} + V_{i} \cdot x_{t} + b_{i}) \label{eqn:lstm_input_gate}\\
	o_{t} = \sigma(W_{o} \cdot h_{t - 1} + V_{o} \cdot x_{t} + b_{o})\\
	f_{t} = \sigma(W_{f} \cdot h_{t - 1} + V_{f} \cdot x_{t} + b_{f})\\
	\tilde{C_{t}} = tanh(W_{c} \cdot h_{t - 1} + V_{c} \cdot x_{t} + b_{c})\\
	C_{t} = i_{t}\odot \tilde{C_{t}} + f_{t} \odot C_{t-1} \label{eqn:cell_state}\\
	h_{t} = o_{t}\odot tanh(C_{t})\\
	z_{t} = h_{t} \label{eqn:lstm_output}
	\end{gather}
\end{subequations}

Compared to the basic RNN cell, the LSTM cell has two components to its state, the hidden state and the internal cell state where the hidden state corresponds to the short-term memory component and the cell state corresponds to the long-term memory. With its Constant Error Carrousel (CEC) capability supported by the internal state of the cells, LSTM avoids the vanishing and exploding gradient issues. Moreover a gating mechanism is introduced which comprises of the input, forget and the output gates. In the Equations \ref{eqn:lstm_input_gate} - \ref{eqn:lstm_output}, $h_{t} \in \mathbb{R}^d$ is a vector which denotes the hidden state of the cell, where d is the cell dimension. Similarly $C_{t} \in \mathbb{R}^d$ is the cell state and  $\tilde{C_{t}} \in \mathbb{R}^d$ is the candidate cell state at time step t which captures the important information to be persisted through to the future. $x_{t} \in  \mathbb{R}^d$ and $z_{t} \in \mathbb{R}^d$ are the same as explained for the basic RNN cell. $W_{i}, W_{o}, W_{f}, W_{c} \in  \mathbb{R}^{d \times d}$ denote the weight matrices of the input gate, output gate, forget gate and the cell state respectively. Likewise,$V_{i}, V_{o}, V_{f}, V_{c} \in  \mathbb{R}^{d \times d}$ and $b_{i}, b_{o}, b_{f}, b_{c} \in  \mathbb{R}^d$ denote the weight matrices corresponding to the current input and the bias vectors respectively. $i_{t}, o_{t}, f_{t} \in  \mathbb{R}^d$ are the input, output and forget gate vectors. 

The activation function $\sigma$ of the gates denotes the sigmoid function which outputs values in the range [0, 1]. In Equation \ref{eqn:cell_state}, the input and the forget gates together determine how much of the past information to retain in the current cell state and how much of the current context to propagate forward to the future time steps. $\odot$ denotes the element wise multiplication which is known as the Hudmard Product. A value of 0 in forget gate $f_{t}$ denotes that nothing should be carried forward from the previous cell state. In other words, the previous cell state should be completely forgotten in the current cell state. Following this argument, a value of 1 implies that the previous cell state should be completely retained. The same notion holds for the other two gates $i_{t}$ and $o_{t}$. A value in between the two extremes of 0 and 1 for both the input and forget gates can carefully control the value of the current cell state using only the important information from both the previous cell state and the current candidate cell state. For the candidate cell state, the activation function is a hyperbolic tangent function which outputs values in the range [-1, 1]. An important difference of the LSTM cell compared to the simple RNN cell is that its output $z_{t}$ is equal to the hidden state $h_{t}$. 

In this review, we use the variant of the vanilla LSTM cell known as the LSTM cell with peephole connections. In this LSTM unit, the states are updated as per the following equations.

\begin{subequations}
	\begin{gather}
	i_{t} = \sigma(W_{i} \cdot h_{t - 1} + V_{i} \cdot x_{t} + P_{i} \cdot C_{t - 1} + b_{i})\label{eqn:peephole_input_gate}\\
	o_{t} = \sigma(W_{o} \cdot h_{t - 1} + V_{o} \cdot x_{t} + P_{o} \cdot C_{t} + b_{o})\\
	f_{t} = \sigma(W_{f} \cdot h_{t - 1} + V_{f} \cdot x_{t} + P_{f} \cdot C_{t - 1} +  b_{f})\\
	\tilde{C_{t}} = tanh(W_{c} \cdot h_{t - 1} + V_{c} \cdot x_{t} + b_{c})\\
	C_{t} = i_{t}\odot \tilde{C_{t}} + f_{t} \odot C_{t-1}\\
	h_{t} = o_{t}\odot tanh(C_{t})\\
	z_{t} = h_{t} \label{eqn:peephole_output}
	\end{gather}
\end{subequations}

The difference is that the peephole connections let the forget and the input gates of the cell look at the previous cell state $C_{t - 1}$ before updating it. In the Equations  \ref{eqn:peephole_input_gate} to \ref{eqn:peephole_output}, $P_{i}, P_{o}, P_{f} \in \mathbb{R}^{d \times d}$ represent the weight matrices of the input, output, and forget gates respectively. As for the output gate, it can now inspect the current cell state for generating the output. The implementation used in this research is the default implementation of peephole connections in the Tensorflow framework.

The GRU is another variant introduced by \citet{Cho2014-vi} which is comparatively simpler than the LSTM unit as well as faster in computations. This is due to the LSTM unit having three gates within the internal gating mechanism, whereas the GRU has only two, the update gate and the reset gate. The update gate in this unit plays the role of the forget gate and the input gate combined. Furthermore, similar to the vanilla RNN cell, the GRU cell also has only one component to the state, i.e the hidden state. Figure \ref{fig:gru_cell} and the equations display the functionality of the GRU cell.

\begin{center}
	\captionsetup{type=figure}
	\includegraphics[scale=0.33]{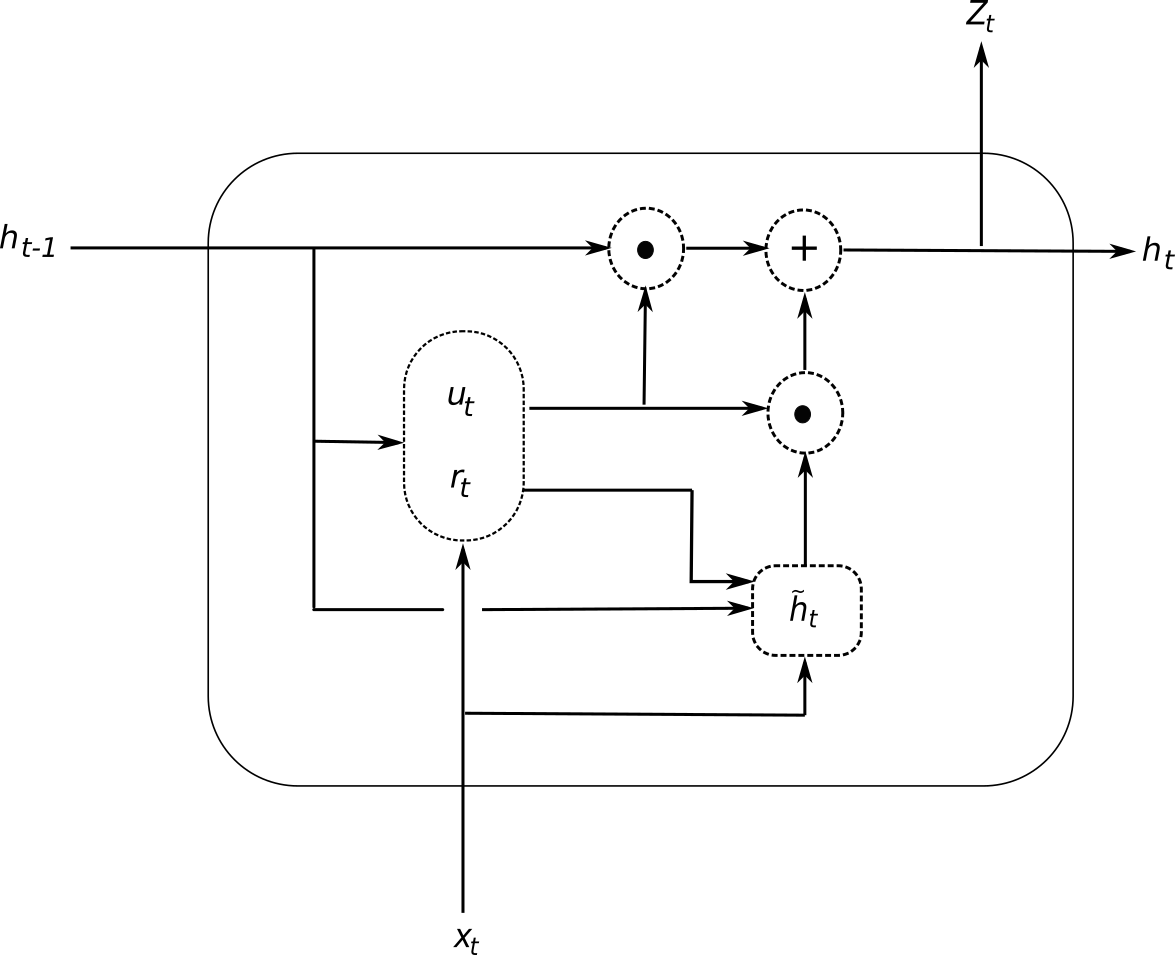}
	\captionof{figure}{Gated Recurrent Unit}
	\label{fig:gru_cell}
\end{center}

\begin{subequations}
	\begin{gather}
	u_{t} = \sigma(W_{u} \cdot h_{t - 1} + V_{u} \cdot x_{t} + b_{u})\\
	r_{t} = \sigma(W_{r} \cdot h_{t - 1} + V_{r} \cdot x_{t} + b_{r})\\
	\tilde{h_{t}} = tanh(W_{h} \cdot r_{t} \cdot h_{t - 1} + V_{h} \cdot x_{t} + b_{h})\\
	h_{t} = u_{t} \odot \tilde{h_{t}} + (1 - u_{t}) \odot h_{t - 1}\\
	z_{t} = h_{t}
	\end{gather}
	\label{eqn:gru_cell}
\end{subequations}

$u_{t}, r_{t} \in \mathbb{R}^{d}$ denote the update and reset gates respectively. $\tilde{h_{t}} \in \mathbb{R}^{d}$ indicates the candidate hidden state and $h_{t} \in \mathbb{R}^{d}$ indicates the current hidden state at time step t. The weights and biases follow the same notation as mentioned before. The reset gate decides how much of the previous hidden state contributes to the candidate state of the current step. Since the update gate functions alone without a forget gate, $(1 - u_{t})$ is used as an alternative. The GRU has gained a lot of popularity owing to its simplicity (lesser no. of parameters) compared to the LSTM cell and also its efficiency in training.

\subsubsection{Recurrent Neural Network Architectures}
\label{sec:architectures_review}
We use the following RNN architectures for our study.

\paragraph{Stacked Architecture}
\label{sec:stacked_architecture_review}

\begin{figure*}[!htb]
	\centering
	\includegraphics[width=\textwidth]{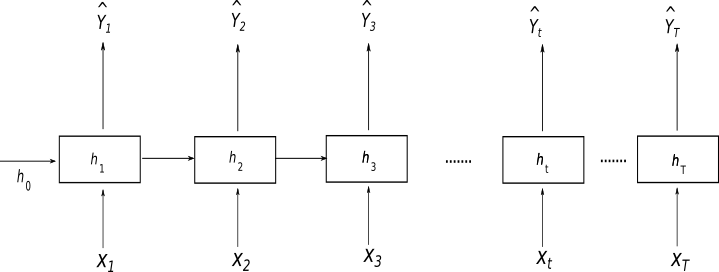}
	\caption{Stacked Architecture}
	\label{fig:stacked_architecture}
\end{figure*}

 The stacked architecture used in this study closely relates to the architecture mentioned in the work of \citet{Bandara2017-gb}. It is as illustrated in Figure \ref{fig:stacked_architecture}. 
\begin{figure*}[!htb]
	\centering
	\includegraphics[]{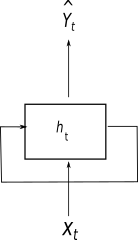}
	\caption{Folded Version of RNN}
	\label{fig:folded_version}
\end{figure*}

\begin{figure*}[!ht]
	\centering
	\includegraphics[scale=0.7]{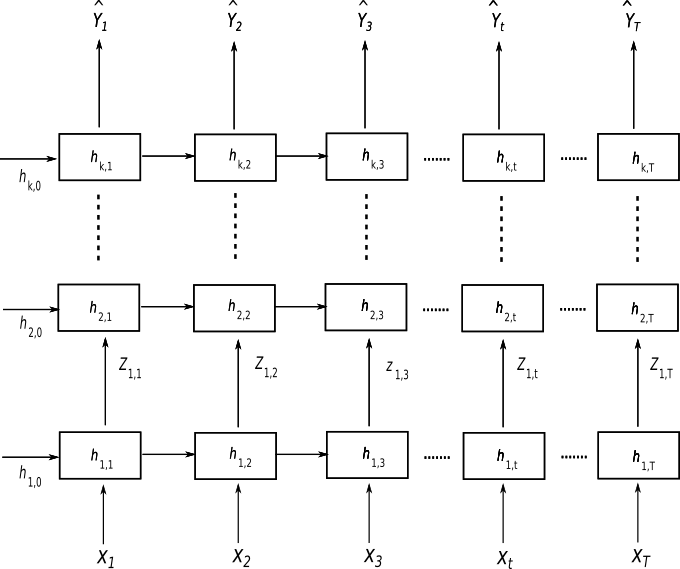}
	\caption{Multi-layer Stacked Architecture}
	\label{fig:multilayered_stacked_architecture}
\end{figure*}

Figure \ref{fig:folded_version} shows the folded version of the RNN while Figure \ref{fig:stacked_architecture} demonstrates the unfolded version through time. The idea is that the same RNN unit repeats for every time step, sharing the same weights and biases between each of them. The feedback loop of the cell helps the network to propagate the state $h_{t}$ to the future time steps. For the purpose of generalization only the state $h_{t}$ is shown in the diagram; However, for an LSTM cell, $h_{t}$ should be accompanied by the cell state $C_{t}$. The notion of stacking means that multiple LSTM layers can be stacked on top of one another. In the most basic setup, the model has only one LSTM layer. Figure \ref{fig:stacked_architecture} is for such a basic case, whereas Figure \ref{fig:multilayered_stacked_architecture} is for a multi-layered scenario. As seen in Figure \ref{fig:multilayered_stacked_architecture}, for many hidden layers the same structure as in Figure \ref{fig:stacked_architecture} is repeated multiple times stacked on top of each other where the output from every layer is directly fed as input to the next immediate layer above and the final forecasts retrieved from the last layer.

As seen in Figure \ref{fig:stacked_architecture}, $X_{t}$ denotes the input to the cell at time step t and $\hat{Y_{t}}$ corresponds to the output. $X_{t}$ and $\hat{Y_{t}}$ used for the stacking architecture are vectors instead of single data points. This is done according to the moving window scheme explained later in Section \ref{sec:multiple_output_strategy}. At every time step, the cell functions using its existing weights and produces the output $Z_t$ which corresponds to the next immediate output window of the time series. Likewise, the output of the RNN cell instance of the final time step $\hat{Y}_{T}$ corresponds to the expected forecasts for the particular time series. However, the output of the cell does not conform to the expected dimension which is the forecasting horizon (H). Since the cell dimension (d) is an externally tuned hyperparameter it may take on any appropriate value. Therefore, to project the output of the cell to the expected forecasting horizon, an affine neural layer is connected on top of every recurrent cell, whose weights are trained altogether with the recurrent network itself. In Figure \ref{fig:stacked_architecture}, this fully connected layer is not shown explicitly and the output $\hat{Y_{t}} \in \mathbb{R}^{H}$ corresponds to the output of the combined RNN cell and the dense layer.

During the model training process, the error is calculated per each time step and accumulated until the end of the time series. Let the error per each time step t be $e_t$. Then,

\begin{equation}
e_{t} = Y_{t} - \hat{Y}_t
\end{equation}

where ${Y}_{t}$ is the actual output vector at time step t, also preprocessed according to the moving window strategy. For all the time steps, the accumulated error E can be defined as follows.

\begin{equation}
E = \sum_{t=1}^{T} e_{t}
\end{equation}

At the end of every time series, the accumulated error E is used for the Backpropagation Through Time (BPTT) once for the whole sequence. This BPTT then updates the weights and biases of the RNN cells according to the optimizer algorithm used.

\paragraph{Sequence to Sequence Architecture}
\label{sec:seq2seq_architecture_review}

\begin{figure*}[!htb]
	\centering
	\includegraphics[width=\textwidth]{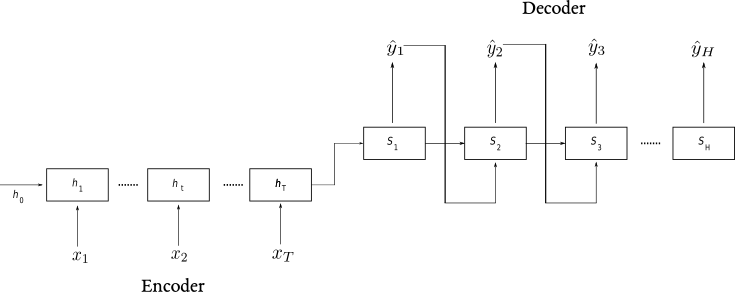}
	\caption{Sequence to Sequence with Decoder Architecture}
	\label{fig:seq2seq_with_dense}
\end{figure*}

The S2S architecture used in this study is as illustrated in Figure \ref{fig:seq2seq_with_dense}.
There are few major differences of the S2S network compared to the Stacked architecture. The first is the input format. The input $x_{t}$ fed to each cell instance of this network is a single data point instead of a vector. Stated differently, this network does not use the moving window scheme of data preprocessing. The RNN cells keep getting input at each time step and consequently build the state of the network. This component is known as the Encoder. However, in contrast to the Stacked architecture, the output is not considered per each time step; rather only the forecasts produced after the last input point of the Encoder are considered. Here, every $y_{t}$ corresponds to a single forecasted data point in the forecast horizon. The component that produces the outputs in this manner is called the Decoder. 

The Decoder comprises of a set of RNN cell instances as well, one per each step of the forecast horizon. The initial state of the Decoder is the final state built from the Encoder which is also known as the context vector. A distinct feature of the Decoder is that it contains autoregressive connections from the output of the previous time step into the input of the cell instance of the next time step. During training, these connections are disregarded and the externally fed actual output of each previous time step is used as means of teacher forcing. This teaches the Decoder component how inaccurate it is in predicting the previous output and how much it should be corrected. During testing, since the actual targets are not available, the autoregressive connections are used instead to substitute them with the generated forecasts. This is known as scheduled sampling where a decision is taken either to sample from the outputs or the external inputs at the Decoder. The decoder can also accept external inputs to support exogenous variables whose values are known for the future time steps. Since the hidden state size may not be equal to 1, which is the expected output size of each cell of the Decoder, an affine neural layer is applied on top of every cell instance of the Decoder similar to the Stacked architecture. The error computed for backpropagation in the S2S architecture differs from that of the Stacked architecture since no error accumulation happens over the time steps of the Encoder. Only the error at the Decoder is considered for the loss function of the optimizer. 

\begin{equation}
E = \sum_{t=1}^{H} y_{t} - \hat{y}_t 
\end{equation}

\begin{figure*}[!htb]
	\centering
	\includegraphics[width=\textwidth]{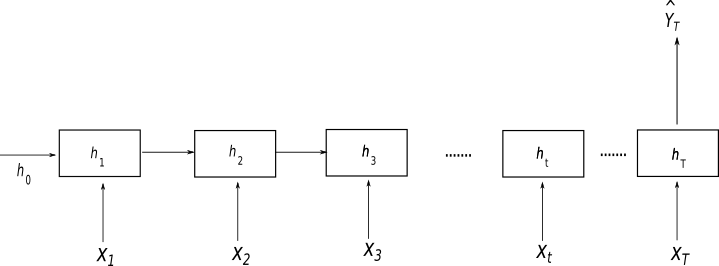}
	\caption{Sequence to Sequence with Dense Layer Architecture}
	\label{fig:seq2seq_with_dense_layer_architecture}
\end{figure*}

There are basically two types of components for output in a S2S network. The most common is the Decoder as mentioned above. Inspired by the work of \citet{Wen2017-xz}, a dense layer can also be used in place of the Decoder. We implement both these possibilities in our experiments. However, since we do not feed any future information into the decoder, we do not use the concept of local MLP as in \citet{Wen2017-xz} and use only one global MLP for the forecast horizon. This technique is expected to obviate the error propagation issue in the Decoder with autoregressive connections. This model is illustrated in Figure \ref{fig:seq2seq_with_dense_layer_architecture}.

As seen in Figure \ref{fig:seq2seq_with_dense_layer_architecture}, the network no longer contains a Decoder. Only the Encoder exists to take input per each time step. However, the format of the input in this model can be two fold; either with a moving window scheme or without it. Both these are tested in our study. In the scheme with the moving window, each Encoder cell receives the vector of inputs $X_t$ whereas without the moving window scheme, the input $x_t$ corresponds to a single scalar input as explained before for the network with the Decoder. The forecast for both these is simply the output of the last Encoder time step projected to the desired forecast horizon using a dense layer without bias (Fully connected layer). Therefore, the output of the last Encoder step is a vector both with a moving window or without. The error considered for the backpropagation is the error produced by this forecast. 

\begin{equation}
E = Y_{T} - \hat{Y}_T  
\end{equation}

In comparison to the Stacked architecture which is also fed with a moving window input, the only difference in this S2S architecture with the dense layer and the moving window input format is that in the former, the error is calculated per each time step and the latter calculates the error only for the last time step.

The set of models selected for implementation by considering all the aforementioned architectures and input formats is shown in Table \ref{tab:rnn_architectures}. All the models are implemented using the three RNN units Elman RNN cell, LSTM cell and the GRU cell and tested across the five datasets detailed further in Section \ref{sec:dataset_overview}. 

\begin{table*}
	\begin{center}
		\begin{tabular}{lccr}
			\toprule
			Architecture & Output Component & Input Format & Error Computation \\
			\hline
			Stacked & Dense Layer & Moving Window & Accumulated Error\\
			Sequence to Sequence & Decoder & Without Moving Window & Last Step Error\\
			Sequence to Sequence & Dense Layer & Without Moving Window & Last Step Error\\
			Sequence to Sequence & Dense Layer & Moving Window & Last Step Error\\
			\hline
		\end{tabular}
		\caption{RNN Architecture Information}
		\label{tab:rnn_architectures}
	\end{center}
\end{table*}

\subsection{Learning Algorithms}
\label{sec:learning_algorithms}

Three learning algorithms are tested in our framework; the Adam optimizer, the Adagrad optimizer and the COntinuous COin Betting (COCOB) optimizer~\citep{Kingma2014-pa, Duchi2011-ef, Orabona2017-qi}. Both Adam and Adagrad have built-in Tensorflow implementations while the COCOB optimizer has an open-source implementation which uses Tensorflow~\citep{cocob}. The Adam optimizer and the Adagrad optimizer both require the hyperparameter learning rate, which if poorly tuned may perturb the whole learning process. The Adagrad optimizer introduces the Adaptive Learning Rate concept where a separate learning rate is kept for each variable of the function to be optimized. Consequently, the different weights are updated using separate equations. However, the Adagrad optimizer is prone to shrinking learning rates over time as its learning rate update equations have accumulating gradients in the denominator. This slows down the learning process of the optimizer over time. The Adam optimizer, similar to Adagrad, keeps one learning rate per each parameter. However, to address the issue of vanishing learning rates, the Adam optimizer uses both exponentially decaying average of gradient moments and the gradients squared in its update equations. The Adam optimizer is generally expected to perform better than the other optimizers and \citet{Kingma2014-pa} empirically demonstrate this.

The COCOB optimizer attempts to minimize the loss function by self-tuning its learning rate. Therefore, this is one step closer to fully automating the NN modelling process since the user is relieved from the burden of defining the initial learning rate ranges for the hyperparameter tuning algorithm. Learning algorithms are extremely sensitive to their learning rate. Therefore, finding the optimal learning rate is crucial for the model performance. The COCOB algorithm is based on a coin betting scheme where during each iteration, an amount of money is bet on the outcome of a coin toss such that the total wealth in possession is maximized. \citet{Orabona2017-qi} apply the same idea to a function optimization where the bet corresponds to the size of the step taken along the axis of the independent variable. The total wealth and the outcome of the coin flip correspond to the optimum point of the function and the negative subgradient of the function at the bet point, respectively. During each round, a fraction of the current total wealth (optimum point) is bet. The betting strategy is designed such that the total wealth does not become negative at any point and the fraction of the money bet in each round increases until the outcome of the coin toss remains constant. For our context this means that as long as the signs of the negative subgradient evaluations remain the same, the algorithm keeps on making bigger steps along the same direction. This makes the convergence faster in contrast to other gradient descent algorithms which have a constant learning rate or a decaying learning rate where the convergence becomes slower close to the optimum. Similar to the Adagrad optimizer, COCOB too maintains separate coins (learning rates) for each parameter.    

%% file: sections/experimental_framework.tex
\section{Experimental Framework}
\label{sec:experimental_framework}
To implement the models we use version 1.12.0 of the Tensorflow open-source deep learning framework introduced by \citet{tensorflow2015-whitepaper}. This section details the different datasets used for the experiments along with their associated preprocessing steps, the information of the model training and testing procedures as well as the benchmarks used for comparison.

\subsection{Datasets}
\label{sec:dataset_overview}

The datasets used for the experiments are taken from the following forecasting competitions, held during the past few years. 

\begin{itemize}
\item CIF 2016 Forecasting Competition Dataset
\item NN5 Forecasting Competition Dataset
\item M3 Forecasting Competition Dataset
\item M4 Forecasting Competition Dataset
\item Wikipedia Web Traffic Time Series Forecasting Competition Dataset
\item Tourism Forecasting Competition Dataset
\end{itemize}

As mentioned earlier, our study is limited to using RNN architectures on univariate, multi step ahead forecasting considering only single seasonality, to be able to straightforwardly compare against automatic standard benchmark methods. For the NN5 dataset and the Wikipedia web traffic dataset, since they contain daily data with less than two years of data, we consider only the weekly seasonality. From the M3 and M4 datasets, we only use the monthly time series which contain a single yearly seasonality. The NN5, Wikipedia web traffic and the Tourism datasets contain non-negative series meaning that they also have 0 values. Table~\ref{tab:dataset_overview} gives further insight into these datasets.

\begin{table*}
\small
\begin{adjustwidth}{-0.7cm}{}
\begin{center}
\begin{tabular}{lccccr}
\toprule
Dataset Name & No. of Time Series & Forecasting Horizon & Frequency & Max. Length & Min. Length\\
\hline
CIF 2016 & 72 & 6, 12 & Monthly & 108 & 22\\
NN5 & 111 & 56 & Daily & 735 & 735\\
M3 &  1428 & 18 & Monthly &  126 & 48\\
M4 & 48,000 & 18 & Monthly & 2794 & 42\\
Wikipedia Web Traffic & 997 & 59 & Daily & 550 & 550\\
Tourism & 366 & 24 & Monthly & 309 & 67 \\
\hline
\end{tabular}
\caption{Dataset Information}
\label{tab:dataset_overview}
\end{center}
\end{adjustwidth}
\end{table*}

The CIF 2016 competition dataset has 72 monthly time series with 57 of them having a prediction horizon of 12 and the remaining 15 having a prediction horizon of 6 in the original competition. Some of the series having prediction horizon 6 are shorter than two full periods and are thus considered as having no seasonality. Out of all the time series, 48 series are artificially generated while the rest are real time series originating from the banking domain~\citep{Stepnicka2017-ly}. The NN5 competition was held in 2008. This dataset has in total 111 daily time series which represent close to two years of daily cash withdrawal data from ATM machines in the UK~\citep{Ben_Taieb2011-op}. The forecasting horizon for all time series is 56. The NN5 dataset also contains missing values. The methods used to compensate for these issues are as detailed in Section \ref{sec:missing_values}.

Two of the datasets are selected from the M competition series held over the years. From both M3 and M4 competitions we select only the monthly category which contains a single yearly seasonality. The yearly data contain no seasonality and the series are relatively shorter. In the quarterly data too, although they contain yearly seasonality, the series are shorter and contain fewer series compared to the monthly category. The M3 competition, held in 2000, has monthly time series with a prediction horizon of 18. In particular, this dataset consists of time series from a number of different categories namely, micro, macro, industry, finance, demography and other. The total number of time series in the monthly category is 1428~\citep{Makridakis2000-kk}. The M4 competition held in 2018 has a dataset with similar format to that in the M3 competition. It has the same 6 categories as stated before and the participants were required to make 18 months ahead predictions for the monthly series. Compared to the previous competitions, one of the key objectives of the M4 is to increase the number of time series available for forecasting~\citep{Makridakis2018-fc}. Consequently, the whole dataset consists of 100,000 series and the monthly subcategory alone has a total of 48,000 series. 

The other two datasets that we select for the experiments are both taken from Kaggle challenges, the Web Traffic Time Series Forecasting Competition for Wikipedia articles and the Tourism Dataset Competition~\citep{kaggle, tourism_challenge}. The task of the first competition is to predict the future web traffic (number of hits) of a given set of Wikipedia pages (145,000 articles), given their history of traffic for about two years. The dataset involves some metadata as well to denote whether the traffic comes from desktop, mobile, spider or all these sources. For this study, this problem is reduced to the history of the first 997 articles for the period of $1^{st}$ July 2015 to $31^{st}$ December 2016. The expected forecast horizon is from $1^{st}$ January 2017 to $28^{th}$ February 2017, covering 59 days. One important distinction in this dataset compared to the others is that all the values are integers. Since the NN outputs continuous values both positive and negative, to obtain the final forecasts, the values need to be rounded to the closest non-negative integer. The tourism dataset contains monthly, yearly and quarterly time series. Again, for our experiments we select the monthly category which has 366 series in total. The requirement in this competition for the monthly category is to predict the next 24 months of the given series. However, the nature of the data is generically termed as 'tourism related' and no specific details are given about what values the individual time series hold. The idea of the competition is to encourage the public to come up with new models that can beat the results which are initially published by \citet{tourism_athanasopoulos} using the same tourism dataset.  

Table \ref{tab:dataset_overview} also gives details of the lengths of different time series of the datasets. The CIF 2016, M3 and the Tourism datasets have relatively short time series (maximum length being 108, 126 and 309 respectively). The NN5 and Wikipedia Web Traffic dataset time series are longer. In the M4 monthly dataset, lengths of the series vary considerably from 42 to 2794.

\begin{figure*}[!htb]
\centering
    \includegraphics[width=\textwidth]{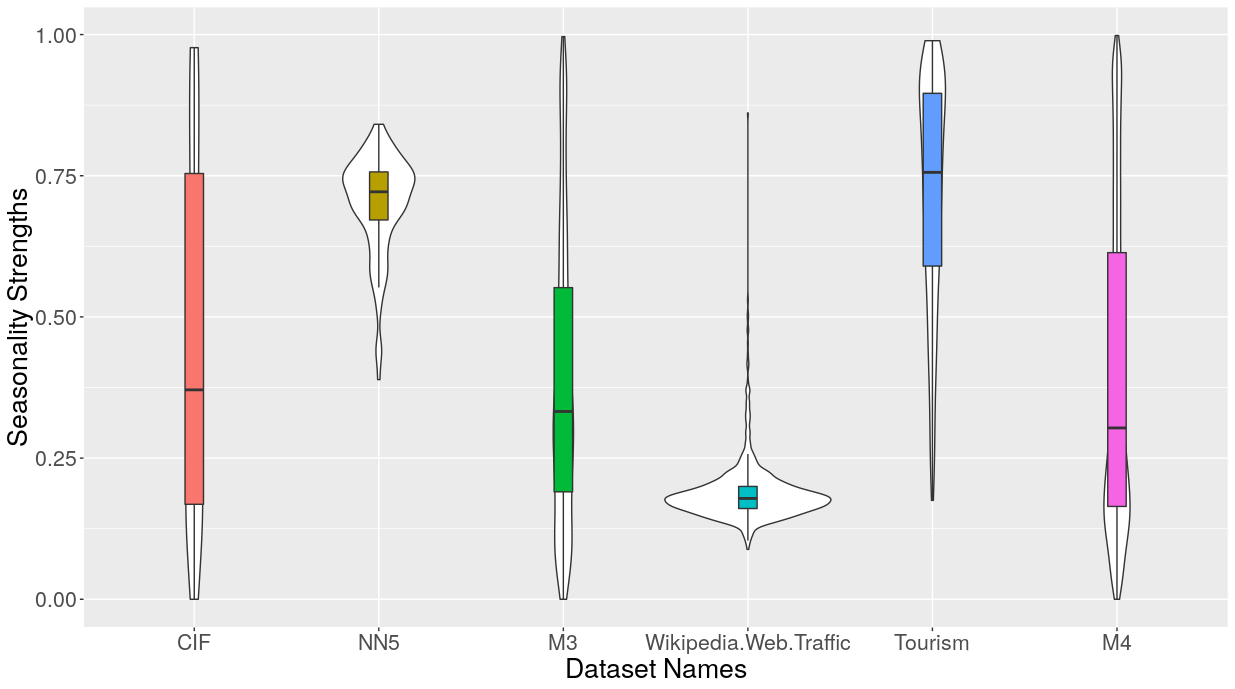}
    \caption{Violin Plots of Seasonality Strengths. The two datasets NN5 and Tourism have higher seasonality strength compared to the others. The inter-quantile ranges of the two respective violin plots are located comparatively higher along the y axis. Among these two, the seasonality strengths of the individual NN5 series vary less from each other since the inter-quantile range is quite narrow. For the Wikipedia Web Traffic dataset, the inter-quantile range is even more narrow and located much lower along the y axis. This indicates that the time series of the Wikipedia Web Traffic dataset carry quite minimal seasonality compared to the other datasets. For the CIF, M3 monthly and M4 monthly datasets, the seasonality strengths of the different time series are spread across a wide spectrum.}
  \label{fig:seasonality_strengths}
\end{figure*}

Figure \ref{fig:seasonality_strengths} shows violin plots of the seasonality strengths of the different datasets. To extract the seasonality strengths, we use the \texttt{tsfeatures} R package~\citep{tsfeatures}.

\subsection{Data Preprocessing}

We apply a number of preprocessing steps in our work that we detail in this section. Most of them are closely related to the ideas presented in~\citet{Bandara2017-gb}.

\subsubsection{Dataset Split}
\label{sec:dataset_split}

\begin{figure*}[!htb]
	\centering
	\captionsetup{justification=centering}
	\includegraphics[width=\textwidth]{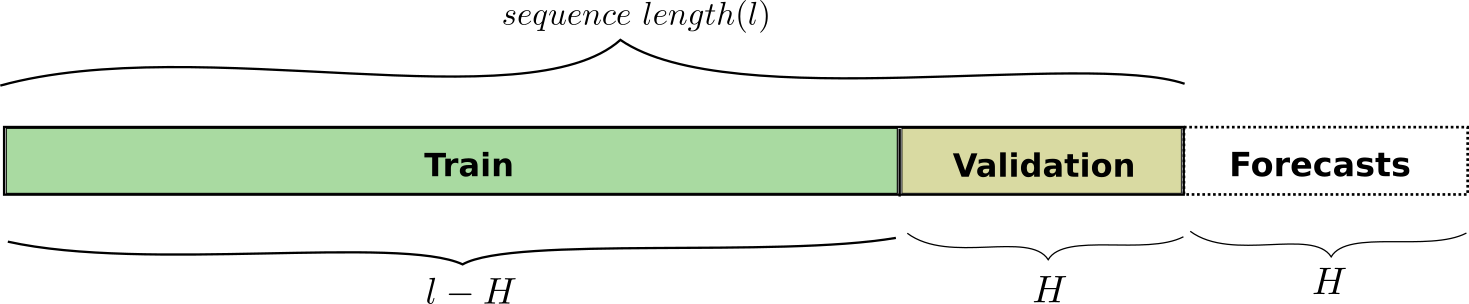}
	\caption{Train Validation Set Split}
	\label{fig:train_test_split}
\end{figure*}

The training and validation datasets are separated similar to the ideas presented by \citet{Suilin_undated-hh} and \citet{Bandara2017-gb}. From each time series, we reserve a part from the end for validation with a length equal to the forecast horizon. This is for finding the optimal values of the hyperparameters using the automated techniques explained in Section~\ref{sec:hyperparam}. The rest of the time series constitutes the training data. This approach is illustrated in Figure \ref{fig:train_test_split}. We use such fixed origin mechanism for validation instead of the rolling origin scheme since we want to replicate a usual competition setup. Also, given the fact that we have 36 RNN models implemented and tested across 6 datasets involving thousands of time series for both training and evaluation, we deem our results representative even with fixed origin evaluation and it would be computationally extremely challenging to perform a rolling origin evaluation.

However, as stated by \citet{Suilin_undated-hh}, this kind of split is problematic since the last part of the sequence is not considered for training the model. The further away the test predictions are from the training set the worse, since the underlying patterns may change during this last part of the sequence. Therefore, in this work, the aforementioned split is used only for the validation phase. For testing, the model is re-trained using the whole sequence without any data split. For each dataset, the models are trained using all the time series available, for the purpose of developing a global model. Different RNN architectures need to be fed data in different formats. 

\subsubsection{Addressing Missing Values}
\label{sec:missing_values}
Many machine learning methods cannot handle missing values, so that these need to be properly replaced by other appropriate substitutes. There are different techniques available to fill in missing values, Mean substitution and Median substitution being two of them. Linear interpolation and replacing by ``0" are other possible methods.
Out of the datasets selected for our experiments, the NN5 dataset and the Kaggle web traffic dataset contain missing values. 
For the NN5 dataset, we use a median substitution method. 
Since the NN5 dataset includes daily data, a missing value on a particular day is replaced by the median across all the same days of the week along the whole series. For example, if a missing value is encountered on a Tuesday, the median of all values on Tuesdays of that time series is taken as the substitute. This approach seems superior to taking the median across all the available data points, since the NN5 series have a strong weekly seasonality. Compared to the NN5 dataset, the Kaggle web traffic dataset has many more missing values. In addition to that, this dataset does not differentiate between missing values and ``0" values. Therefore, for the Kaggle web traffic dataset, a simple substitution by ``0"s is carried out for all the missing values.

\subsubsection{Modelling Seasonality}
\label{sec:deseasonalization}

With regard to modelling seasonality using NNs, there have been mixed notions among the researchers over the years. Some of the early works in this space infer that NNs are capable of modelling seasonality accurately~\citep{Sharda1992-lz, Tang1991-jz}. However, more recent experiments suggest that deseasonalization prior to feeding data to the NNs is essential since NNs are weak in modelling seasonality. Particularly, \citet{Claveria2017-ym} empirically show for a tourism demand forecasting problem that seasonally adjusted data can boost the performance of NNs especially in the case of long forecasting horizons. \citet{Zhang2005-le} conclude that using both detrending and deseasonalization can improve the forecasting accuracy of NNs. Similar observations are recorded in the work of \citet{Zhang2007-jk} as well as \citet{Nelson1999-pa}. More recently, in the winning solution by \citet{smyl2020esrnn} at the M4 forecasting competition, the same argument that NNs are weak at modelling seasonality, is put forward. 
A core part of our research is to investigate whether NNs actually struggle to model seasonality on their own. Therefore, we run experiments with and without removing the seasonality.
In the following, we describe the deseasonalization procedure we use in case we remove the seasonality.

Since in theory deterministic seasonality does not change and is therewith known ahead of time, relieving the NN from the burden of modelling it can ease the task of the NN and let it predict only the non-deterministic parts of the time series. Following this general consensus, we run a set of experiments with prior deseasonalization of the time series data. For that we use the Seasonal and Trend Decomposition using Loess (STL Decomposition) introduced by \citet{Cleveland1990-bt} as a way of decomposing a time series into its seasonal, trend and remainder components. Loess is the underlying method for estimating non-linear relationships in the data. By the application of a sequence of Loess smoothers, the STL Decomposition method can efficiently separate the seasonality, trend and remainder components of the time series. However, this technique can only be used with additive trend and seasonality components. Therefore, in order to convert all multiplicative time series components to additive format, the STL Decomposition is immediately preceded by a variance stabilization technique as discussed in Section~\ref{sec:variance_stabilization}. The STL Decomposition is potentially able to allow the seasonality to change over time. However, we use STL in a deterministic way, where we assume the seasonality of all the time series to be fixed along the whole time span.

In particular, we use the implementation of the STL Decomposition available in the forecast R package introduced by \citet{Hyndman2008-gt}. By specifically setting the s.window parameter in the stl method to ``periodic", we make the seasonality deterministic. Hence, we remove only the deterministic seasonality component from the time series while other stochastic seasonality components may still remain. The NN is expected to model such stochastic seasonality by itself. The STL Decomposition is applied in this manner to all the time series, regardless of whether they actually show seasonal behaviour or not. Nevertheless, this technique requires at least two full periods of the time series data to determine its seasonality component. In extreme cases where the full length of the series is less than two periods, the technique considers such sequences as having no seasonality and returns 0 for the seasonality component.

\subsubsection{Stabilizing the Variance in the Data}
\label{sec:variance_stabilization}
Variance stabilization is necessary if an additive seasonality decomposition technique such as the STL decomposition in Section \ref{sec:deseasonalization} is used. For time series forecasting, variance stabilization can be done in many ways, and applying a logarithmic transformation is arguably the most straightforward way to do so. However, a shortcoming of the logarithm is that it is undefined for negative and zero-valued inputs. All values need to be on the positive scale for the logarithm. For non-negative data, this can be easily resolved by defining a Log transformation as in Equation~\ref{eqn:log}, as a slightly altered version of a pure logarithm transformation. 

\begin{equation} \label{eqn:log}
w_{t}=\begin{cases}
log(y_t) & \text{if}\ min(y) > \epsilon,\\ 
log(y_t + 1) & \text{if}\ min(y) \leq \epsilon
\end{cases}
\end{equation}

$y$ in Equation \ref{eqn:log} denotes the whole time series. For count data, $\epsilon$ can be defined as equal to 0, whereas for real-valued data, $\epsilon$ can be selected as a small positive value close to 0. 
The logarithm is a very strong transformation and may not always be adequate. There are some other similar transformations that attempt to overcome this shortcoming. One such transformation is the Box-Cox transformation. It is defined as follows.

\begin{equation}
\label{eqn:box_cox_transformation}
w_{t}=\begin{cases}
log(y_t) & \text{if}\ \lambda = 0,\\ 
(y_{t}^{\lambda} - 1)/\lambda & \text{if}\ \lambda \neq 0
\end{cases}
\end{equation}

As indicated by Equation \ref{eqn:box_cox_transformation}, the Box-Cox transformation is a combination of a log transformation (when $\lambda = 0$) and a power transformation (when $\lambda \neq 1$) denoted by $y_{t}^{\lambda}$. The parameter $\lambda$ needs to be carefully chosen. In the case when $\lambda$ equals 1, the Box-Cox transformation results in $y_{t} - 1$. Thus, the shape of the data is not changed but the series is simply shifted down by 1~\citep{robjhyndmangeorgeathanasopoulos2018}. Another similar form of power transformation is the Yeo-Johnson transformation~\citep{Yeo2000-rr} shown in Equation~\ref{equ:YeoJohnson}. 

\begin{equation} \label{equ:YeoJohnson}
w_{t}=\begin{cases}
((y_t + 1)^{\mu} - 1)/\mu & \text{if}\ \mu \neq 0, y_t \geq 0,\\ 
log(y_t + 1) & \text{if}\ \mu = 0, y_t \geq 0,\\
-[(-y_t + 1)^{(2-\mu)} - 1]/(2 - \mu) & \text{if}\ \mu \neq 2, y_t < 0,\\
-log(-y_t + 1) & \text{if}\ \mu = 2, y_t < 0
\end{cases}
\end{equation}

The parameter $\mu$ in the Yeo-Johnson transformation is confined in the range $[0 ,2]$ while $\mu = 1$ gives the identity transformation. Values of $y_t$ are allowed to be either positive, negative or zero. Even though this is an advantage over the logarithm, the choice of the optimal value of the parameter $\mu$ is again not trivial. 

Due to the complexities associated with selecting the optimal values of the parameters $\lambda$ and $\mu$ in the Box-Cox and Yeo-Johnson Transformations, and given that all data used in our experiments are non-negative, we use a Log transformation in our experiments.

\subsubsection{Multiple Output Strategy}
\label{sec:multiple_output_strategy}
Forecasting problems are typically multi-step-ahead forecasting problems, which is also what we focus on in our experiments. \citet{Ben_Taieb2011-op} perform an extensive research involving five different techniques for multi-step-ahead forecasting. The Recursive Strategy which is a sequence of one-step-ahead forecasts, involves feeding the prediction from the last time step as input for the next prediction. The Direct Strategy involves different models, one per each time step of the forecasting horizons. The DiRec Strategy combines the concepts from the two above methods and develops multiple models one per each forecasting step where the prediction from each model is fed as input to the next consecutive model. In all these techniques, the model outputs basically a scalar value corresponding to one forecasting step. 
The other two techniques tested by \citet{Ben_Taieb2011-op} use a direct multi-step-ahead forecast where a vector of outputs corresponding to the whole forecasting horizon is directly produced by the model. The first technique under this category is known as the Multi-Input Multi-Output (MIMO) strategy. The advantage of the MIMO strategy comes from producing the forecasts for the whole output window at once, and thus incorporating the inter-dependencies between each time step, rather than forecasting each time step in isolation. The DIRMO technique synthesizes the ideas of the Direct technique and the MIMO technique where each model produces forecasts for windows of size $s$ ($s \in \{1,...,H\}, H$ being the forecast horizon). In the extreme cases where $s = 1$ and $s = H$ the technique narrows down to the Direct strategy and the MIMO strategy, respectively. The experimental analysis in that paper using the NN5 competition dataset demonstrates that the multiple output strategies are the best overall. The DIRMO strategy requires careful selection of the output window size $s$~\citep{Ben_Taieb2011-op}. \citet{Wen2017-xz} also state in their work that the Direct Multi-Horizon strategy also known as the MIMO strategy performs reliably since it avoids error accumulation over the prediction time steps. 

\begin{figure*}[!htb]
	\centering
	\captionsetup{justification=centering}
	\includegraphics[width=\textwidth]{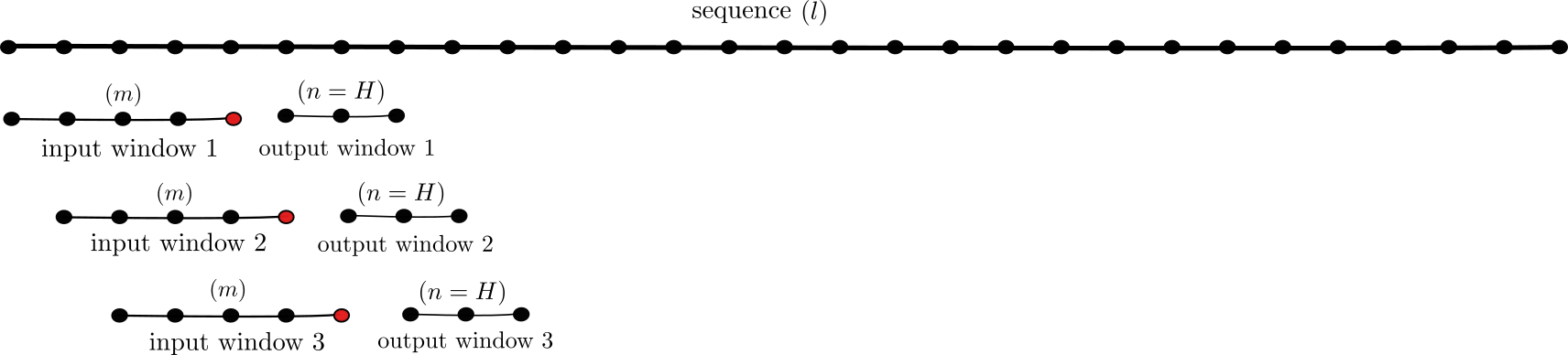}
	\caption{Moving Window Scheme}
	\label{fig:moving_windows}
\end{figure*}

Following these early findings, we use in our study the multiple output strategy in all the RNN architectures. However, the inputs and outputs of each recurrent cell differ between the different architectures. For the S2S model, each recurrent cell of the encoder is fed a single scalar input. For the stacking model and the S2S model with the dense layer output, a moving window scheme is used to feed the inputs and create the outputs as in the work of \citet{Bandara2017-gb}. In this approach, every recurrent cell accepts a window of inputs and produces a window of outputs corresponding to the time steps which immediately follow the fed inputs. The recurrent cell instances of the next consecutive time step accepts an input window of the same size as the previous cell, shifted forward by one. This method can also be regarded as an effective data augmentation mechanism~\citep{Smyl2016-rf}. Figure \ref{fig:moving_windows} illustrates this scheme.

Let the length of the whole sequence be $l$, the size of the input window $m$ and the size of the output window $n$. As mentioned in Section \ref{sec:dataset_split}, during the training phase, the last $n$ sized piece from this sequence is left out for validation. The rest of the sequence (of length $l - n$) is broken down into blocks of size $m + n$ forming the input output combination for each recurrent cell instance. Likewise in total, there are $l - n * 2 - m$ and $l - n - m$ such blocks for the training and the validation stages respectively. Even though the validation stage does not involve explicit training of the model, the created blocks should still be fed in sequence to build up the state. The output window size is set to be equal to the size of the forecasting horizon $H$ ($n = H$). The selection of the input window size is a careful task. The objective of using an input window as opposed to single input is to relax the duty of the recurrent unit to remember the whole history of the sequence. Though theoretically the RNN unit is expected to memorize all the information from the whole sequence it has seen, it is not practically competent in doing so~\citep{Smyl2016-rf}. The importance of the information from distant time steps tend to fade as the model continues to see new inputs. 
However, in terms of the trend, it is intelligible that only the last few time steps contribute the most in forecasting. 

Putting these ideas together, we select the input window size $m$ with two options. One is to make the input window size slightly bigger than the output window size ($m = 1.25 * output\_window\_size$). The other is to make the input window size slightly bigger than the seasonality period ($m = 1.25 * seasonality\_period$). For instance if the time series has weekly seasonality ($seasonality\_period = 7$) with expected forecasting horizon $56$, with the first option we set the input window size to be $70$ ($1.25 * 56$), whereas with the second option it is set to be $9$ ($1.25 * 7$). The constant 1.25 is selected purely as a heuristic. The idea is to ascertain that every recurrent cell instance at each time step gets to see at least its last periodic cycle so that it can model any remaining stochastic seasonality. In case, the total length of the time series is too short, $m$ is selected to be either of the two feasible, or some other possible value if none of them work. For instance, the forecasting horizon 6 subgroup of the CIF 2016 dataset comprises of such short series and $m$ is selected to be 7 (slightly bigger than the output window size) even though the $seasonality\_period$ is equal to 12.

\subsubsection{Trend Normalization}
\label{sec:normalization}

The activation functions used in RNN cells, such as the sigmoid or the hyperbolic tangent function have a saturation area after which the outputs are constant. Hence, when using RNN cells it should be assured that the inputs fed are normalized adequately such that the outputs do not lie in the saturated range~\citep{Smyl2016-rf}. To this end, we have performed a per window local normalization step for the RNN architectures which use the moving window scheme. From each deseasonalized input and corresponding output window pair, the trend value of the last time step of the input window (found by applying the STL Decomposition) is deducted. This is carried out for all the input and output window pairs of the moving window strategy. The red coloured time steps in Figure \ref{fig:moving_windows} correspond to those last time steps of all the input windows. This technique is inspired by the batch normalization scheme and it also helps address the trend in time series~\citep{batch_normalization}. As for the models that do not use the moving window scheme, a per sequence normalization is carried out where the trend value of the last point of the whole training sequence is used for the normalization instead.

\subsubsection{Mean Normalization}

For the set of experiments that do not use STL Decomposition, it is not possible to perform the trend normalization as described in Section \ref{sec:normalization}. For those experiments, we perform a mean normalization of the time series before applying the Log transformation mentioned in Section \ref{sec:variance_stabilization}. In this mean normalization, every time series is divided by the mean of that particular time series. The division instead of subtraction further helps to scale all the time series to a similar range which helps the RNN learning process. The Log transformation is then carried out on the resulting data.

\subsection{Training \& Validation Scheme}

The methodology used for training the models is detailed in this section. Our work uses the idea of global models in Section \ref{sec:cross_series_lit_review} and develops one model for all the available time series. However, it is important to understand that the implemented techniques work well only in the case of related (similar/homogeneous) time series. If the individual time series are from heterogeneous domains, modeling such time series using a single model may not render the best results. 

Due to the complexity and the total number of models associated with the study, we run the experiments on three computing resources in parallel. The details of these machines are as mentioned in Table \ref{tab:hardware_specifications}. Due to the scale of the M4 monthly dataset, the GPU machines are used to run the experiments on it. Resource 2 and Resource 3 indicate allocated resources from the Massive cluster~\citep{m3_cluster}. Resource 2 is used for GPU only operations whereas Resource 3 is used for CPU only operations. Resource 1 is used for both CPU and GPU operations.

\begin{table*}
	\small
	\begin{center}
		\begin{tabular}{lccr}
			\toprule
			Specification & Resource 1 & Resource 2 & Resource 3\\
			\hline
			CPU Model Name & Intel(R) Core(TM) i7-8700 & Intel Xeon Gold 6150
			 & Intel Xeon CPU E5-2680 v3\\
			CPU Architecture & x86\_64 & x86\_64 & x86\_64\\
			CPU Cores & 12 & 1 & 4\\
			Memory(GB) & 64 & 50 & 5\\
			GPU Model Name& GP102 [GeForce GTX 1080 Ti] & nVidia Tesla V100 & -\\
			No. of GPUs & 2 & 2 & -\\
			\hline
		\end{tabular}
		\caption{Hardware Specifications}
		\label{tab:hardware_specifications}
	\end{center}
\end{table*}

\subsubsection{Hyper-parameter Tuning}
\label{sec:hyperparam}

The experiments require the proper tuning of a number of hyperparameters associated with the model training. They are as mentioned below.

\begin{enumerate}
\item Minibatch Size
\item Number of Epochs
\item Epoch Size
\item Learning Rate
\item Standard Deviation of the Gaussian Noise
\item L2 Regularization Parameter
\item RNN Cell Dimension
\item Number of Hidden Layers
\item Standard Deviation of the Random Normal Initializer
\end{enumerate}

Minibatch size denotes the number of time series considered for each full backpropagation in the RNN. This is a more limited version than using all the available time series at once to perform one backpropagation, which poses a significant memory requirement. On the other hand, this could also be regarded as a more generalized version of the extreme case which uses only a single time series per each full backpropagation, also known as stochastic gradient descent. Apart from the explicit regularization schemes used in the RNN models further discussed in Section \ref{sec:model_overfitting}, minibatch gradient descent also introduces some implicit regularization to the deep learning models in specific problem scenarios such as classification~\citep{Soudry2018GradientDescent}. An epoch denotes one full forward and backward pass through the whole dataset. Therefore, the number of epochs denote how many such passes across the dataset are required for the optimal training of the RNN. Even within each epoch, the dataset is traversed a number of times denoted by the epoch size. This is especially useful for those datasets with limited number of time series to increase the number of datapoints available for training. This is because NNs are supposed to be trained best when the amount of data available is higher. 

Out of the three optimizers used in the experiments, the Adam optimizer and the Adagrad optimizer both require the appropriate tuning of the learning rate to converge to the optimal state of the network parameters fast. To reduce the effect of model overfitting, two steps are taken; adding Gaussian noise to the input and using L2 regularization for the loss function (explained further under Section \ref{sec:model_overfitting}). Both these techniques require tuning hyperparameters; the standard deviation of the Gaussian noise distribution and the L2 regularization parameter. Furthermore, the weights of the RNN units are initialized using random samples drawn from a normal distribution whose standard deviation is tuned as another hyperparameter. 

There are two other hyperparameters that directly relate to the RNN network architecture; the number of hidden layers and the cell dimension of each RNN cell. For simplicity, both the Encoder and the Decoder components of the Sequence to Sequence network are composed of the same number of hidden layers. Also, the same cell dimension is used for the RNN cells in both the Encoder and the Decoder. However, as explained before in Section \ref{sec:recurrent_units_review}, the different Recurrent unit types that we employ in this study, namely the LSTM cell, ERNN cell, and the GRU, have different numbers of trainable parameters for the same cell dimension, with the LSTM having the highest and the ERNN having the lowest number of parameters respectively. Therefore, to enable a fair comparison, in other research communities such as natural language processing, it is common practice to consider the total number of trainable parameters in the different models compared~\citep{Collins2016CapacityAT, Ji2016Trainable, Jagannatha2016ncbi}. This is to ascertain that the capacities of all the compared models are the same to clearly distinguish the performance gains achieved purely through the novelty introduced to the models. As our main aim is to compare software frameworks, we use the cell dimension which is the natural hyperparameter to be tuned by practitioners, and additionally perform experiments using the number of trainable hyperparameters instead. For this, we choose the best model combination identified from the preceding experiments involving the cell dimension as a hyperparameter and run it again along with all three recurrent unit types by letting the hyperparameter tuning algorithm choose the optimal number of trainable parameters as a hyperparameter. Hence, for this experiment, instead of setting the cell dimension directly, we set the same initial range for the number of trainable parameters across all the compared models with the three RNN units. 
As the number of trainable parameters is directly proportional to the cell dimension, and the deep learning framework we use allows us to only set the cell dimension, we then calculate the corresponding cell dimension from the number of parameters, and set this parameter accordingly.

For tuning hyperparameters, there are many different techniques available. The most na\"ive and the fundamental approach is hand tuning which requires intensive manual experimentation to find the best possible hyperparamters. However, our approach is more targeted towards a fully automated framework. To this end, Grid Search and Random Search are two of the most basic methods for automated hyperparameter tuning~\citep{Bergstra2012-ze}. Grid Search, as its name suggests explores in the space of a grid of different hyperparameter value combinations. For example, if there are two hyperparameters $\Gamma$ and $\Theta$, a set of values are specified for each hyperparameter as in $(\gamma_1, \gamma_2, \gamma_3, ...)$ for $\Gamma$ and $(\theta_1, \theta_2, \theta_3, ...)$ for $\Theta$. Then the Grid Search algorithm goes through each pair of hyperparameter values $(\gamma, \theta)$ that lie in the grid of $\Gamma$ and $\Theta$ and uses them in the model to calculate the error on a held out validation set. The hyperparameter combination which gives the minimum error is chosen as the optimal hyperparameter values. This is an exhaustive search through the grid of possible hyperparameter values. In the Random Search algorithm, instead of the exact values, distributions for the hyperparameters are provided, from which the algorithm randomly samples values for each model evaluation. A maximum number of iterations are provided for the algorithm until when the model evaluations are performed on the validation set and the optimal combination thus far is selected.There are other more sophisticated hyperparameter tuning techniques as well.    

\paragraph{Bayesian Optimization}

The methods discussed above evaluate an objective function (validation error) point wise. This means retraining machine learning models from scratch which is quite compute intensive. The idea of the Bayesian Optimization is to limit the number of such expensive objective function evaluations. The technique models the objective function using a Gaussian prior over all possible functions~\citep{Snoek2012-sq}. This probabilistic model embeds all prior assumptions regarding the objective function to be optimized. During every iteration, another instance of the hyperparameter space is selected to evaluate the objective function value. The decision of which point to select next, depends on the minimization/maximization of a separate acquisition function which is much cheaper to evaluate than the objective function itself. The acquisition function can take many forms out of which the implementation of \citet{Snoek2012-sq} uses Expected Improvement. The optimization of the acquisition function considers all the previous objective function evaluations to decide the next evaluation point. Therefore, the Bayesian Optimization process takes on smarter decisions of function evaluation compared to the aforementioned Grid Search and Random Search. Given an initial range of values for every hyperparameter and a defined number of iterations, the Bayesian Optimization method initializes using a set of prior known parameter configurations and thus attempts to find the optimal values for those hyperparameters for the given problem. There are many practical implementations and variants of the basic Bayesian Optimization technique such as hyperopt, spearmint and bayesian-optimization~\citep{hyperopt, spearmint, bayes-opt}.

\paragraph{Sequential Model based Algorithm Configuration (SMAC)}

SMAC for hyperparameter tuning, is a variant of Bayesian Optimization proposed by \citet{Hutter_undated-kp}. This implementation is based on the Sequential Model Based Optimization (SMBO) concept. SMBO is a technique founded on the ideas of the Bayesian Optimization but uses a Tree-structured Parzen Estimator (TPE) for modelling the objective function instead of the Gaussian prior. The TPE algorithm produces a set of candidate optimal values for the hyperparameters in each iteration, as opposed to just one candidate in the usual Bayesian Optimization. The tree structure lets the user define conditional hyperparameters which depend on one another. The process is carried out until a given time bound or a number of iterations is reached. The work by \citet{Hutter_undated-kp} further enhances the basic SMBO algorithm by adding the capability to work with categorical hyperparameters. The Python implementation of SMAC used for this study is available as a Python package~\citep{smac-2017}. We use the version 0.8.0 for our experiments. The initial hyperparameter ranges used for the NN models across the different datasets are as listed in Table \ref{tab:hyperparameter_ranges}. Additionally, for the experiment which involves the number of trainable parameters instead of the cell dimension as a hyperparameter, we set an initial range of $2000 - 25000$ across all the compared models in all the selected datasets. This range is selected based on the initial experiments involving the cell dimension.

\begin{table*}
	\begin{center}
		\resizebox{\textwidth}{!}{
		\begin{tabular}{lcccccccccc}
			\hline
			Dataset & Batch Size & Epochs & Epoch Size & Std. Noise & L2 Reg. & Cell Dim. & Layers & Std. Initializer & \multicolumn{2}{c}{Learning Rate}\\
			\cline{10-11}
			&&&&&&&&& Adam & Adagrad\\
			\hline 
			CIF (12) & 10 - 30 & 3 - 25 & 5 - 20 & 0.01 - 0.08 & 0.0001 - 0.0008 & 20 - 50 & 1 - 2 & 0.0001 - 0.0008 & 0.001 - 0.1 & 0.01 - 0.9\\
			CIF (6) & 2 - 5 & 3 - 30 & 5 - 15 & 0.0001 - 0.0008 & 0.0001 - 0.0008 & 20 - 50 & 1 - 5 & 0.0001 - 0.0008 & 0.001 - 0.1 & 0.01 - 0.9\\
			NN5 & 5 - 15 & 3 - 25 & 2 - 10 & 0.0001 - 0.0008 & 0.0001 - 0.0008 & 20 - 25 & 1 - 2 & 0.0001 - 0.0008 & 0.001 - 0.1 & 0.01 - 0.9\\
			M3(Mic) & 40 - 100 & 3 - 30 & 2 - 10 & 0.0001 - 0.0008 & 0.0001 - 0.0008 & 20 - 50 & 1 - 2 & 0.0001 - 0.0008 & 0.001 - 0.1 & 0.01 - 0.9\\
			M3(Mac) & 30 - 70 & 3 - 30 & 2 - 10 & 0.0001 - 0.0008 & 0.0001 - 0.0008 & 20 - 50 & 1 - 2 & 0.0001 - 0.0008 & 0.001 - 0.1 & 0.01 - 0.9\\
			M3(Ind) & 30 - 70 & 3 - 30 & 2 - 10 & 0.0001 - 0.0008 & 0.0001 - 0.0008 & 20 - 50 & 1 - 2 & 0.0001 - 0.0008 & 0.001 - 0.1 & 0.01 - 0.9\\
			M3(Dem) & 20 - 60 & 3 - 30 & 2 - 10 & 0.0001 - 0.0008 & 0.0001 - 0.0008 & 20 - 50 & 1 - 2 & 0.0001 - 0.0008 & 0.001 - 0.1 & 0.01 - 0.9\\
			M3(Fin) & 20 - 60 & 3 - 30 & 2 - 10 & 0.0001 - 0.0008 & 0.0001 - 0.0008 & 20 - 50 & 1 - 2 & 0.0001 - 0.0008 & 0.001 - 0.1 & 0.01 - 0.9\\
			M3(Oth) & 10 - 30 & 3 - 30 & 5 - 20 & 0.0001 - 0.0008 & 0.0001 - 0.0008 & 20 - 50 & 1 - 2 & 0.0001 - 0.0008 & 0.001 - 0.1 & 0.01 - 0.9\\
			M4(Mic) & 1000 - 1500 & 3 - 25 & 2 - 10 & 0.0001 - 0.0008 & 0.0001 - 0.0008 & 20 - 50 & 1 - 2 & 0.0001 - 0.0008 & 0.001 - 0.1 & 0.01 - 0.9\\
			M4(Mac) & 1000 - 1500 & 3 - 25 & 2 - 10 & 0.0001 - 0.0008 & 0.0001 - 0.0008 & 20 - 50 & 1 - 2 & 0.0001 - 0.0008 & 0.001 - 0.1 & 0.01 - 0.9\\
			M4(Ind) & 1000 - 1500 & 3 - 25 & 2 - 10 & 0.0001 - 0.0008 & 0.0001 - 0.0008 & 20 - 50 & 1 - 2 & 0.0001 - 0.0008 & 0.001 - 0.1 & 0.01 - 0.9\\
			M4(Dem) & 850 - 1000 & 3 - 25 & 2 - 10 & 0.0001 - 0.0008 & 0.0001 - 0.0008 & 20 - 50 & 1 - 2 & 0.0001 - 0.0008 & 0.001 - 0.1 & 0.01 - 0.9\\
			M4(Fin) & 1000 - 1500 & 3 - 25 & 2 - 10 & 0.0001 - 0.0008 & 0.0001 - 0.0008 & 20 - 50 & 1 - 2 & 0.0001 - 0.0008 & 0.001 - 0.1 & 0.01 - 0.9\\
			M4(Oth) & 50 - 60 & 3 - 25 & 2 - 10 & 0.0001 - 0.0008 & 0.0001 - 0.0008 & 20 - 25 & 1 - 2 & 0.0001 - 0.0008 & 0.001 - 0.1 & 0.01 - 0.9\\
			Wikipedia & 200 - 700 & 3 - 25 & 2 - 10 & 0.0001 - 0.0008 & 0.0001 - 0.0008 & 20 - 25 & 1 - 2 & 0.0001 - 0.0008 & 0.001 - 0.1 & 0.01 - 0.9\\
			Tourism & 10 - 90 & 3 - 25 & 2 - 10 & 0.0001 - 0.0008 & 0.0001 - 0.0008 & 20 - 25 & 1 - 2 & 0.0001 - 0.0008 & 0.001 - 0.1 & 0.01 - 0.9
			\\\hline
		\end{tabular}
	}
		\caption{Initial Hyperparameter Ranges}
		\label{tab:hyperparameter_ranges}
	\end{center}
\end{table*}

\subsubsection{Dealing with Model Overfitting}
\label{sec:model_overfitting}

Overfitting in a machine learning model is when the performance on the validation dataset is much worse than the performance on the training dataset. This means that the model has fitted quite well onto the training data to the extent that it has lost its generalization capability to model other unseen data. In NN models, this mostly happens due to the complexity ensued from the large number of parameters. In order to make the models more versatile and capture unseen patterns, we are using two techniques in our framework. One method is to add noise to the input to distort it partially. The noise distribution applied here is a Gaussain distribution with zero mean and the standard deviation treated as another hyperparameter. The other technique is L2 weight regularization which uses a Ridge penalty in the loss function. Weight regularization avoids the weights of the network from growing excessively by incorporating them in the loss function and thus penalizing the network for increasing its complexity. Given the L2 regularization parameter $\psi$, the Loss function for the NN can be defined as below.

\begin{equation}
    \label{eqn:l2_regularization}
    L = E + \underbrace{\psi\sum_{i=1}^{p} w_{i}^2}_\text{L2 regularization}
\end{equation}

Applied to our context, $E$ in Equation \ref{eqn:l2_regularization} denotes the Mean Absolute Error from the network outputs. $w_i$ represents the trainable parameters of the network where p is the number of all such parameters. Hence, as seen in Equation \ref{eqn:l2_regularization}, L2 regularization adds the squared magnitudes of the weights multiplied by the regularization parameter $\psi$. The selection of $\psi$ is paramount since too large a value results in underfitting and on the other hand very small values let the model become overfitted. In our study, $\psi$ is also optimized as a hyperparameter. 

\subsection{Model Testing}
Once an optimal hyperparameter combination is found, those values are used to train the model and get the final forecasts. During testing, the model is trained using all the data available and the final forecasts are written to files. We address parameter uncertainty by training all the models on 10 different Tensorflow graph seeds but using the same initially tuned hyperparameters. We do not re-run the hyperparameter tuning as this would be computationally very expensive. The different seeds give 10 different initializations to the networks. Once the RNN forecasts are obtained from every model for the different seeds, they are ensembled by taking the median across the seeds. 

\subsubsection{Data Post-Processing}
To calculate the final error metrics, the effects of the prior preprocessing is properly reversed on the generated forecasts. For the experiments which use STL Decomposition, this post-processing is carried out as follows.

\begin{enumerate}
    \item Reverse the local normalization by adding the trend value of the last input point.
    \item Reverse deseasonalization by adding back the seasonality components.
    \item Reverse the log transformation by taking the exponential.
    \item Subtract 1, if the data contain 0s
    \item For integer data, round the forecasts to the closest integer.
    \item Clip all negative values at 0 (To allow for only positive values in the forecasts).
\end{enumerate}

For those experiments which do not use STL Decomposition, the post-processing of the forecasts is done as follows.

\begin{enumerate}
	\item Reverse the log transformation by taking the exponential.
	\item Subtract 1, if the data contain 0s
	\item Multiply the forecasts of every series by its corresponding mean to reverse the effect of mean scaling.
	\item For integer data, round the forecasts to the closest integer.
	\item Clip all negative values at 0 (To allow for only positive values in the forecasts).
\end{enumerate}

\subsubsection{Performance Measures} 
\label{sec:performance_measures}

We measure the performance of the models in terms of a number of metrics. Symmetric Mean Absolute Percentage Error (SMAPE) is the most common performance measure used in many forecasting competitions. 

\begin{equation}
\label{eqn:smape}
    SMAPE = \frac{100\%}{H}\sum_{k=1}^{H} \frac{|F_{k} - Y_{k}|}{(|Y_{k}| + |F_{k}|)/2} 
\end{equation}

H, $F_{k}$ and $Y_{k}$ indicate the size of the horizon, the forecast of the NN, and the actual forecast, respectively. As seen in Equation \ref{eqn:smape}, the SMAPE is a metric based on percentage errors. However, according to \citet{Hyndman2006-oa}, SMAPE measure is susceptible to instability with values close to 0. The forecasts of the Wikipedia Web Traffic, NN5 and Tourism datasets are heavily affected by this since they all have 0 values. Therefore, to overcome this we use another variant of the SMAPE proposed by \citet{Suilin_undated-hh}. In this metric, the denominator of the above SMAPE is changed as follows.

\begin{equation}
    max(|Y_{k}| + |F_{k}| + \epsilon, 0.5 + \epsilon)    
\end{equation}

where the parameter $\epsilon$ is set to 0.1 following \citet{Suilin_undated-hh}. The idea is that the above metric can avoid division by values close to zero by switching to an alternate positive constant for the denominator in SMAPE when the forecasts are too small values. Yet, the SMAPE error metric has several other pitfalls such as its lack of interpretability and high skewness~\citep{Hyndman2006-oa}. Based on these issues, another metric is proposed by \citet{Hyndman2006-oa} known as the Mean Absolute Scaled Error (MASE) to address them. MASE is defined as follows. 

\begin{equation}
	MASE = \frac{\frac{1}{H}\sum_{k=1}^{H} |F_{k} - Y_{k}|}{\frac{1}{T - M}\sum_{k=M+1}^{T} |Y_{k} - Y_{k - M}|}\\
\end{equation}

MASE is a scale-independent measure, where the numerator is the same as in SMAPE, but normalized by the average in-sample one step na\"ive forecast error or the seasonal na\"ive forecast error in case of seasonal data. A value greater than 1 for MASE, indicates that the performance of the tested model is worse on average than the na\"ive benchmark and a value less than 1 denotes the opposite. Therefore, this error metric provides a direct indication of the performance of the model relative to the na\"ive benchmark.

The model evaluation of this study is presented using six metrics; Mean SMAPE, Median SMAPE, Mean MASE, Median MASE, Rank SMAPE and Rank MASE of the time series of each dataset. The rank measures compare the performance of every model against each other with respect to every time series in the dataset. Apart from them, when considering all the datasets together, ranks of models within each dataset with respect to both mean SMAPE and mean MASE metrics are plotted, since the original SMAPE and MASE values lie in very different ranges for the different datasets. These metrics are referred to as Mean SMAPE Ranks and Mean MASE Ranks respectively.

The literature introduces several other performance measures for forecasting such as the Geometric Mean Relative Absolute Error (GMRAE)~\citep{Hyndman2006-oa}. The GMRAE too is a relative error measure similar to MASE which measures the performance with respect to some benchmark method, but uses the geometric mean instead of the mean. However, as \citet{Chen2017-tm} state in their work, due to the usage of the geometric mean, values from the GMRAE can be quite small close to zero, especially in situations like ours where the expected forecasting horizon is very long such as in the NN5 and the Wikipedia Web Traffic datasets and the number of time series is large. 

\subsection{Benchmarks}

As for the benchmarks for comparison, we select two strong, well established traditional univariate techniques, \texttt{ets} and \texttt{auto.arima} with their default parameters from the \texttt{forecast}~\citep{Hyndman2008-gt} package in R. We perform experiments with the two techniques on all the aforementioned datasets and measure the performance in terms of the same metrics mentioned in Section \ref{sec:performance_measures}. Furthermore, we perform experiments with a pooled regression similar to the work of \citet{pooled_regression}. As mentioned in Section \ref{sec:cross_series_lit_review}, the pooled regression model differs from the RNN in that it does not maintain a state per every series and it models a linear relationship between the lags and the target variable. However, similar to the RNNs, the pooled regression models also calculate their weights globally by considering cross-series information, i.e., a pooled regression model works as a global AR model. Therefore, we use them in this study as a benchmark against RNNs to differentiate between the performance gains in RNNs purely due to building a global model versus due to the choice of the RNN architectures. We also train unpooled versions of regression models on all the datasets meaning that the models are built per every series similar to \texttt{ets} and \texttt{arima}. The unpooled regression models are used in this study to observe the accuracy gains from training regression models as global models rather than usual univariate models.

For implementing the regression models, we use the linear regression implementation from the \texttt{glmnet} package in the R programming language~\citep{Friedman2010glmnet}. Before feeding the series into the pooled regression models, we normalize them by dividing every series by its mean value. For the number of lags, we experiment with two options. The \texttt{auto.arima} model from the \texttt{forecast} package selects by default up to a maximum of 5 lags for both the AR and MA components combined. An ARMA model can be approximated by a pure AR model with a higher number of lags~\citep{robjhyndmangeorgeathanasopoulos2018}. Since the pooled regression model is a pure AR model, to be approximately compatible with the complexity of the \texttt{auto.arima} model, we use 10 lags (as opposed to 5 lags) as the first option for the number of lags in the pooled regression models. On the other hand, the moving window scheme as mentioned in Section \ref{sec:multiple_output_strategy}, already imposes a number of lags on the input window of the relevant RNN architectures. Therefore, we also build versions of the regression models with this number of lagged values. Furthermore, as mentioned in Section \ref{sec:model_overfitting}, the RNN models are regularized using a Ridge penalty. Hence, to enable a fair comparison, we develop the regression models both with and without the L2 regularization. To find the L2 regularization parameter we use two methods. The first is the built-in grid search technique in the \texttt{glmnet} package using a 10-fold cross-validation with a mean squared error as the validation error metric. However, to be more comparable with the RNNs, we also use the Bayesian optimization hyperparameter tuning along with the SMAPE as the validation error metric to find the L2 regularization parameter in the regression models. For this we use a simple 70\%-30\% split of training and validation sets respectively. For the pooled regression models, the range for the L2 regularization parameter is set to the 0-1 interval with 50 iterations in the tuning process. For the unpooled versions of the regression models this initial range is set to the 0-200 interval, since per one series there is a smaller amount of data available and the regularization parameter may need a larger value to avoid overfitting. For the Bayesian optimization we use the \texttt{rBayesianOptimization} package in the R programming language~\citep{Yan2016Bayesopt}. During testing, a recursive strategy is carried out to produce forecasts up to the forecasting horizon where one point is forecasted at every step, taking the last forecast as the most recent lag.

\subsection{Statistical Tests of the Results}
\label{sec:statistical_tests}
To evaluate the statistical significance of the differences of the performance of multiple techniques, we perform a non-parametric Friedman rank-sum test. This test determines whether the performance differences are statistically significant. To further explore these differences with respect to a control technique, specifically the best performing technique, we then perform Hochberg's post hoc procedure~\citep{Garcia2010-wc}. The significance level used is $\alpha = 0.05$. For those tests that have only two techniques to compare, we use a non-parametric paired Wilcoxon signed-rank test with Bonferroni correction, to measure the statistical significance of the differences. For this we apply the \texttt{wilcox.test} function from the \texttt{stats} package of the R core libraries~\citep{r_language}. The Bonferroni procedure is used to adjust the significance level used for comparison at each step, of a number of successive comparisons~\citep{Garcia2010-wc}. In particular, the procedure divides the significance level by the total number of comparisons for use in every individual comparison. The mean and the median of the SMAPE measure are used where relevant for all the statistical testing.

%% file: sections/results.tex
\section{Analysis of Results}
\label{sec:analysis_of_results}

This section provides a comprehensive analysis of the results obtained from the experiments. The datasets that we use for our experiments are quite different from each other in terms of the seasonality strength, number of time series, and length of individual time series. Thus, it seems reasonable to assume that the selected datasets cover a sufficient number of different time series characteristics to arrive at generalized conclusions. The results of all the implemented models in terms of the mean SMAPE metric are as shown in Table \ref{tab:mean_smape_results}. Results in terms of all the other error metrics are available in an Online Appendix\footnote{\url{https://drive.google.com/file/d/16rdzLTFwkKs-_eG_MU0i_rDnBQBCjsGe/view?usp=sharing}}. In the tables presenting the results, we use the abbreviations `NSTL', `MW' and `NMW' to denote `without using STL Decomposition', `moving window' and `non moving window' respectively. Furthermore, `IW+' indicates an increased input window size. 

\begin{filecontents*}{results/mean_SMAPE.csv}
Model Name,CIF,Kaggle,M3,NN5,Tourism,M4
auto.arima,11.7,47.96,14.25,25.91,19.74,13.08
ets,11.88,53.5,14.14,21.57,19.02,13.53
Pooled Regression Lags 10,14.67,122.43,14.73,30.21,31.29,14.18
Pooled Regression Lags 10 Bayesian Regularized,14.67,122.43,14.85,30.21,31.29,14.18
Pooled Regression Lags 10 Regularized,14.52,123.94,14.75,30.21,31.29,14.19
Pooled Regression Lags Window,12.89,47.47,14.36,22.41,21.1,13.75
Pooled Regression Lags Window Bayesian Regularized,12.89,47.55,14.42,22.41,21.1,13.75
Pooled Regression Lags Window Regularized,12.89,47.46,14.38,22.41,21.1,13.75
Unpooled Regression Lags 10,15.35,75.39,14.81,30.06,27.39,13.38
Unpooled Regression Lags 10 Bayesian Regularized,12.85,84.36,15.99,30.21,30.41,14.2
Unpooled Regression Lags 10 Regularized,12.74,94.28,15.32,30.06,27.61,14
Unpooled Regression Lags Window,14.34,56.75,14.93,22.63,20.71,13.46
Unpooled Regression Lags Window Bayesian Regularized,12.46,60.66,15.72,23.15,21.58,14.15
Unpooled Regression Lags Window Regularized,11.61,54.61,14.72,22.67,21.13,13.65
S2S GRU adagrad,11.57,51.77,15.91,28.33,20.57,13.53
S2S GRU adam,11.66,49.9,14.53,28.61,20.62,14.52
S2S GRU cocob,10.79,46.89,14.74,28.36,20.35,13.61
S2S LSTM adagrad,11.23,51.77,14.29,28.34,20.9,14.13
S2S LSTM adam,10.71,49.34,14.77,28.04,20.5,13.99
S2S LSTM cocob,11.04,52.46,14.6,24.77,20.51,14.02
S2S ERNN adagrad,10.47,51.45,14.7,28.99,20.44,14.24
S2S ERNN adam,11.19,50.7,14.66,28.32,20.63,13.86
S2S ERNN cocob,10.48,48.95,14.56,26.35,20.77,13.8
S2SD GRU MW adagrad,11.6,51.77,14.11,28.73,19.7,13.38
S2SD GRU MW adam,10.66,51.76,14.78,25.17,19.5,13.79
S2SD GRU MW cocob,10.3,48.8,14.38,27.92,19.67,13.57
S2SD GRU NMW adagrad,10.65,52.29,14.27,28.73,20.99,13.51
S2SD GRU NMW adam,10.41,46.9,14.35,25.42,20.5,13.92
S2SD GRU NMW cocob,10.48,47.33,14.35,24.8,20.8,13.75
S2SD LSTM MW adagrad,11.58,51.76,14.19,28.73,20.36,13.39
S2SD LSTM MW adam,10.28,51.09,14.68,26,20.18,13.76
S2SD LSTM MW cocob,11.23,51.77,14.45,24.29,19.76,13.59
S2SD LSTM NMW adagrad,10.81,52.28,14.39,28.73,20.22,13.5
S2SD LSTM NMW adam,10.34,47.37,14.43,25.36,20.33,13.62
S2SD LSTM NMW cocob,10.23,49.74,14.31,27.82,20.29,13.66
S2SD ERNN MW adagrad,11.34,51.58,14.2,28.58,19.58,13.42
S2SD ERNN MW adam,10.73,49.12,14.75,27.49,19.69,13.83
S2SD ERNN MW cocob,10.39,47.96,14.73,25.58,19.44,14.01
S2SD ERNN NMW adagrad,10.31,49.72,14.86,28.65,21.27,13.52
S2SD ERNN NMW adam,10.3,47.76,14.79,29.1,20.04,14.32
S2SD ERNN NMW cocob,10.28,47.94,14.94,26.37,19.68,13.57
Stacked GRU adagrad,10.54,47.23,14.64,21.84,21.56,14.15
Stacked GRU adam,10.55,47.11,14.76,21.93,21.92,14.29
Stacked GRU cocob,10.54,46.73,14.67,22.18,21.66,14.57
Stacked LSTM adagrad,10.51,47.12,14.34,22.12,20.61,13.97
Stacked LSTM adam,10.51,47.13,14.39,21.53,22.28,14.23
Stacked LSTM cocob,10.4,47.14,14.44,21.61,21.12,14.13
Stacked ERNN adagrad,10.53,47.79,14.73,23.83,21.24,14.3
Stacked ERNN adam,10.51,46.7,14.71,23.53,22.22,14.51
Stacked ERNN cocob,10.57,47.38,14.93,23.59,21.72,14.24
Stacked GRU adagrad(IW+),-,47.09,-,23.43,-,-
Stacked GRU adam(IW+),-,46.35,-,24.96,-,-
Stacked GRU cocob(IW+),-,46.46,-,21.92,-,-
Stacked LSTM adagrad(IW+),-,46.41,-,22.97,-,-
Stacked LSTM adam(IW+),-,46.52,-,21.59,-,-
Stacked LSTM cocob(IW+),-,46.54,-,21.54,-,-
Stacked ERNN adagrad(IW+),-,47.08,-,22.96,-,-
Stacked ERNN adam(IW+),-,46.76,-,22.5,-,-
Stacked ERNN cocob(IW+),-,47.59,-,23.13,-,-
NSTL S2S GRU adagrad,20.02,68.6,18.32,30.98,199.97,-
NSTL S2S GRU adam,16.98,50.48,18.52,28.09,48.2,-
NSTL S2S GRU cocob,17.96,50.44,17.56,26.01,95.19,-
NSTL S2S LSTM adagrad,17.54,71.25,17.75,29.52,41.41,-
NSTL S2S LSTM adam,16.11,47.8,17.37,36.33,32.69,-
NSTL S2S LSTM cocob,17.8,68.36,18.42,30.29,48.44,-
NSTL S2S ERNN adagrad,19.98,51.4,19.73,26.33,49.83,-
NSTL S2S ERNN adam,18.12,49.84,19.06,29.81,51.24,-
NSTL S2S ERNN cocob,17.97,51.69,18.85,31.77,43.68,-
NSTL S2SD GRU MW adagrad,16.12,51.08,20.87,24.45,199.97,19.47
NSTL S2SD GRU MW adam,13.74,51.2,18.09,24.08,36.02,-
NSTL S2SD GRU MW cocob,17.77,51.71,19.48,24.31,37.91,-
NSTL S2SD GRU NMW adagrad,14.42,46.57,15.8,25.43,199.97,-
NSTL S2SD GRU NMW adam,15.71,46.08,16.17,23.85,38.63,-
NSTL S2SD GRU NMW cocob,15.05,47.69,16.14,24.85,31.57,-
NSTL S2SD LSTM MW adagrad,27.06,51.24,21.29,40.96,40.27,-
NSTL S2SD LSTM MW adam,15.23,51.11,19.33,25.7,38.52,-
NSTL S2SD LSTM MW cocob,18.61,49.83,23.05,24.14,24.34,-
NSTL S2SD LSTM NMW adagrad,14.81,49.77,15.9,25.03,25.47,-
NSTL S2SD LSTM NMW adam,14.84,46.07,17.05,24.55,23.91,-
NSTL S2SD LSTM NMW cocob,22.74,47.72,16.35,25.31,31.52,-
NSTL S2SD ERNN MW adagrad,19.19,50.71,19.22,27.35,22.29,-
NSTL S2SD ERNN MW adam,13.92,49.69,18.55,24.52,30.79,-
NSTL S2SD ERNN MW cocob,14.88,50.82,16.42,24.87,26.16,-
NSTL S2SD ERNN NMW adagrad,15.78,47.41,17.22,27.47,38.19,-
NSTL S2SD ERNN NMW adam,15.02,46.37,16.34,22.81,28.55,-
NSTL S2SD ERNN NMW cocob,14.24,48.27,15.26,23.22,25.82,-
NSTL Stacked GRU adagrad,16.77,45.68,16.39,24.39,199.97,-
NSTL Stacked GRU adam,16.34,45.62,16.76,23.47,38.66,-
NSTL Stacked GRU cocob,17.64,46.35,16.23,23.55,41.52,-
NSTL Stacked LSTM adagrad,17.08,45.88,16.69,25.1,23.46,-
NSTL Stacked LSTM adam,15.83,46.12,17.54,24.44,21.73,-
NSTL Stacked LSTM cocob,15.27,45.9,17.97,26.28,25.24,17.71
NSTL Stacked ERNN adagrad,13.25,46,16.78,24.22,25.36,-
NSTL Stacked ERNN adam,16.05,45.77,16.43,23.04,20.67,-
NSTL Stacked ERNN cocob,15.74,46.02,17.32,23.55,28.71,-
NSTL Stacked GRU adagrad(IW+),-,45.9,-,23.53,-,-
NSTL Stacked GRU adam(IW+),-,45.69,-,22.2,-,-
NSTL Stacked GRU cocob(IW+),-,46.01,-,22.78,-,-
NSTL Stacked LSTM adagrad(IW+),-,45.68,-,23.15,-,-
NSTL Stacked LSTM adam(IW+),-,46.32,-,22.45,-,-
NSTL Stacked LSTM cocob(IW+),-,46.12,-,22.66,-,-
NSTL Stacked ERNN adagrad(IW+),-,45.69,-,22.79,-,-
NSTL Stacked ERNN adam(IW+),-,45.92,-,22.13,-,-
NSTL Stacked ERNN cocob(IW+),-,45.79,-,22.8,-,-
\end{filecontents*}
\pgfplotstabletypeset[
font=\footnotesize,
begin table=\begin{longtable},
	multicolumn names, 
	col sep=comma, 
	display columns/0/.style={
		column type={l},string type},  
	display columns/1/.style={
		column type={c},string type},
	display columns/2/.style={
		column type={c},string type},
	display columns/3/.style={
		column type={c},string type},
	display columns/4/.style={
		column type={c},string type},
	display columns/5/.style={
		column type={c},string type},
	display columns/6/.style={
		column type={c},string type},
	every head row/.style={
		before row={\toprule},
		after row={\toprule} 
	},
	every row 34 column 1/.style={postproc cell content/.style={@cell content=\textbf{##1}}},
	every row 23 column 3/.style={postproc cell content/.style={@cell content=\textbf{##1}}},
	every row 1 column 5/.style={postproc cell content/.style={@cell content=\textbf{##1}}},
	every row 0 column 6/.style={postproc cell content/.style={@cell content=\textbf{##1}}},
	every row 10 column 2/.style={postproc cell content/.style={@cell content={*$##1^\dagger$}}},
	every row 101 column 2/.style={postproc cell content/.style={@cell content={**$##1^\ddagger$}}},
	every row 102 column 2/.style={postproc cell content/.style={@cell content={**$##1^\ddagger$}}},
	every row 103 column 2/.style={postproc cell content/.style={@cell content={**$##1^\ddagger$}}},
	every row 11 column 2/.style={postproc cell content/.style={@cell content={*$##1^\dagger$}}},
	every row 12 column 2/.style={postproc cell content/.style={@cell content={*$##1^\dagger$}}},
	every row 18 column 2/.style={postproc cell content/.style={@cell content={**$##1^\ddagger$}}},
	every row 2 column 2/.style={postproc cell content/.style={@cell content={*$##1^\dagger$}}},
	every row 22 column 2/.style={postproc cell content/.style={@cell content={**$##1^\dagger$}}},
	every row 26 column 2/.style={postproc cell content/.style={@cell content={*$##1^\dagger$}}},
	every row 3 column 2/.style={postproc cell content/.style={@cell content={*$##1^\dagger$}}},
	every row 32 column 2/.style={postproc cell content/.style={@cell content={*$##1^\dagger$}}},
	every row 37 column 2/.style={postproc cell content/.style={@cell content={**$##1^\dagger$}}},
	every row 39 column 2/.style={postproc cell content/.style={@cell content={**$##1^\dagger$}}},
	every row 4 column 2/.style={postproc cell content/.style={@cell content={*$##1^\dagger$}}},
	every row 40 column 2/.style={postproc cell content/.style={@cell content={**$##1^\dagger$}}},
	every row 59 column 2/.style={postproc cell content/.style={@cell content={*$##1^\dagger$}}},
	every row 62 column 2/.style={postproc cell content/.style={@cell content={*$##1^\dagger$}}},
	every row 63 column 2/.style={postproc cell content/.style={@cell content={**$##1^\dagger$}}},
	every row 64 column 2/.style={postproc cell content/.style={@cell content={*$##1^\dagger$}}},
	every row 66 column 2/.style={postproc cell content/.style={@cell content={**$##1^\dagger$}}},
	every row 73 column 2/.style={postproc cell content/.style={@cell content={**$##1^\dagger$}}},
	every row 79 column 2/.style={postproc cell content/.style={@cell content={**$##1^\dagger$}}},
	every row 8 column 2/.style={postproc cell content/.style={@cell content={*$##1^\dagger$}}},
	every row 86 column 2/.style={postproc cell content/.style={@cell content={**$##1^\ddagger$}}},
	every row 87 column 2/.style={postproc cell content/.style={@cell content={**$\textbf{##1}^\ddagger$}}},
	every row 89 column 2/.style={postproc cell content/.style={@cell content={**$##1^\ddagger$}}},
	every row 9 column 2/.style={postproc cell content/.style={@cell content={*$##1^\dagger$}}},
	every row 93 column 2/.style={postproc cell content/.style={@cell content={**$##1^\ddagger$}}},
	every row 96 column 2/.style={postproc cell content/.style={@cell content={**$##1^\ddagger$}}},
	every row 98 column 2/.style={postproc cell content/.style={@cell content={**$##1^\ddagger$}}},
	every row 99 column 2/.style={postproc cell content/.style={@cell content={**$##1^\ddagger$}}},
	every row 5 column 2/.style={postproc cell content/.style={@cell content={**$##1$}}},
	every row 7 column 2/.style={postproc cell content/.style={@cell content={**$##1$}}},
	every row 33 column 2/.style={postproc cell content/.style={@cell content={**$##1$}}},
	every row 43 column 2/.style={postproc cell content/.style={@cell content={**$##1$}}},
	every row 48 column 2/.style={postproc cell content/.style={@cell content={**$##1$}}},
	every row 94 column 2/.style={postproc cell content/.style={@cell content={**$##1$}}},
	every row 49 column 2/.style={postproc cell content/.style={@cell content={**$##1$}}},
	every row 95 column 2/.style={postproc cell content/.style={@cell content={**$##1$}}},
	every row 97 column 2/.style={postproc cell content/.style={@cell content={**$##1$}}},
	every row 50 column 2/.style={postproc cell content/.style={@cell content={**$##1$}}},
	every row 51 column 2/.style={postproc cell content/.style={@cell content={**$##1$}}},
	every row 52 column 2/.style={postproc cell content/.style={@cell content={**$##1$}}},
	every row 41 column 2/.style={postproc cell content/.style={@cell content={**$##1$}}},
	every row 78 column 2/.style={postproc cell content/.style={@cell content={**$##1$}}},
	every row 71 column 2/.style={postproc cell content/.style={@cell content={**$##1$}}},
	every row 72 column 2/.style={postproc cell content/.style={@cell content={**$##1$}}},
	every row 42 column 2/.style={postproc cell content/.style={@cell content={**$##1$}}},
	every row 16 column 2/.style={postproc cell content/.style={@cell content={**$##1$}}},
	every row 54 column 2/.style={postproc cell content/.style={@cell content={**$##1$}}},
	every row 55 column 2/.style={postproc cell content/.style={@cell content={**$##1$}}},
	every row 56 column 2/.style={postproc cell content/.style={@cell content={**$##1$}}},
	every row 57 column 2/.style={postproc cell content/.style={@cell content={**$##1$}}},
	every row 58 column 2/.style={postproc cell content/.style={@cell content={**$##1$}}},
	every row 88 column 2/.style={postproc cell content/.style={@cell content={**$##1$}}},
	every row 90 column 2/.style={postproc cell content/.style={@cell content={**$##1$}}},
	every row 91 column 2/.style={postproc cell content/.style={@cell content={**$##1$}}},
	every row 92 column 2/.style={postproc cell content/.style={@cell content={**$##1$}}},
	every row 83 column 2/.style={postproc cell content/.style={@cell content={**$##1$}}},
	every row 84 column 2/.style={postproc cell content/.style={@cell content={**$##1$}}},
	every row 27 column 2/.style={postproc cell content/.style={@cell content={**$##1$}}},
	every row 44 column 2/.style={postproc cell content/.style={@cell content={**$##1$}}},
	every row 45 column 2/.style={postproc cell content/.style={@cell content={**$##1$}}},
	every row 46 column 2/.style={postproc cell content/.style={@cell content={**$##1$}}},
	every row 47 column 2/.style={postproc cell content/.style={@cell content={**$##1$}}},
	every row 53 column 2/.style={postproc cell content/.style={@cell content={**$##1$}}},
	every row 28 column 2/.style={postproc cell content/.style={@cell content={**$##1$}}},
	every row 100 column 2/.style={postproc cell content/.style={@cell content={**$##1$}}},
	every row 6 column 2/.style={postproc cell content/.style={@cell content={**$##1$}}},
	every row 19 column 2/.style={postproc cell content/.style={@cell content={$##1\dagger$}}},
	every row 30 column 2/.style={postproc cell content/.style={@cell content={$##1\dagger$}}},
	every row 31 column 2/.style={postproc cell content/.style={@cell content={$##1\dagger$}}},
	every row 38 column 2/.style={postproc cell content/.style={@cell content={$##1\dagger$}}},
	every row 60 column 2/.style={postproc cell content/.style={@cell content={$##1\dagger$}}},
	every row 20 column 2/.style={postproc cell content/.style={@cell content={$##1\dagger$}}},
	every row 61 column 2/.style={postproc cell content/.style={@cell content={$##1\dagger$}}},
	every row 65 column 2/.style={postproc cell content/.style={@cell content={$##1\dagger$}}},
	every row 25 column 2/.style={postproc cell content/.style={@cell content={$##1\dagger$}}},
	every row 68 column 2/.style={postproc cell content/.style={@cell content={$##1\dagger$}}},
	every row 21 column 2/.style={postproc cell content/.style={@cell content={$##1\dagger$}}},
	every row 13 column 2/.style={postproc cell content/.style={@cell content={$##1\dagger$}}},
	every row 34 column 2/.style={postproc cell content/.style={@cell content={$##1\dagger$}}},
	every row 35 column 2/.style={postproc cell content/.style={@cell content={$##1\dagger$}}},
	every row 36 column 2/.style={postproc cell content/.style={@cell content={$##1\dagger$}}},
	every row 75 column 2/.style={postproc cell content/.style={@cell content={$##1\dagger$}}},
	every row 76 column 2/.style={postproc cell content/.style={@cell content={$##1\dagger$}}},
	every row 77 column 2/.style={postproc cell content/.style={@cell content={$##1\dagger$}}},
	every row 80 column 2/.style={postproc cell content/.style={@cell content={$##1\dagger$}}},
	every row 69 column 2/.style={postproc cell content/.style={@cell content={$##1\dagger$}}},
	every row 70 column 2/.style={postproc cell content/.style={@cell content={$##1\dagger$}}},
	every row 14 column 2/.style={postproc cell content/.style={@cell content={$##1\dagger$}}},
	every row 15 column 2/.style={postproc cell content/.style={@cell content={$##1\dagger$}}},
	every row 17 column 2/.style={postproc cell content/.style={@cell content={$##1\dagger$}}},
	every row 82 column 2/.style={postproc cell content/.style={@cell content={$##1\dagger$}}},
	every row 85 column 2/.style={postproc cell content/.style={@cell content={$##1\dagger$}}},
	every row 74 column 2/.style={postproc cell content/.style={@cell content={$##1\dagger$}}},
	every row 81 column 2/.style={postproc cell content/.style={@cell content={$##1\dagger$}}},
	every row 67 column 2/.style={postproc cell content/.style={@cell content={$##1\dagger$}}},
	every row 23 column 2/.style={postproc cell content/.style={@cell content={$##1\dagger$}}},
	every row 24 column 2/.style={postproc cell content/.style={@cell content={$##1\dagger$}}},
	every row 29 column 2/.style={postproc cell content/.style={@cell content={$##1\dagger$}}},
	every row 10 column 4/.style={postproc cell content/.style={@cell content={*$##1^\dagger$}}},
	every row 2 column 4/.style={postproc cell content/.style={@cell content={*$##1^\dagger$}}},
	every row 3 column 4/.style={postproc cell content/.style={@cell content={*$##1^\dagger$}}},
	every row 4 column 4/.style={postproc cell content/.style={@cell content={*$##1^\dagger$}}},
	every row 59 column 4/.style={postproc cell content/.style={@cell content={*$##1^\dagger$}}},
	every row 60 column 4/.style={postproc cell content/.style={@cell content={*$##1^\dagger$}}},
	every row 62 column 4/.style={postproc cell content/.style={@cell content={*$##1^\dagger$}}},
	every row 63 column 4/.style={postproc cell content/.style={@cell content={*$##1^\dagger$}}},
	every row 64 column 4/.style={postproc cell content/.style={@cell content={*$##1^\dagger$}}},
	every row 65 column 4/.style={postproc cell content/.style={@cell content={*$##1^\dagger$}}},
	every row 66 column 4/.style={postproc cell content/.style={@cell content={*$##1^\dagger$}}},
	every row 67 column 4/.style={postproc cell content/.style={@cell content={*$##1^\dagger$}}},
	every row 74 column 4/.style={postproc cell content/.style={@cell content={*$##1^\dagger$}}},
	every row 8 column 4/.style={postproc cell content/.style={@cell content={*$##1^\dagger$}}},
	every row 80 column 4/.style={postproc cell content/.style={@cell content={*$##1^\dagger$}}},
	every row 83 column 4/.style={postproc cell content/.style={@cell content={*$##1^\dagger$}}},
	every row 9 column 4/.style={postproc cell content/.style={@cell content={*$##1^\dagger$}}},
	every row 91 column 4/.style={postproc cell content/.style={@cell content={*$##1^\dagger$}}},
	every row 71 column 4/.style={postproc cell content/.style={@cell content={*$##1$}}},
	every row 103 column 4/.style={postproc cell content/.style={@cell content={*$##1$}}},
	every row 68 column 4/.style={postproc cell content/.style={@cell content={*$##1$}}},
	every row 69 column 4/.style={postproc cell content/.style={@cell content={*$##1$}}},
	every row 70 column 4/.style={postproc cell content/.style={@cell content={*$##1$}}},
	every row 72 column 4/.style={postproc cell content/.style={@cell content={*$##1$}}},
	every row 73 column 4/.style={postproc cell content/.style={@cell content={*$##1$}}},
	every row 75 column 4/.style={postproc cell content/.style={@cell content={*$##1$}}},
	every row 76 column 4/.style={postproc cell content/.style={@cell content={*$##1$}}},
	every row 77 column 4/.style={postproc cell content/.style={@cell content={*$##1$}}},
	every row 78 column 4/.style={postproc cell content/.style={@cell content={*$##1$}}},
	every row 79 column 4/.style={postproc cell content/.style={@cell content={*$##1$}}},
	every row 61 column 4/.style={postproc cell content/.style={@cell content={*$##1$}}},
	every row 88 column 4/.style={postproc cell content/.style={@cell content={*$##1$}}},
	every row 30 column 4/.style={postproc cell content/.style={@cell content={*$##1$}}},
	every row 31 column 4/.style={postproc cell content/.style={@cell content={*$##1$}}},
	every row 34 column 4/.style={postproc cell content/.style={@cell content={*$##1$}}},
	every row 86 column 4/.style={postproc cell content/.style={@cell content={*$##1$}}},
	every row 87 column 4/.style={postproc cell content/.style={@cell content={*$##1$}}},
	every row 89 column 4/.style={postproc cell content/.style={@cell content={*$##1$}}},
	every row 90 column 4/.style={postproc cell content/.style={@cell content={*$##1$}}},
	every row 92 column 4/.style={postproc cell content/.style={@cell content={*$##1$}}},
	every row 93 column 4/.style={postproc cell content/.style={@cell content={*$##1$}}},
	every row 94 column 4/.style={postproc cell content/.style={@cell content={*$##1$}}},
	every row 18 column 4/.style={postproc cell content/.style={@cell content={*$##1$}}},
	every row 20 column 4/.style={postproc cell content/.style={@cell content={*$##1$}}},
	every row 21 column 4/.style={postproc cell content/.style={@cell content={*$##1$}}},
	every row 22 column 4/.style={postproc cell content/.style={@cell content={*$##1$}}},
	every row 6 column 4/.style={postproc cell content/.style={@cell content={*$##1$}}},
	every row 5 column 4/.style={postproc cell content/.style={@cell content={*$##1$}}},
	every row 7 column 4/.style={postproc cell content/.style={@cell content={*$##1$}}},
	every row 36 column 4/.style={postproc cell content/.style={@cell content={*$##1$}}},
	every row 98 column 4/.style={postproc cell content/.style={@cell content={*$##1$}}},
	every row 81 column 4/.style={postproc cell content/.style={@cell content={*$##1$}}},
	every row 82 column 4/.style={postproc cell content/.style={@cell content={*$##1$}}},
	every row 84 column 4/.style={postproc cell content/.style={@cell content={*$##1$}}},
	every row 85 column 4/.style={postproc cell content/.style={@cell content={*$##1$}}},
	every row 37 column 4/.style={postproc cell content/.style={@cell content={*$##1$}}},
	every row 39 column 4/.style={postproc cell content/.style={@cell content={*$##1$}}},
	every row 40 column 4/.style={postproc cell content/.style={@cell content={*$##1$}}},
	every row 15 column 4/.style={postproc cell content/.style={@cell content={*$##1$}}},
	every row 101 column 4/.style={postproc cell content/.style={@cell content={*$##1$}}},
	every row 12 column 4/.style={postproc cell content/.style={@cell content={*$##1$}}},
	every row 11 column 4/.style={postproc cell content/.style={@cell content={*$##1$}}},
	every row 13 column 4/.style={postproc cell content/.style={@cell content={*$##1$}}},
	every row 95 column 4/.style={postproc cell content/.style={@cell content={*$##1$}}},
	every row 97 column 4/.style={postproc cell content/.style={@cell content={*$##1$}}},
	every row 24 column 4/.style={postproc cell content/.style={@cell content={*$##1$}}},
	every row 25 column 4/.style={postproc cell content/.style={@cell content={*$##1$}}},
	every row 27 column 4/.style={postproc cell content/.style={@cell content={*$##1$}}},
	every row 57 column 4/.style={postproc cell content/.style={@cell content={$##1\ddagger$}}},
	every row 52 column 4/.style={postproc cell content/.style={@cell content={$##1\ddagger$}}},
	every row 42 column 4/.style={postproc cell content/.style={@cell content={$##1\ddagger$}}},
	every row 43 column 4/.style={postproc cell content/.style={@cell content={$##1\ddagger$}}},
	every row 44 column 4/.style={postproc cell content/.style={@cell content={$##1\ddagger$}}},
	every row 45 column 4/.style={postproc cell content/.style={@cell content={$\textbf{##1}\ddagger$}}},
	every row 46 column 4/.style={postproc cell content/.style={@cell content={$##1\ddagger$}}},
	every row 47 column 4/.style={postproc cell content/.style={@cell content={$##1\ddagger$}}},
	every row 54 column 4/.style={postproc cell content/.style={@cell content={$##1\ddagger$}}},
	every row 55 column 4/.style={postproc cell content/.style={@cell content={$##1\ddagger$}}},
	every row 41 column 4/.style={postproc cell content/.style={@cell content={$##1\ddagger$}}},
	every row 48 column 4/.style={postproc cell content/.style={@cell content={$##1\ddagger$}}},
	every row 49 column 4/.style={postproc cell content/.style={@cell content={$##1\ddagger$}}},
	every row 10 column 5/.style={postproc cell content/.style={@cell content={*$##1^\dagger$}}},
	every row 11 column 5/.style={postproc cell content/.style={@cell content={*$##1^\dagger$}}},
	every row 12 column 5/.style={postproc cell content/.style={@cell content={*$##1^\dagger$}}},
	every row 13 column 5/.style={postproc cell content/.style={@cell content={*$##1^\dagger$}}},
	every row 14 column 5/.style={postproc cell content/.style={@cell content={*$##1^\dagger$}}},
	every row 15 column 5/.style={postproc cell content/.style={@cell content={*$##1^\dagger$}}},
	every row 16 column 5/.style={postproc cell content/.style={@cell content={*$##1^\dagger$}}},
	every row 17 column 5/.style={postproc cell content/.style={@cell content={*$##1^\dagger$}}},
	every row 18 column 5/.style={postproc cell content/.style={@cell content={*$##1^\dagger$}}},
	every row 19 column 5/.style={postproc cell content/.style={@cell content={*$##1^\dagger$}}},
	every row 2 column 5/.style={postproc cell content/.style={@cell content={*$##1^\dagger$}}},
	every row 20 column 5/.style={postproc cell content/.style={@cell content={*$##1^\dagger$}}},
	every row 21 column 5/.style={postproc cell content/.style={@cell content={*$##1^\dagger$}}},
	every row 22 column 5/.style={postproc cell content/.style={@cell content={*$##1^\dagger$}}},
	every row 26 column 5/.style={postproc cell content/.style={@cell content={*$##1^\dagger$}}},
	every row 27 column 5/.style={postproc cell content/.style={@cell content={*$##1^\dagger$}}},
	every row 28 column 5/.style={postproc cell content/.style={@cell content={*$##1^\dagger$}}},
	every row 29 column 5/.style={postproc cell content/.style={@cell content={*$##1^\dagger$}}},
	every row 3 column 5/.style={postproc cell content/.style={@cell content={*$##1^\dagger$}}},
	every row 32 column 5/.style={postproc cell content/.style={@cell content={*$##1^\dagger$}}},
	every row 33 column 5/.style={postproc cell content/.style={@cell content={*$##1^\dagger$}}},
	every row 34 column 5/.style={postproc cell content/.style={@cell content={*$##1^\dagger$}}},
	every row 38 column 5/.style={postproc cell content/.style={@cell content={*$##1^\dagger$}}},
	every row 4 column 5/.style={postproc cell content/.style={@cell content={*$##1^\dagger$}}},
	every row 41 column 5/.style={postproc cell content/.style={@cell content={*$##1^\dagger$}}},
	every row 42 column 5/.style={postproc cell content/.style={@cell content={*$##1^\dagger$}}},
	every row 43 column 5/.style={postproc cell content/.style={@cell content={*$##1^\dagger$}}},
	every row 44 column 5/.style={postproc cell content/.style={@cell content={*$##1^\dagger$}}},
	every row 45 column 5/.style={postproc cell content/.style={@cell content={*$##1^\dagger$}}},
	every row 46 column 5/.style={postproc cell content/.style={@cell content={*$##1^\dagger$}}},
	every row 48 column 5/.style={postproc cell content/.style={@cell content={*$##1^\dagger$}}},
	every row 49 column 5/.style={postproc cell content/.style={@cell content={*$##1^\dagger$}}},
	every row 5 column 5/.style={postproc cell content/.style={@cell content={*$##1^\dagger$}}},
	every row 59 column 5/.style={postproc cell content/.style={@cell content={*$##1^\dagger$}}},
	every row 6 column 5/.style={postproc cell content/.style={@cell content={*$##1^\dagger$}}},
	every row 60 column 5/.style={postproc cell content/.style={@cell content={*$##1^\dagger$}}},
	every row 61 column 5/.style={postproc cell content/.style={@cell content={*$##1^\dagger$}}},
	every row 62 column 5/.style={postproc cell content/.style={@cell content={*$##1^\dagger$}}},
	every row 63 column 5/.style={postproc cell content/.style={@cell content={*$##1^\dagger$}}},
	every row 64 column 5/.style={postproc cell content/.style={@cell content={*$##1^\dagger$}}},
	every row 65 column 5/.style={postproc cell content/.style={@cell content={*$##1^\dagger$}}},
	every row 66 column 5/.style={postproc cell content/.style={@cell content={*$##1^\dagger$}}},
	every row 67 column 5/.style={postproc cell content/.style={@cell content={*$##1^\dagger$}}},
	every row 68 column 5/.style={postproc cell content/.style={@cell content={*$##1^\dagger$}}},
	every row 69 column 5/.style={postproc cell content/.style={@cell content={*$##1^\dagger$}}},
	every row 7 column 5/.style={postproc cell content/.style={@cell content={*$##1^\dagger$}}},
	every row 70 column 5/.style={postproc cell content/.style={@cell content={*$##1^\dagger$}}},
	every row 71 column 5/.style={postproc cell content/.style={@cell content={*$##1^\dagger$}}},
	every row 72 column 5/.style={postproc cell content/.style={@cell content={*$##1^\dagger$}}},
	every row 73 column 5/.style={postproc cell content/.style={@cell content={*$##1^\dagger$}}},
	every row 74 column 5/.style={postproc cell content/.style={@cell content={*$##1^\dagger$}}},
	every row 75 column 5/.style={postproc cell content/.style={@cell content={*$##1^\dagger$}}},
	every row 76 column 5/.style={postproc cell content/.style={@cell content={*$##1^\dagger$}}},
	every row 77 column 5/.style={postproc cell content/.style={@cell content={*$##1^\dagger$}}},
	every row 78 column 5/.style={postproc cell content/.style={@cell content={*$##1^\dagger$}}},
	every row 79 column 5/.style={postproc cell content/.style={@cell content={*$##1^\dagger$}}},
	every row 8 column 5/.style={postproc cell content/.style={@cell content={*$##1^\dagger$}}},
	every row 80 column 5/.style={postproc cell content/.style={@cell content={*$##1^\dagger$}}},
	every row 81 column 5/.style={postproc cell content/.style={@cell content={*$##1^\dagger$}}},
	every row 82 column 5/.style={postproc cell content/.style={@cell content={*$##1^\dagger$}}},
	every row 83 column 5/.style={postproc cell content/.style={@cell content={*$##1^\dagger$}}},
	every row 84 column 5/.style={postproc cell content/.style={@cell content={*$##1^\dagger$}}},
	every row 85 column 5/.style={postproc cell content/.style={@cell content={*$##1^\dagger$}}},
	every row 86 column 5/.style={postproc cell content/.style={@cell content={*$##1^\dagger$}}},
	every row 87 column 5/.style={postproc cell content/.style={@cell content={*$##1^\dagger$}}},
	every row 88 column 5/.style={postproc cell content/.style={@cell content={*$##1^\dagger$}}},
	every row 89 column 5/.style={postproc cell content/.style={@cell content={*$##1^\dagger$}}},
	every row 9 column 5/.style={postproc cell content/.style={@cell content={*$##1^\dagger$}}},
	every row 90 column 5/.style={postproc cell content/.style={@cell content={*$##1^\dagger$}}},
	every row 91 column 5/.style={postproc cell content/.style={@cell content={*$##1^\dagger$}}},
	every row 92 column 5/.style={postproc cell content/.style={@cell content={*$##1^\dagger$}}},
	every row 94 column 5/.style={postproc cell content/.style={@cell content={*$##1^\dagger$}}},
	every row 47 column 5/.style={postproc cell content/.style={@cell content={*$##1$}}},
	every row 30 column 5/.style={postproc cell content/.style={@cell content={*$##1$}}},
	every row 93 column 5/.style={postproc cell content/.style={@cell content={*$##1$}}},
	every row 59 column 1/.style={postproc cell content/.style={@cell content={*$##1^\dagger$}}},
	every row 60 column 1/.style={postproc cell content/.style={@cell content={*$##1^\dagger$}}},
	every row 61 column 1/.style={postproc cell content/.style={@cell content={*$##1^\dagger$}}},
	every row 62 column 1/.style={postproc cell content/.style={@cell content={*$##1^\dagger$}}},
	every row 63 column 1/.style={postproc cell content/.style={@cell content={*$##1^\dagger$}}},
	every row 64 column 1/.style={postproc cell content/.style={@cell content={*$##1^\dagger$}}},
	every row 65 column 1/.style={postproc cell content/.style={@cell content={*$##1^\dagger$}}},
	every row 66 column 1/.style={postproc cell content/.style={@cell content={*$##1^\dagger$}}},
	every row 67 column 1/.style={postproc cell content/.style={@cell content={*$##1^\dagger$}}},
	every row 68 column 1/.style={postproc cell content/.style={@cell content={*$##1^\dagger$}}},
	every row 70 column 1/.style={postproc cell content/.style={@cell content={*$##1^\dagger$}}},
	every row 72 column 1/.style={postproc cell content/.style={@cell content={*$##1^\dagger$}}},
	every row 73 column 1/.style={postproc cell content/.style={@cell content={*$##1^\dagger$}}},
	every row 74 column 1/.style={postproc cell content/.style={@cell content={*$##1^\dagger$}}},
	every row 76 column 1/.style={postproc cell content/.style={@cell content={*$##1^\dagger$}}},
	every row 77 column 1/.style={postproc cell content/.style={@cell content={*$##1^\dagger$}}},
	every row 79 column 1/.style={postproc cell content/.style={@cell content={*$##1^\dagger$}}},
	every row 80 column 1/.style={postproc cell content/.style={@cell content={*$##1^\dagger$}}},
	every row 82 column 1/.style={postproc cell content/.style={@cell content={*$##1^\dagger$}}},
	every row 83 column 1/.style={postproc cell content/.style={@cell content={*$##1^\dagger$}}},
	every row 86 column 1/.style={postproc cell content/.style={@cell content={*$##1^\dagger$}}},
	every row 87 column 1/.style={postproc cell content/.style={@cell content={*$##1^\dagger$}}},
	every row 88 column 1/.style={postproc cell content/.style={@cell content={*$##1^\dagger$}}},
	every row 89 column 1/.style={postproc cell content/.style={@cell content={*$##1^\dagger$}}},
	every row 90 column 1/.style={postproc cell content/.style={@cell content={*$##1^\dagger$}}},
	every row 91 column 1/.style={postproc cell content/.style={@cell content={*$##1^\dagger$}}},
	every row 93 column 1/.style={postproc cell content/.style={@cell content={*$##1^\dagger$}}},
	every row 94 column 1/.style={postproc cell content/.style={@cell content={*$##1^\dagger$}}},
	every row 10 column 1/.style={postproc cell content/.style={@cell content={*$##1$}}},
	every row 71 column 1/.style={postproc cell content/.style={@cell content={*$##1$}}},
	every row 75 column 1/.style={postproc cell content/.style={@cell content={*$##1$}}},
	every row 78 column 1/.style={postproc cell content/.style={@cell content={*$##1$}}},
	every row 81 column 1/.style={postproc cell content/.style={@cell content={*$##1$}}},
	every row 84 column 1/.style={postproc cell content/.style={@cell content={*$##1$}}},
	every row 85 column 1/.style={postproc cell content/.style={@cell content={*$##1$}}},
	every row 69 column 1/.style={postproc cell content/.style={@cell content={*$##1$}}},
	every row 92 column 1/.style={postproc cell content/.style={@cell content={*$##1$}}},
	every row 3 column 1/.style={postproc cell content/.style={@cell content={*$##1$}}},
	every row 2 column 1/.style={postproc cell content/.style={@cell content={*$##1$}}},
	every row 4 column 1/.style={postproc cell content/.style={@cell content={*$##1$}}},
	every row 6 column 1/.style={postproc cell content/.style={@cell content={*$##1$}}},
	every row 5 column 1/.style={postproc cell content/.style={@cell content={*$##1$}}},
	every row 7 column 1/.style={postproc cell content/.style={@cell content={*$##1$}}},
	every row 12 column 3/.style={postproc cell content/.style={@cell content={*$##1^\dagger$}}},
	every row 14 column 3/.style={postproc cell content/.style={@cell content={*$##1^\dagger$}}},
	every row 16 column 3/.style={postproc cell content/.style={@cell content={*$##1^\dagger$}}},
	every row 2 column 3/.style={postproc cell content/.style={@cell content={*$##1^\dagger$}}},
	every row 20 column 3/.style={postproc cell content/.style={@cell content={*$##1^\dagger$}}},
	every row 21 column 3/.style={postproc cell content/.style={@cell content={*$##1^\dagger$}}},
	every row 22 column 3/.style={postproc cell content/.style={@cell content={*$##1^\dagger$}}},
	every row 26 column 3/.style={postproc cell content/.style={@cell content={*$##1^\dagger$}}},
	every row 3 column 3/.style={postproc cell content/.style={@cell content={*$##1^\dagger$}}},
	every row 32 column 3/.style={postproc cell content/.style={@cell content={*$##1^\dagger$}}},
	every row 38 column 3/.style={postproc cell content/.style={@cell content={*$##1^\dagger$}}},
	every row 39 column 3/.style={postproc cell content/.style={@cell content={*$##1^\dagger$}}},
	every row 4 column 3/.style={postproc cell content/.style={@cell content={*$##1^\dagger$}}},
	every row 40 column 3/.style={postproc cell content/.style={@cell content={*$##1^\dagger$}}},
	every row 5 column 3/.style={postproc cell content/.style={@cell content={*$##1^\dagger$}}},
	every row 59 column 3/.style={postproc cell content/.style={@cell content={*$##1^\dagger$}}},
	every row 6 column 3/.style={postproc cell content/.style={@cell content={*$##1^\dagger$}}},
	every row 60 column 3/.style={postproc cell content/.style={@cell content={*$##1^\dagger$}}},
	every row 61 column 3/.style={postproc cell content/.style={@cell content={*$##1^\dagger$}}},
	every row 62 column 3/.style={postproc cell content/.style={@cell content={*$##1^\dagger$}}},
	every row 63 column 3/.style={postproc cell content/.style={@cell content={*$##1^\dagger$}}},
	every row 64 column 3/.style={postproc cell content/.style={@cell content={*$##1^\dagger$}}},
	every row 65 column 3/.style={postproc cell content/.style={@cell content={*$##1^\dagger$}}},
	every row 66 column 3/.style={postproc cell content/.style={@cell content={*$##1^\dagger$}}},
	every row 67 column 3/.style={postproc cell content/.style={@cell content={*$##1^\dagger$}}},
	every row 68 column 3/.style={postproc cell content/.style={@cell content={*$##1^\dagger$}}},
	every row 69 column 3/.style={postproc cell content/.style={@cell content={*$##1^\dagger$}}},
	every row 7 column 3/.style={postproc cell content/.style={@cell content={*$##1^\dagger$}}},
	every row 70 column 3/.style={postproc cell content/.style={@cell content={*$##1^\dagger$}}},
	every row 71 column 3/.style={postproc cell content/.style={@cell content={*$##1^\dagger$}}},
	every row 72 column 3/.style={postproc cell content/.style={@cell content={*$##1^\dagger$}}},
	every row 73 column 3/.style={postproc cell content/.style={@cell content={*$##1^\dagger$}}},
	every row 74 column 3/.style={postproc cell content/.style={@cell content={*$##1^\dagger$}}},
	every row 75 column 3/.style={postproc cell content/.style={@cell content={*$##1^\dagger$}}},
	every row 76 column 3/.style={postproc cell content/.style={@cell content={*$##1^\dagger$}}},
	every row 77 column 3/.style={postproc cell content/.style={@cell content={*$##1^\dagger$}}},
	every row 78 column 3/.style={postproc cell content/.style={@cell content={*$##1^\dagger$}}},
	every row 79 column 3/.style={postproc cell content/.style={@cell content={*$##1^\dagger$}}},
	every row 80 column 3/.style={postproc cell content/.style={@cell content={*$##1^\dagger$}}},
	every row 81 column 3/.style={postproc cell content/.style={@cell content={*$##1^\dagger$}}},
	every row 82 column 3/.style={postproc cell content/.style={@cell content={*$##1^\dagger$}}},
	every row 83 column 3/.style={postproc cell content/.style={@cell content={*$##1^\dagger$}}},
	every row 84 column 3/.style={postproc cell content/.style={@cell content={*$##1^\dagger$}}},
	every row 85 column 3/.style={postproc cell content/.style={@cell content={*$##1^\dagger$}}},
	every row 86 column 3/.style={postproc cell content/.style={@cell content={*$##1^\dagger$}}},
	every row 87 column 3/.style={postproc cell content/.style={@cell content={*$##1^\dagger$}}},
	every row 88 column 3/.style={postproc cell content/.style={@cell content={*$##1^\dagger$}}},
	every row 89 column 3/.style={postproc cell content/.style={@cell content={*$##1^\dagger$}}},
	every row 9 column 3/.style={postproc cell content/.style={@cell content={*$##1^\dagger$}}},
	every row 90 column 3/.style={postproc cell content/.style={@cell content={*$##1^\dagger$}}},
	every row 91 column 3/.style={postproc cell content/.style={@cell content={*$##1^\dagger$}}},
	every row 92 column 3/.style={postproc cell content/.style={@cell content={*$##1^\dagger$}}},
	every row 93 column 3/.style={postproc cell content/.style={@cell content={*$##1^\dagger$}}},
	every row 94 column 3/.style={postproc cell content/.style={@cell content={*$##1^\dagger$}}},
	every row 24 column 3/.style={postproc cell content/.style={@cell content={$##1\dagger$}}},
	every row 25 column 3/.style={postproc cell content/.style={@cell content={$##1\dagger$}}},
	every row 27 column 3/.style={postproc cell content/.style={@cell content={$##1\dagger$}}},
	every row 28 column 3/.style={postproc cell content/.style={@cell content={$##1\dagger$}}},
	every row 33 column 3/.style={postproc cell content/.style={@cell content={$##1\dagger$}}},
	every row 36 column 3/.style={postproc cell content/.style={@cell content={$##1\dagger$}}},
	every row 37 column 3/.style={postproc cell content/.style={@cell content={$##1\dagger$}}},
	every row 8 column 3/.style={postproc cell content/.style={@cell content={$##1\dagger$}}},
	every row 10 column 3/.style={postproc cell content/.style={@cell content={$##1\dagger$}}},
	every row 15 column 3/.style={postproc cell content/.style={@cell content={$##1\dagger$}}},
	every row 17 column 3/.style={postproc cell content/.style={@cell content={$##1\dagger$}}},
	every row 34 column 3/.style={postproc cell content/.style={@cell content={$##1\dagger$}}},
	every row 18 column 3/.style={postproc cell content/.style={@cell content={$##1\dagger$}}},
	every row 10 column 6/.style={postproc cell content/.style={@cell content={*$##1^\dagger$}}},
	every row 11 column 6/.style={postproc cell content/.style={@cell content={*$##1^\dagger$}}},
	every row 12 column 6/.style={postproc cell content/.style={@cell content={*$##1^\dagger$}}},
	every row 13 column 6/.style={postproc cell content/.style={@cell content={*$##1^\dagger$}}},
	every row 14 column 6/.style={postproc cell content/.style={@cell content={*$##1^\dagger$}}},
	every row 15 column 6/.style={postproc cell content/.style={@cell content={*$##1^\dagger$}}},
	every row 16 column 6/.style={postproc cell content/.style={@cell content={*$##1^\dagger$}}},
	every row 17 column 6/.style={postproc cell content/.style={@cell content={*$##1^\dagger$}}},
	every row 18 column 6/.style={postproc cell content/.style={@cell content={*$##1^\dagger$}}},
	every row 19 column 6/.style={postproc cell content/.style={@cell content={*$##1^\dagger$}}},
	every row 2 column 6/.style={postproc cell content/.style={@cell content={*$##1^\dagger$}}},
	every row 20 column 6/.style={postproc cell content/.style={@cell content={*$##1^\dagger$}}},
	every row 21 column 6/.style={postproc cell content/.style={@cell content={*$##1^\dagger$}}},
	every row 22 column 6/.style={postproc cell content/.style={@cell content={*$##1^\dagger$}}},
	every row 23 column 6/.style={postproc cell content/.style={@cell content={*$##1^\dagger$}}},
	every row 24 column 6/.style={postproc cell content/.style={@cell content={*$##1^\dagger$}}},
	every row 25 column 6/.style={postproc cell content/.style={@cell content={*$##1^\dagger$}}},
	every row 26 column 6/.style={postproc cell content/.style={@cell content={*$##1^\dagger$}}},
	every row 27 column 6/.style={postproc cell content/.style={@cell content={*$##1^\dagger$}}},
	every row 28 column 6/.style={postproc cell content/.style={@cell content={*$##1^\dagger$}}},
	every row 29 column 6/.style={postproc cell content/.style={@cell content={*$##1^\dagger$}}},
	every row 3 column 6/.style={postproc cell content/.style={@cell content={*$##1^\dagger$}}},
	every row 30 column 6/.style={postproc cell content/.style={@cell content={*$##1^\dagger$}}},
	every row 31 column 6/.style={postproc cell content/.style={@cell content={*$##1^\dagger$}}},
	every row 32 column 6/.style={postproc cell content/.style={@cell content={*$##1^\dagger$}}},
	every row 33 column 6/.style={postproc cell content/.style={@cell content={*$##1^\dagger$}}},
	every row 34 column 6/.style={postproc cell content/.style={@cell content={*$##1^\dagger$}}},
	every row 35 column 6/.style={postproc cell content/.style={@cell content={*$##1^\dagger$}}},
	every row 36 column 6/.style={postproc cell content/.style={@cell content={*$##1^\dagger$}}},
	every row 37 column 6/.style={postproc cell content/.style={@cell content={*$##1^\dagger$}}},
	every row 38 column 6/.style={postproc cell content/.style={@cell content={*$##1^\dagger$}}},
	every row 39 column 6/.style={postproc cell content/.style={@cell content={*$##1^\dagger$}}},
	every row 4 column 6/.style={postproc cell content/.style={@cell content={*$##1^\dagger$}}},
	every row 40 column 6/.style={postproc cell content/.style={@cell content={*$##1^\dagger$}}},
	every row 41 column 6/.style={postproc cell content/.style={@cell content={*$##1^\dagger$}}},
	every row 42 column 6/.style={postproc cell content/.style={@cell content={*$##1^\dagger$}}},
	every row 43 column 6/.style={postproc cell content/.style={@cell content={*$##1^\dagger$}}},
	every row 44 column 6/.style={postproc cell content/.style={@cell content={*$##1^\dagger$}}},
	every row 45 column 6/.style={postproc cell content/.style={@cell content={*$##1^\dagger$}}},
	every row 46 column 6/.style={postproc cell content/.style={@cell content={*$##1^\dagger$}}},
	every row 47 column 6/.style={postproc cell content/.style={@cell content={*$##1^\dagger$}}},
	every row 48 column 6/.style={postproc cell content/.style={@cell content={*$##1^\dagger$}}},
	every row 49 column 6/.style={postproc cell content/.style={@cell content={*$##1^\dagger$}}},
	every row 5 column 6/.style={postproc cell content/.style={@cell content={*$##1^\dagger$}}},
	every row 6 column 6/.style={postproc cell content/.style={@cell content={*$##1^\dagger$}}},
	every row 68 column 6/.style={postproc cell content/.style={@cell content={*$##1^\dagger$}}},
	every row 7 column 6/.style={postproc cell content/.style={@cell content={*$##1^\dagger$}}},
	every row 8 column 6/.style={postproc cell content/.style={@cell content={*$##1^\dagger$}}},
	every row 9 column 6/.style={postproc cell content/.style={@cell content={*$##1^\dagger$}}},
	every row 91 column 6/.style={postproc cell content/.style={@cell content={*$##1^\dagger$}}},
	every last row/.style={after row=\bottomrule\caption{Mean SMAPE Results}\label{tab:mean_smape_results}\\}, 
	end table=\end{longtable},
]{results/mean_SMAPE.csv}

In Table \ref{tab:mean_smape_results}, the best model in every dataset is indicated in boldface. Furthermore, as mentioned before in Section \ref{sec:statistical_tests}, since \texttt{auto.arima} and \texttt{ets} are the two established benchmarks chosen, we perform paired Wilcoxon signed-rank tests with Bonferroni correction for every model against the two benchmarks. In Table \ref{tab:mean_smape_results}, $\dagger$ and * denote those models which are significantly worse than \texttt{auto.arima} and \texttt{ets} respectively. Similarly, $\ddagger$ and ** denote the models which are significantly better than \texttt{auto.arima} and \texttt{ets} respectively. We see that on the CIF2016, M3, Tourism and M4 datasets no model is significantly better than the two benchmarks. On the NN5 dataset, the RNN models have only managed to perform significantly better than \texttt{auto.arima} but not \texttt{ets}. Moreover, across all the datasets it can be seen that the RNN models which perform significantly better than the benchmarks are mostly the variants of the Stacked architecture. Also, it is only on the Wikipedia Web Traffic dataset, that the pooled versions of the regression models perform significantly better than the benchmark \texttt{auto.arima}.

We observe that in general, increasing the number of lags from 10 to the size of the input window of the RNNs, improves accuracy of the regression models across all the datasets. On the datasets Wikipedia Web Traffic, M3 and NN5, the inclusion of the cross-series information alone has an effect since the pooled regression models with increased number of lags have performed better than the unpooled versions of regression. Furthermore, on the two datasets Wikipedia Web Traffic and NN5, the pooled regression models with increased number of lags have outperformed the \texttt{auto.arima} model which acts upon every series independently. Although the pooled models have performed worse than the unpooled models on the CIF, Tourism and M4 datasets, the RNN models have mostly outperformed both the unpooled and pooled regression models on these datasets. This indicates that even though the series in those datasets are not quite homogeneous to build global models, the use of RNN architectures has a clear effect on the final accuracy. In fact, on the CIF dataset the effect from the choice of the RNN architectures has managed to outperform the two statistical benchmarks \texttt{ets} and \texttt{auto.arima}. On the Wikipedia Web Traffic, M3 and NN5 datasets too, the RNNs have outperformed the statistical benchmarks as well as both the versions of the regression models implying that on these datasets the cross-series inclusion together with the choice of the RNNs have lead to better accuracy. 

Another significant observation is that on many datasets, both the Bayesian optimization and the hyperparameter tuning with 10-fold cross validation for the L2 regularization parameter of the pooled regression models select values close to 0 so that the models effectively have nearly no L2 regularization. This is why in many datasets the mean SMAPE values are the same for both with and without regularization. Therefore, for the pooled regression models we cannot conclude that L2 weight regularization improves the model performance. The model without any L2 regularization is already an appropriate fit for the data. For the unpooled regression models, we obtain mixed results. As discussed by \citet{BERGMEIR2018CV}, while being applicable to detect overfitting, a generic k-fold cross-validation may lead to an underestimation of the cross-validation errors for models that underfit. However, our pooled regression models do not show this behaviour, as they have very small L2 regularization parameters. As for the unpooled models, although for some series the L2 regularization parameters are high, it is the same case for both 10-fold cross-validation and the normal cross-validation. 

\subsection{Relative Performance of RNN Architectures}

\begin{figure*}[htbp!]
	\captionsetup{justification=centering}
	\hspace*{-0.75cm}
	\includegraphics[width=0.55\textwidth]{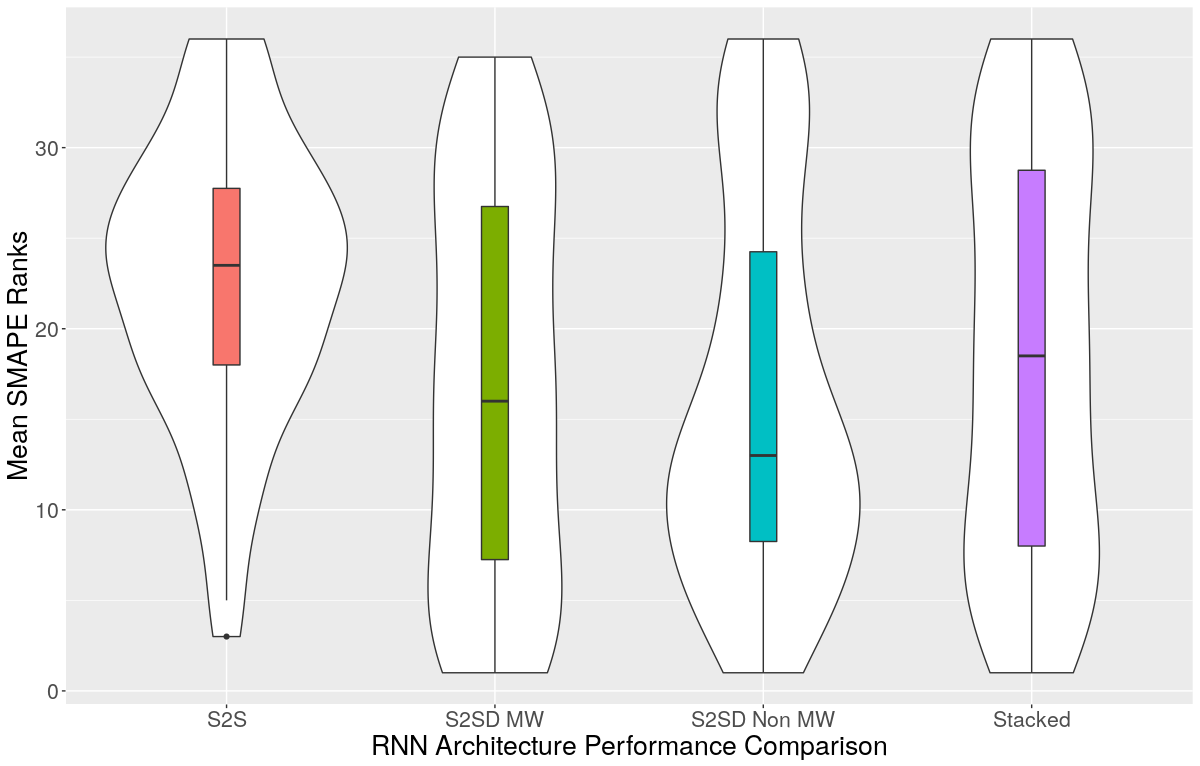}
	\includegraphics[width=0.55\textwidth]{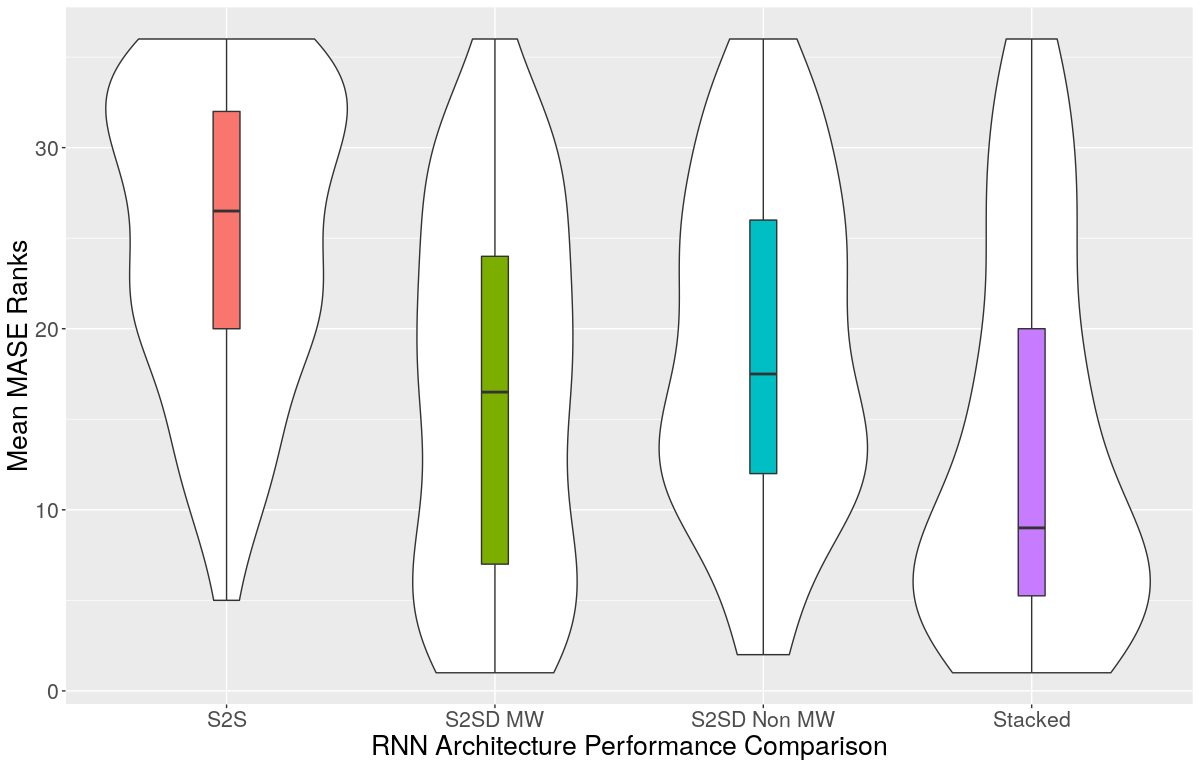}\\
	\hspace*{-0.75cm}
	\includegraphics[width=0.55\textwidth]{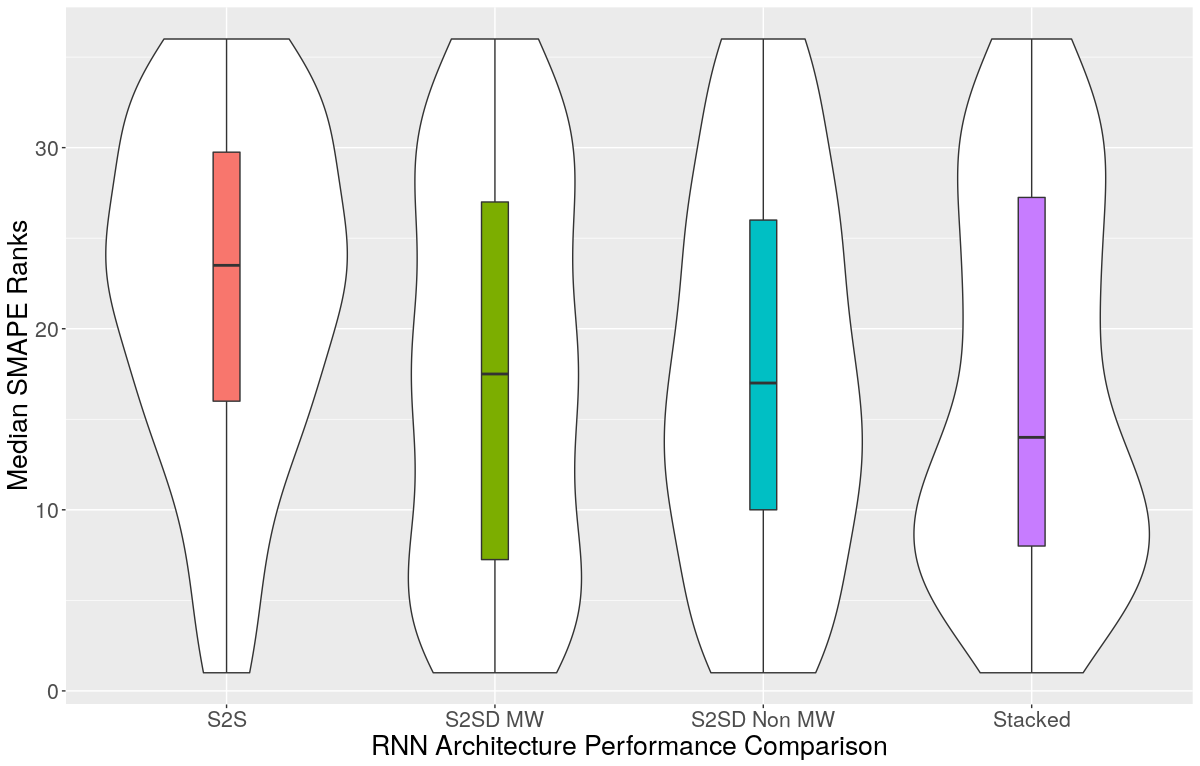}
	\includegraphics[width=0.55\textwidth]{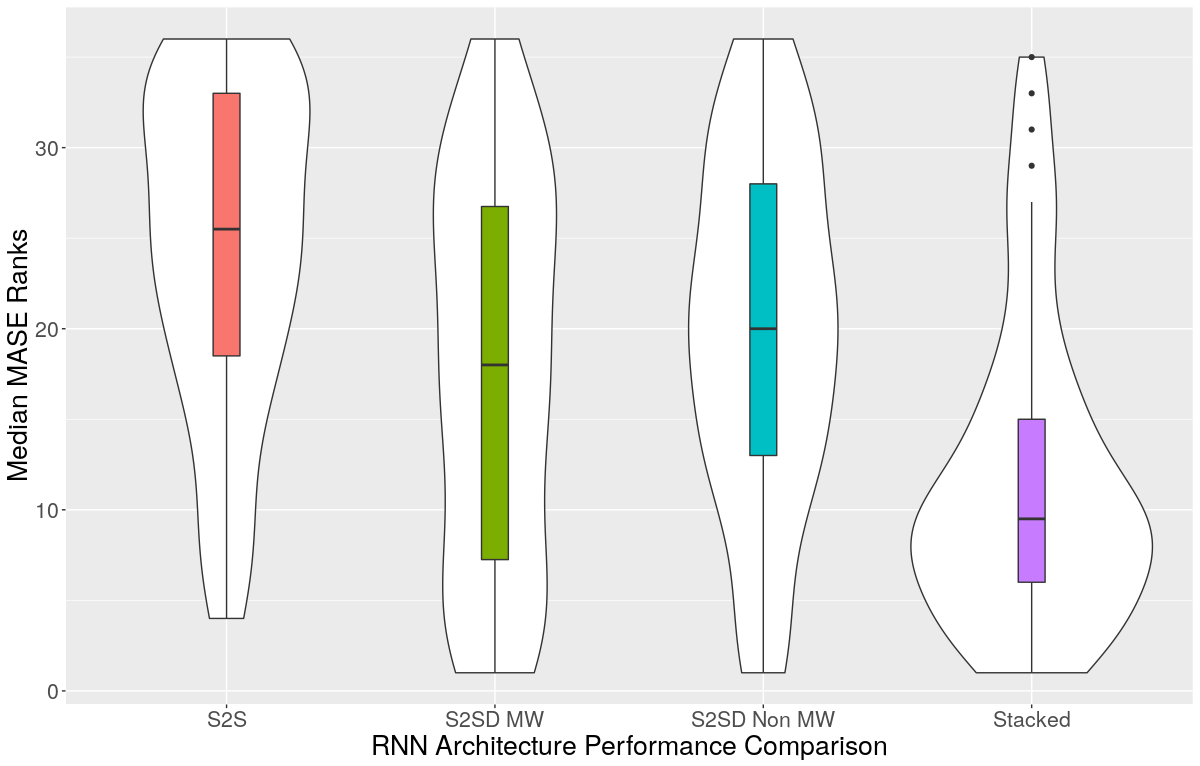}\\
	\hspace*{-0.75cm}
	\includegraphics[width=0.55\textwidth]{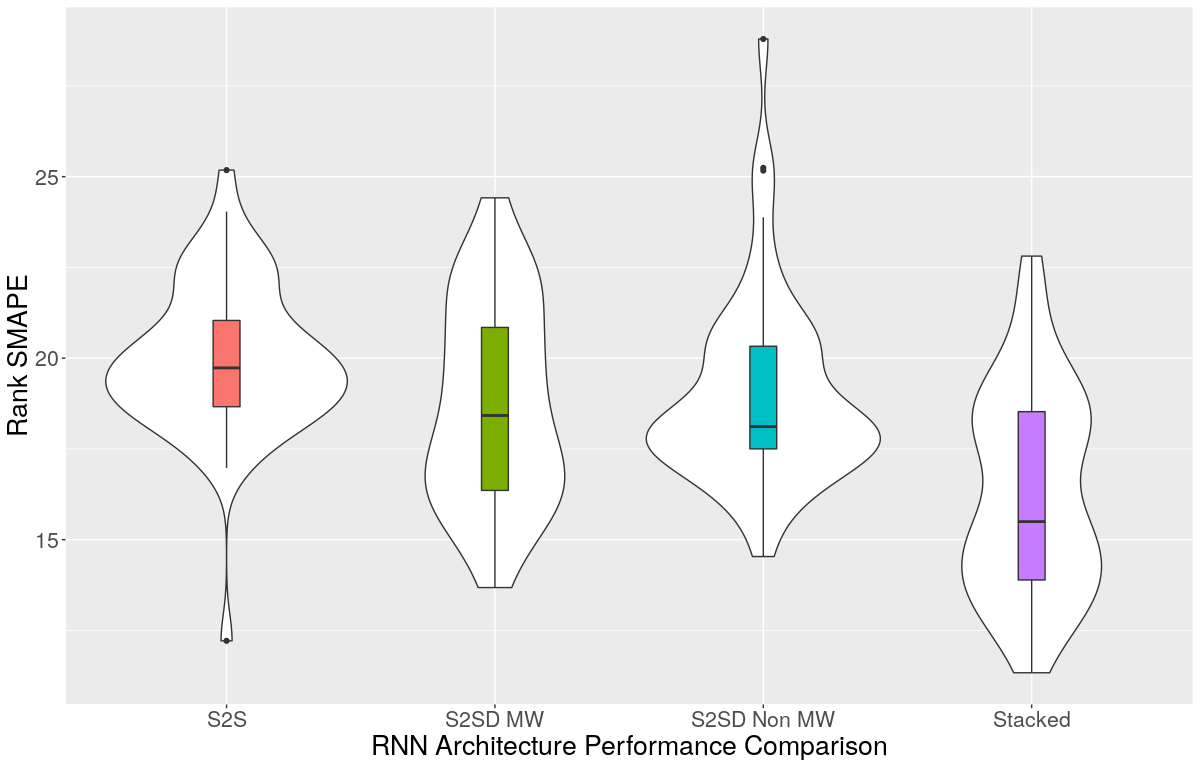}
	\includegraphics[width=0.55\textwidth]{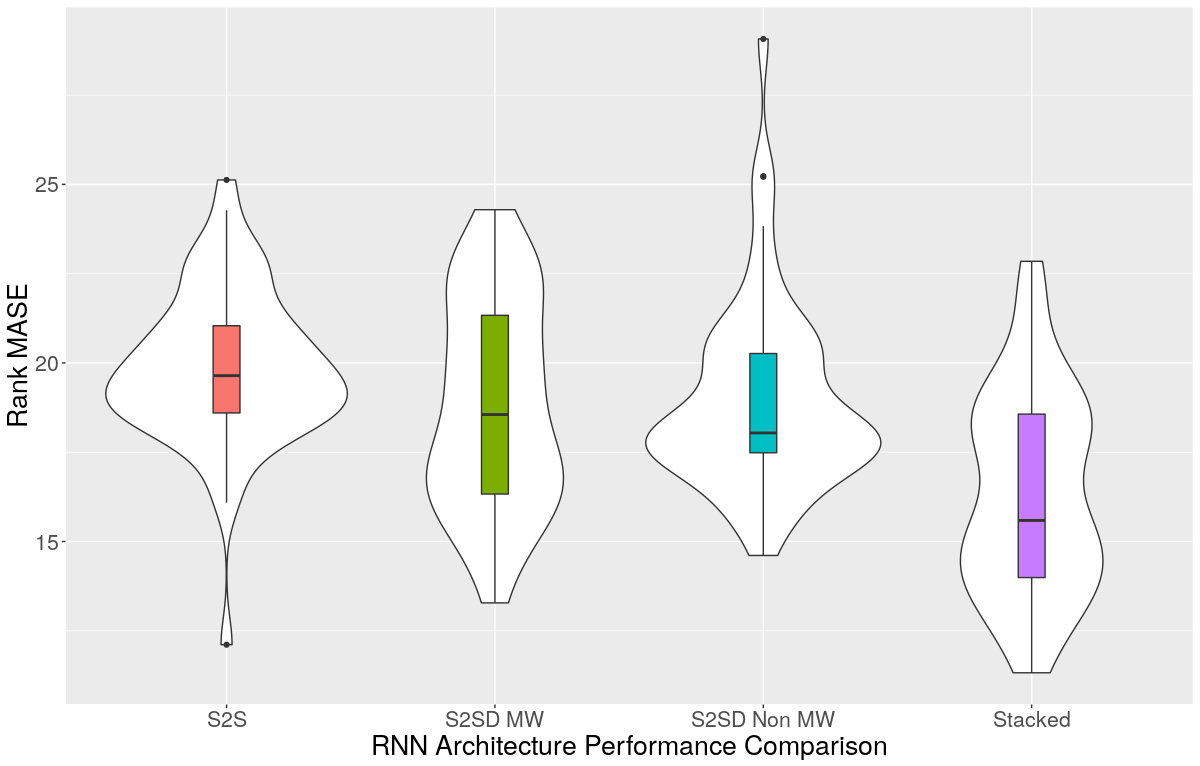}
	\caption{Relative Performance of Different RNN Architectures}
	\label{fig:rnn_performance}
\end{figure*}

We compare the relative performance of the different RNN architectures against each other across all the datasets, in terms of all the error metrics. The violin plots in Figure \ref{fig:rnn_performance} illustrate these results. 

We see that the best architecture differs based on the error metric. On mean SMAPE, the S2S with the Dense Layer (S2SD) architecture without the moving window, performs the best. However, on all the other error metrics, the Stacked architecture performs the best. With respect to the rank error plots, clearly the Stacked architecture produces the best results for most of the time series. Thus, the results on the mean SMAPE indicate that the Stacked architecture results in higher errors for certain outlying time series. The two versions of the S2SD architectures (with and without the moving window) perform comparably well, although not as good as the Stacked architecture on most cases. Therefore, in between the two versions of the S2SD architectures, it is hard to derive a clear conclusion about the best model. The S2S architecture (with the decoder) performs the worst overall.

With respect to the mean SMAPE values, the Friedman test of statistical significance gives an overall $p$-value of $\num{9.3568e-3}$ implying that the differences are statistically significant. The Hochberg's post-hoc procedure is performed by using as the control method the S2SD MW architecture which performs the best. The Stacked and the S2SD Non MW architectures do not perform significantly worse, with an adjusted $p$-value of 0.602. However, the S2S architecture performs significanlty worse with an adjusted $p$-value of \num{5.24e-3}. In terms of the median SMAPE, the Friedman test of statistical significance gives an overall $p$-value of $\num{0.349}$, implying that the differences of the models are not statistically significant. 

\subsection{Performance of Recurrent Units}
\label{sec:recurrent_unit_performance}

\begin{figure*}[htbp!]
	\captionsetup{justification=centering}
	\hspace{-1cm}
	\includegraphics[width=0.55\textwidth]{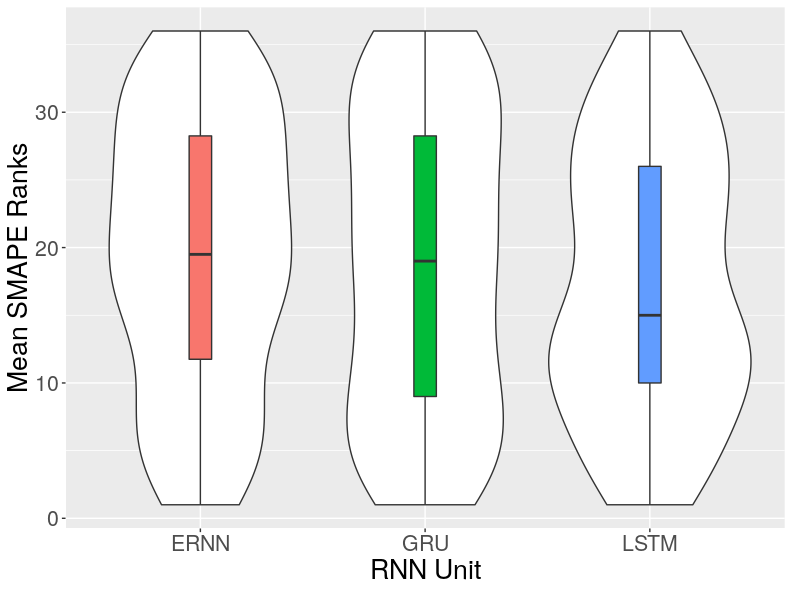}
	\includegraphics[width=0.55\textwidth]{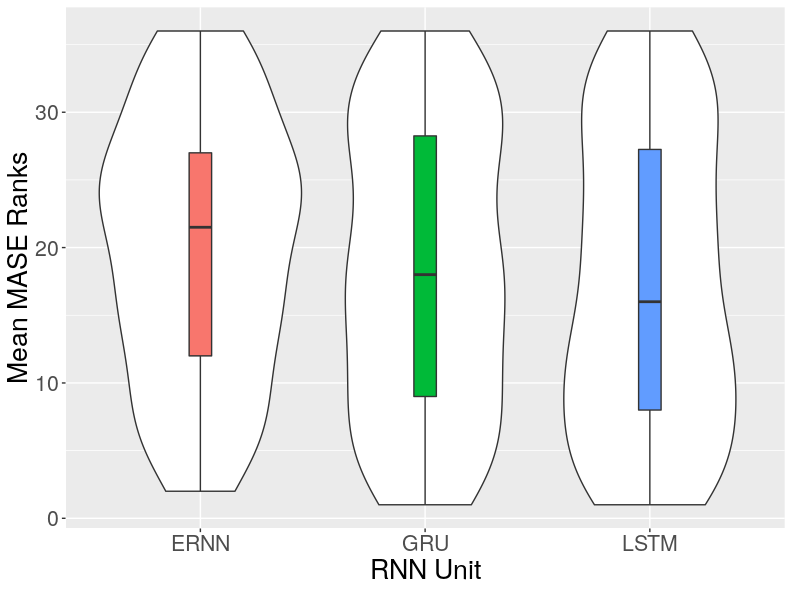}\\
	\caption{Relative Performance of Different Recurrrent Cell Types}
	\label{fig:rnn_unit_performance}
\end{figure*}

The violin plots in Figure \ref{fig:rnn_unit_performance} demonstrate the performance comparison of the different RNN units, in terms of mean SMAPE ranks and mean MASE ranks. The Friedman test of statistical significance with respect to the mean SMAPE values produces an overall $p$-value of $0.101$. Though the $p$-value does not indicate statistical significance, from the plots, we can derive that the LSTM with peephole connections cell performs the best. The ERNN cell performs the worst and the GRU exhibits a performance in-between these two.

\subsection{Performance of Optimizers}

\begin{figure*}[htbp!]
	\captionsetup{justification=centering}
	\hspace{-1cm}
	\includegraphics[width=0.55\textwidth]{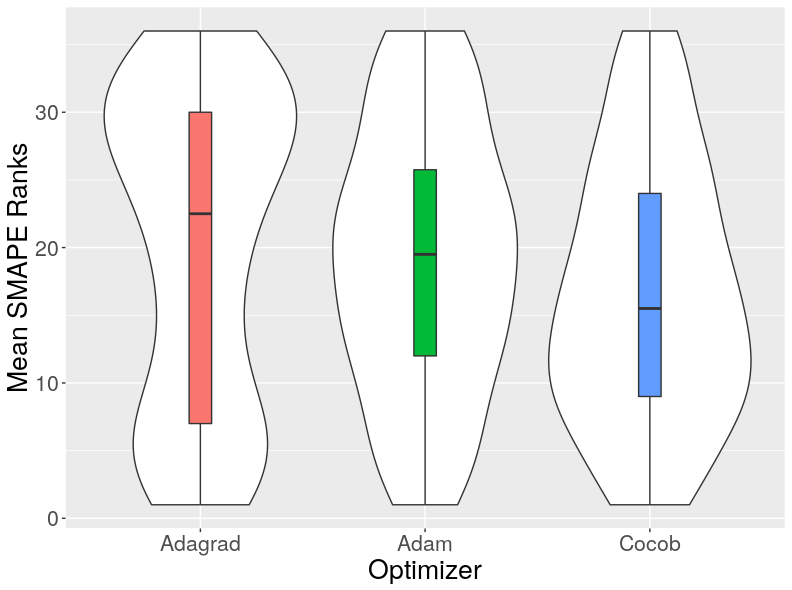}
	\includegraphics[width=0.55\textwidth]{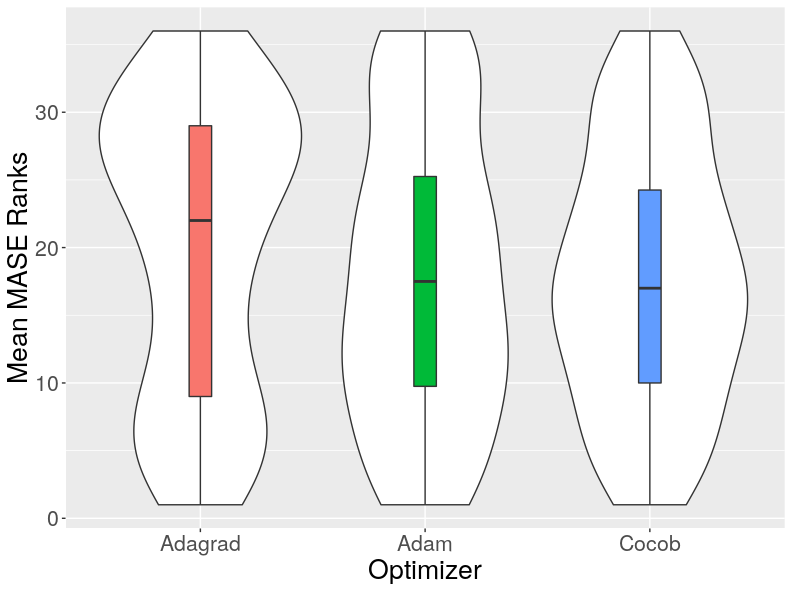}\\
	\caption{Relative Performance of Different Optimizers}
	\label{fig:optimizer_performance}
\end{figure*}

The violin plots in Figure \ref{fig:optimizer_performance} illustrate the performance comparison of the different optimizers, in terms of both the mean SMAPE ranks and the mean MASE ranks. From the plots we see that the Adagrad optimizer performs the worst complying with the findings in the literature stated under the Section \ref{sec:learning_algorithms}. Eventhough the Adam optimizer has been the best optimizer thus far, we can conclude from this study that the COCOB optimizer performs the best out of the three. However, the Adam optimizer also shows quite competitive performance in-between these two. 

The Friedman test of statistical significance with respect to the mean SMAPE values gives an overall $p$-value of $0.115$. Although there is no strong statistical evidence in terms of significance, we conclude that the Cocob optimizer is further preferred over the other two, since it does not require to set an initial learning rate. This supports fully automating the forecasting approach, since it eliminates the tuning of one more external hyperparameter. 

\subsection{Performance of the Output Components for the Sequence to Sequence Architecture}

\begin{figure*}[htbp!]
	\captionsetup{justification=centering}
	\hspace{-1cm}
	\includegraphics[width=0.55\textwidth]{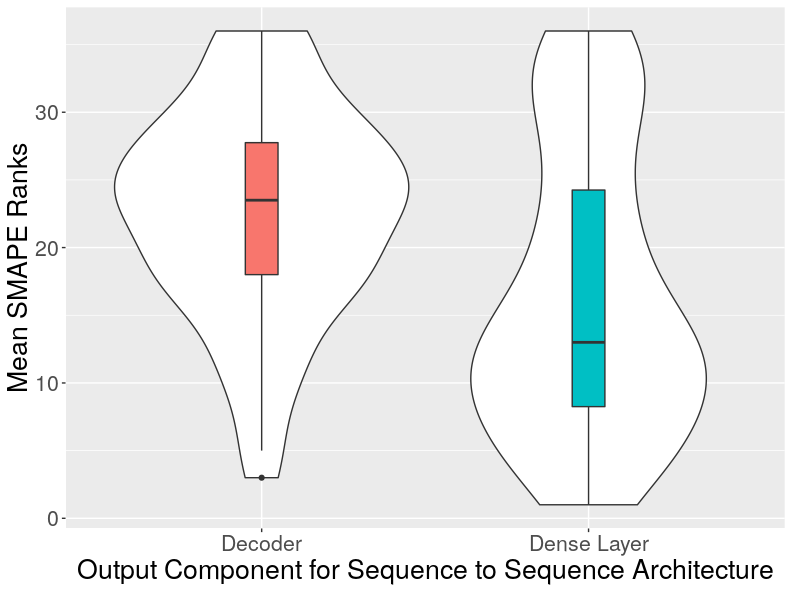}
	\includegraphics[width=0.55\textwidth]{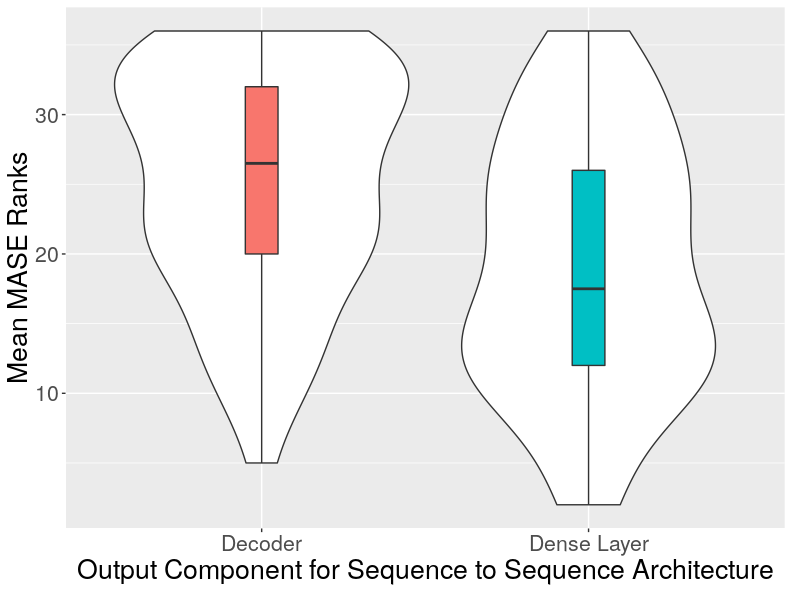}\\
	\caption{Comparison of the Ouput Component for the Sequence to Sequence with the Dense Layer Architecture}
	\label{fig:output_component_performance}
\end{figure*}

Figure \ref{fig:output_component_performance} shows the relative performance of the two output components for the S2S architecture. The dense layer performs better than the decoder owing to the error accumulation issue associated with the teacher forcing used in the decoder as mentioned in Section \ref{sec:rnn_architectures_review}. With teacher forcing, the autoregressive connections in the decoder tend to carry forward the errors generating from each forecasting step, resulting in even more uncertainty of the forecasts along the prediction horizon. The output from the paired Wilcoxon signed-rank test with respect to the mean SMAPE values gives an overall $p$-value of \num{2.064e-4} which indicates that the decoder performs significantly worse than the dense layer. 

\subsection{Comparison of Input Window Sizes for the Stacked Architecture}

\begin{figure*}[htbp!]
	\captionsetup{justification=centering}
	\hspace*{-0.75cm}
	\includegraphics[width=0.55\textwidth]{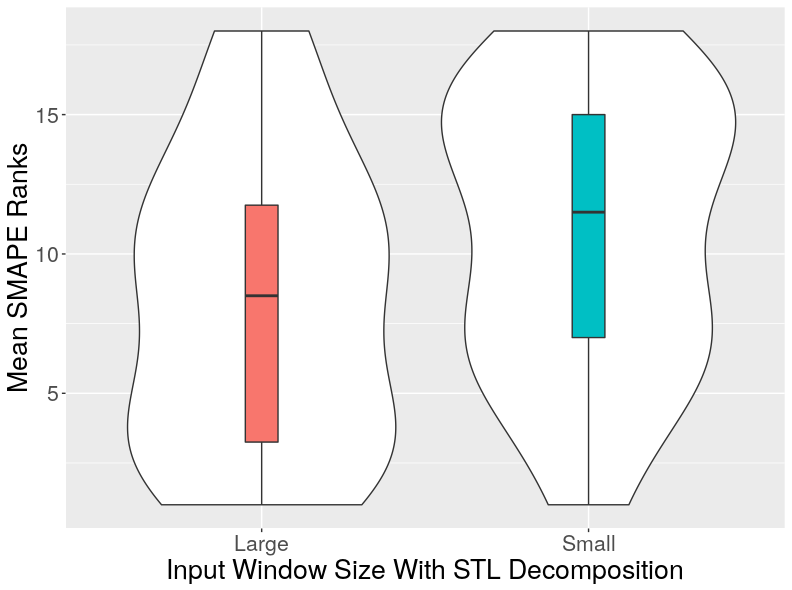}
	\includegraphics[width=0.55\textwidth]{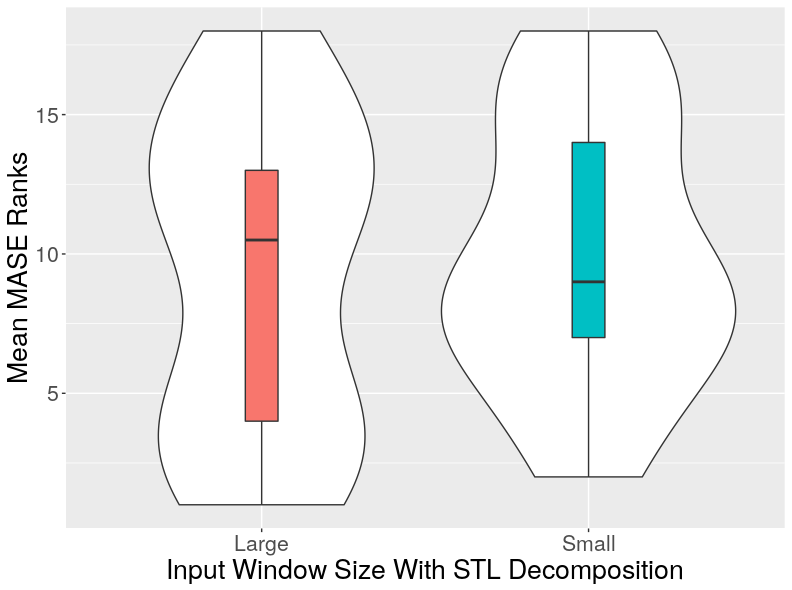}\\
	\hspace*{-0.75cm}
	\includegraphics[width=0.55\textwidth]{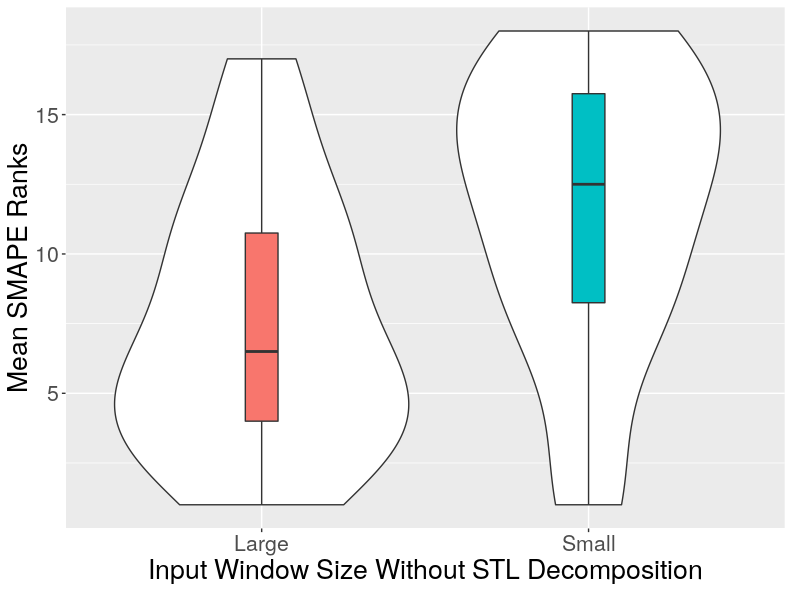}
	\includegraphics[width=0.55\textwidth]{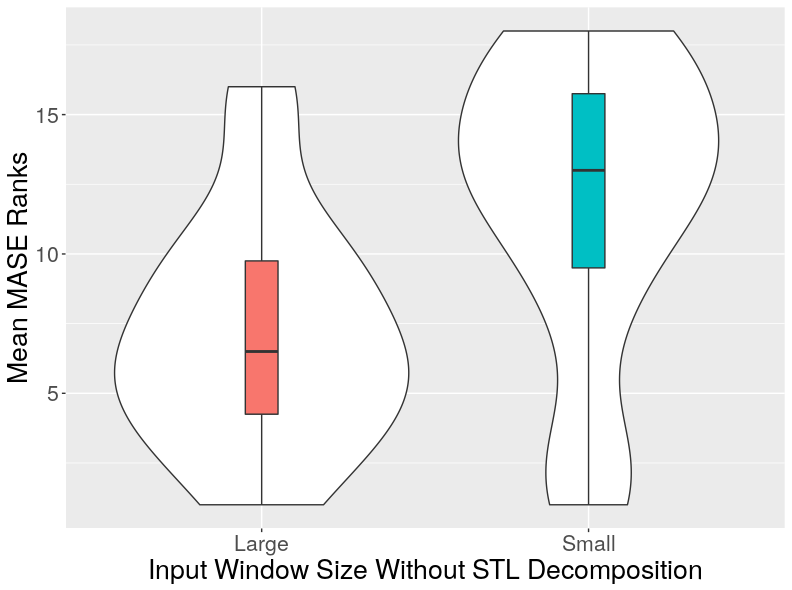}\\
	\caption{Comparison of Input Window Sizes for the Stacked Architecture}
	\label{fig:input_window_size_comparison}
\end{figure*}

Figure \ref{fig:input_window_size_comparison} shows the comparison of the two input window size options for the Stacked architecture on the two daily datasets, NN5 and Wikipedia Web Traffic. Both with and without STL Decomposition results are plotted. Here, `Large' denotes an input window size slightly larger than the expected prediction horizon while `Small' denotes an input window size slightly larger than the seasonality period which is 7 (for the daily data). From the plots, we can state that large input window sizes help the Stacked architecture both when STL Decomposition is used and not used. However, with mean MASE ranks, small input window sizes also perform comparably. For the case when the seasonality is not removed, making the input window size large improves the accuracy by a huge margin. Thus, large input windows make it easier for the Stacked architecture to learn the underlying seasonal patterns in the time series. The output from the paired Wilcoxon signed-rank test with respect to the mean SMAPE values gives an overall $p$-value of \num{2.206e-3} which indicates that the small input window size performs significantly worse than the large input window size when both with STL Decomposition and without STL Decomposition cases are considered together. 

\subsection{Analysis of Seasonality Modelling}

\begin{figure*}
	\captionsetup{justification=centering}
	\vspace{-1cm}
	\includegraphics[width=0.4\textwidth]{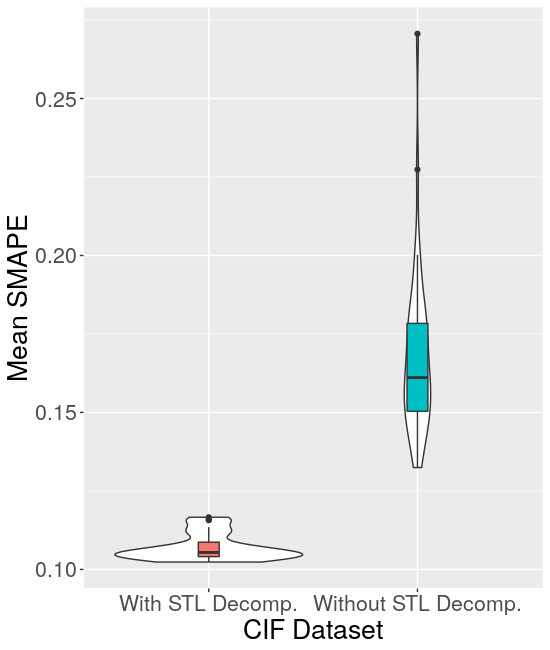}
	\includegraphics[width=0.4\textwidth]{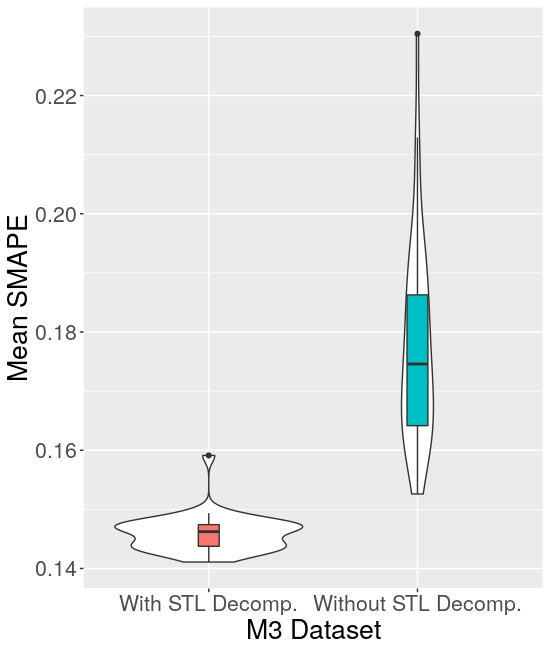}
	\includegraphics[width=0.4\textwidth]{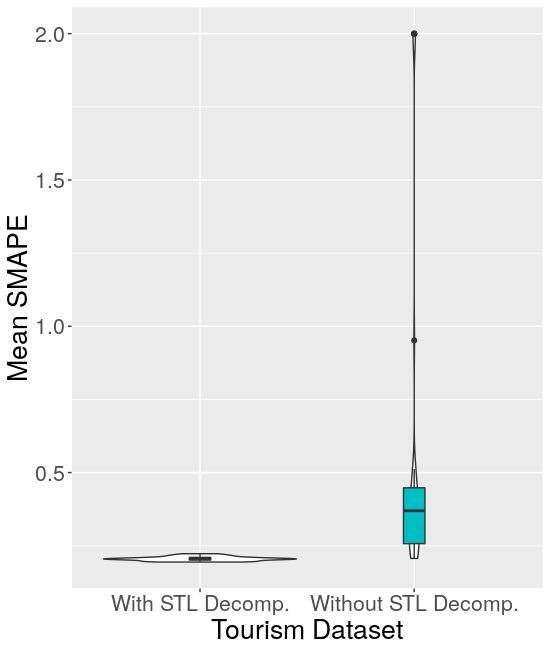}
	\includegraphics[width=0.4\textwidth]{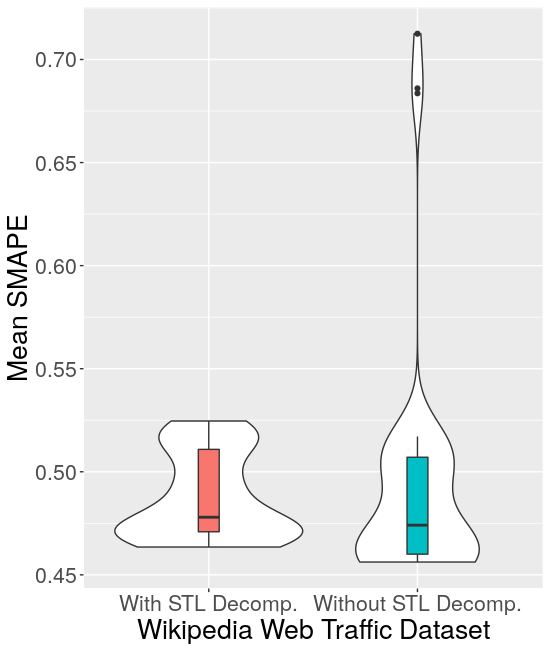}
	\includegraphics[width=0.4\textwidth]{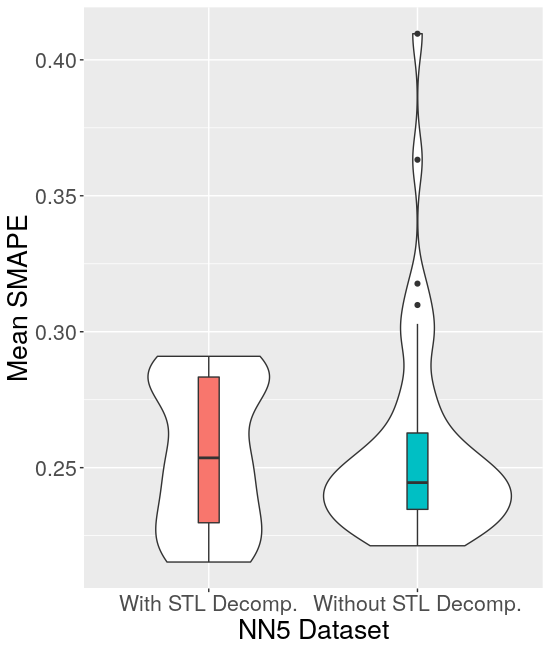}
	\caption{Comparison of the Performance with and without STL Decomposition - Mean SMAPE}
	\label{fig:stl_comparison_mean_smape}
\end{figure*}

\begin{figure*}[htbp!]
	\captionsetup{justification=centering}
	\vspace{-1cm}
	\includegraphics[width=0.4\textwidth]{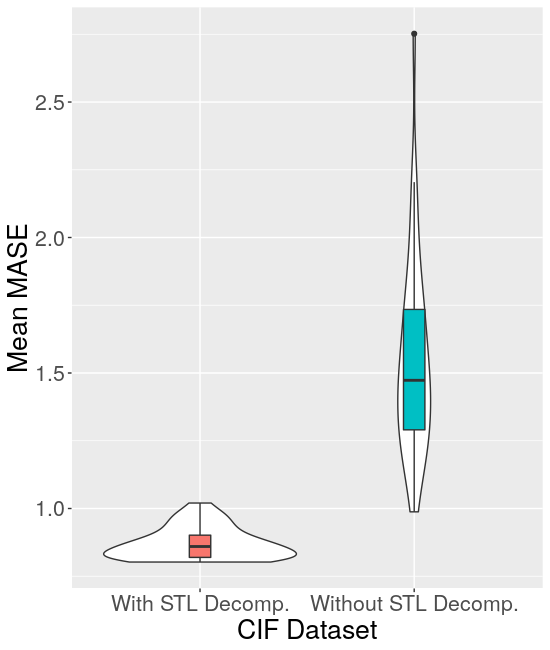}
	\includegraphics[width=0.4\textwidth]{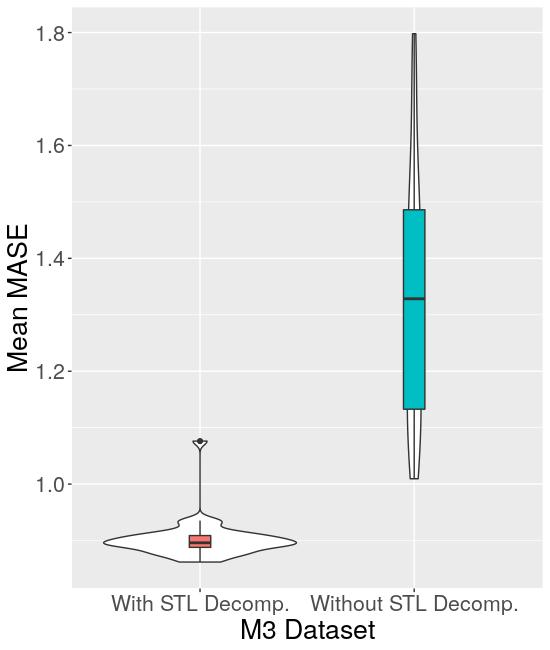}
	\includegraphics[width=0.4\textwidth]{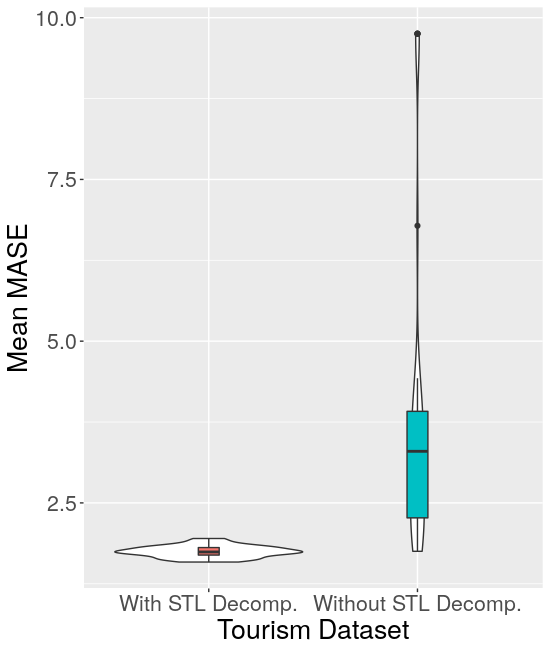}
	\includegraphics[width=0.4\textwidth]{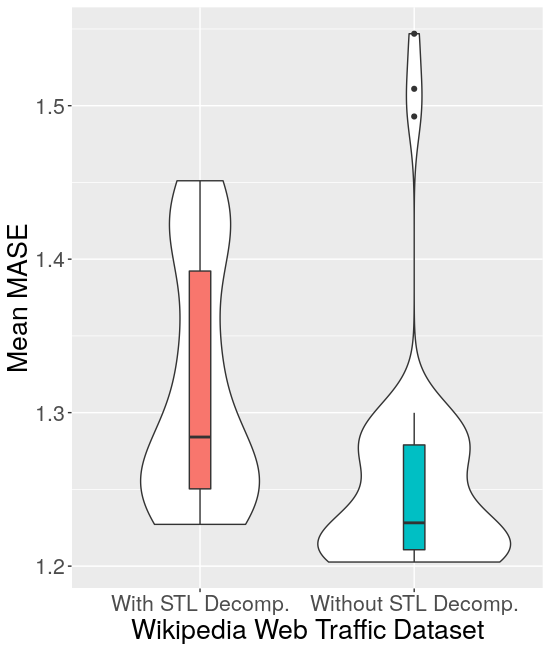}
	\includegraphics[width=0.4\textwidth]{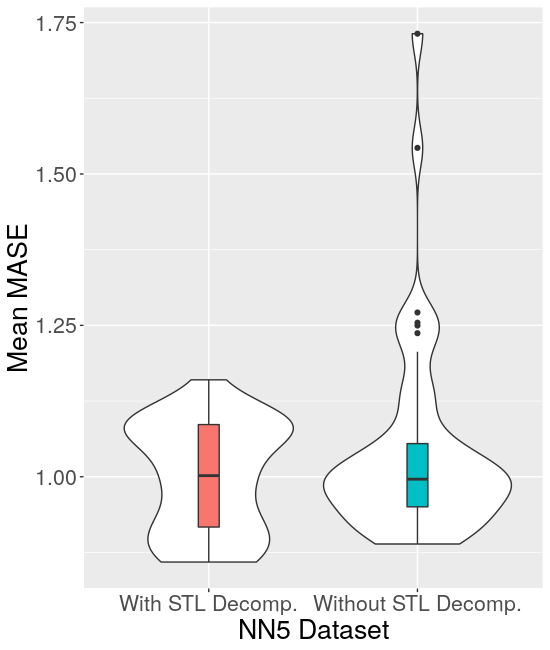}
	\caption{Comparison of the Performance with and without STL Decomposition - Mean MASE}
	\label{fig:stl_comparison_mean_mase}
\end{figure*}

The results from comparing the models with STL Decomposition and without STL Decomposition are as illustrated in Figures \ref{fig:stl_comparison_mean_smape} and \ref{fig:stl_comparison_mean_mase}, for the mean SMAPE metric and the mean MASE metric respectively. 

Due to the scale of the M4 monthly dataset, we do not run all the models without removing seasonality on that dataset. Rather, only the best few models by looking at the results from the first stage with removed seasonality, are selected to run without removing seasonality. As a consequence, we do not plot those results here. The plots indicate that on the CIF, M3 and the Tourism datasets, removing seasonality works better than modelling seasonality with the NN itself. On the M4 monthly dataset too, the same observation holds looking at the results in Table \ref{tab:mean_smape_results}. However, on the Wikipedia Web Traffic dataset, except for few outliers, modelling seasonality using the NN itself works almost as good as removing seasonality beforehand. This can be attributed to the fact that in the Wikipedia Web Traffic dataset, all the series have quite minimal seasonality according to the violin plots in Figure \ref{fig:seasonality_strengths}, so that there is no big difference in the data if the seasonality is removed or not. On the other hand, on the NN5 dataset too, NNs seem to be able to reasonably model seasonality on their own, except for few outliers.

\begin{figure*}[htbp!]
	\vspace*{-1.5cm}
	\includegraphics[width=\textwidth]{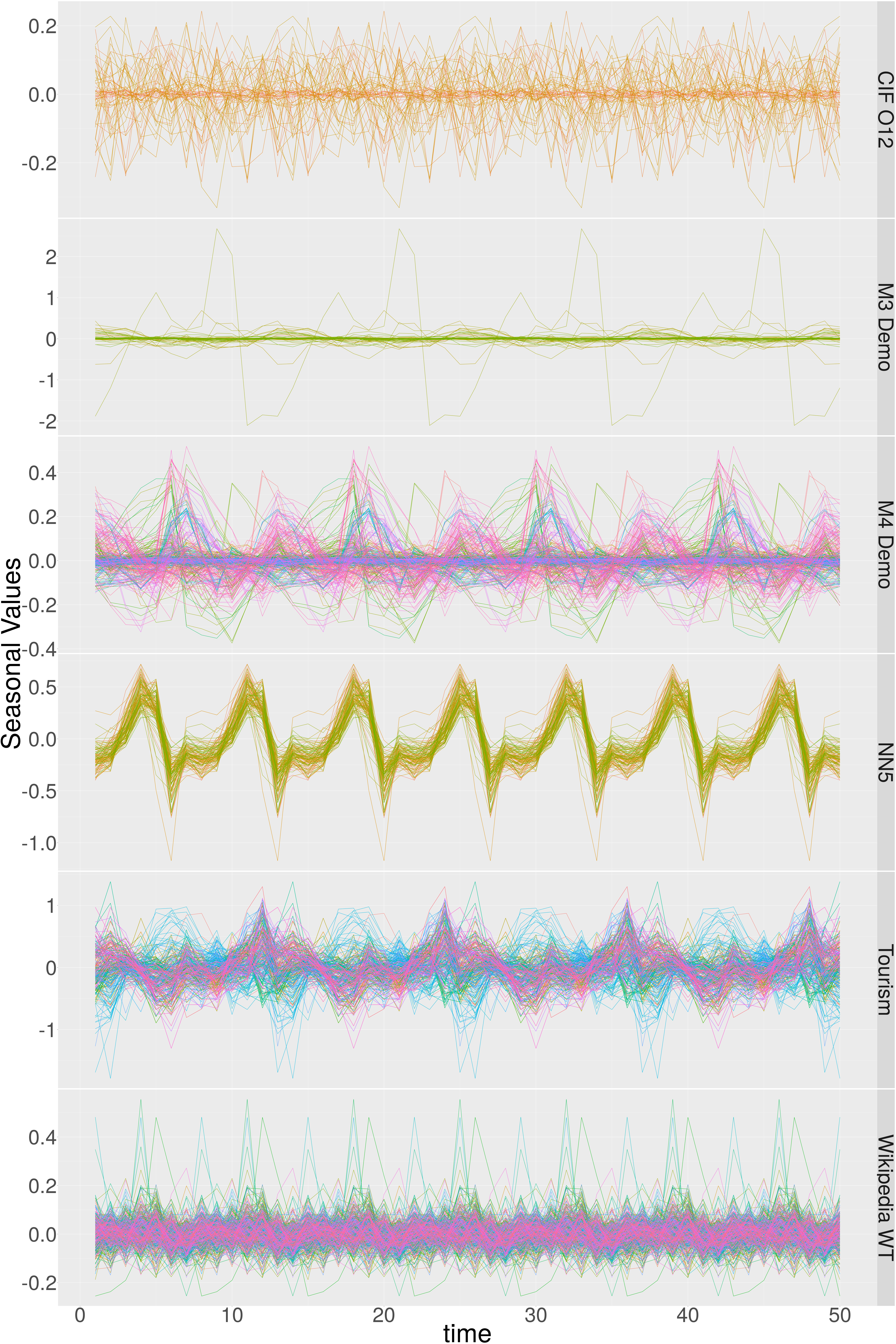}
	\caption{Seasonal Patterns of All the Datasets}
	\label{fig:seasonality_patterns}
\end{figure*}

To further explain this result, we also plot the seasonal patterns of the different datasets as shown in the Figure \ref{fig:seasonality_patterns}.
For the convenience of illustration, in every category of the M4 monthly dataset and the Wikipedia Web Traffic dataset, we plot only the first 400 series. Furthermore, from every series of every dataset, only the first 50 time steps are plotted. More seasonal pattern plots for the different categories of the CIF, M3 monthly and M4 monthly datasets are available in the Online Appendix\footnote{\url{https://drive.google.com/file/d/16rdzLTFwkKs-_eG_MU0i_rDnBQBCjsGe/view?usp=sharing}}.

In the NN5 dataset, almost all the series have the same seasonal pattern as in the $4^{th}$ plot of Figure \ref{fig:seasonality_patterns}, with all of them having the same length as indicated in Table \ref{tab:dataset_overview}. All the series in the NN5 dataset start and end on the same dates\citep{Crone2008-ye}. Also, according to Figure \ref{fig:seasonality_strengths}, the NN5 dataset has higher seasonality, with the seasonality strengths and patterns having very little variation among the individual series. In contrast, for the Tourism dataset, although it contains series with higher seasonality, it has a high variation of the individual seasonality strengths and patterns as well as the lengths of the series. The starting and ending dates are different among the series. Looking at the $5^{th}$ plot of Figure \ref{fig:seasonality_patterns}, it is also evident that the series have very different seasonal patterns. For the rest of the datasets too, the lengths as well as the seasonal patterns of the individual series vary considerably. Therefore, from these observations we can conclude that NNs are capable of modelling seasonality on their own, when all the series in the dataset have similar seasonal patterns and the lengths of the time series are equal, with the start and the end dates coinciding. 

Table \ref{tab:statistical_test_seasonality_modelling} shows the rankings as well as the overall $p$-values obtained from the paired Wilcoxon signed-rank tests for comparing the cases with and without STL Decomposition in every dataset. The overall $p$-values are calculated for each dataset separately by comparing the with STL Decomposition and without STL Decomposition cases in all the available models.

\begin{table*}[!htp]
	\centering
	\begin{tabular}{cccc}
		\hline
		Dataset&\multicolumn{2}{c}{Ranking}&Overall $p$-value\\
		\cline{2-3}
		& With STL Decomp. & Without STL Decomp. &\\
		\hline
		CIF&\textbf{1.0}&2.0&\num{2.91e-11}\\
		M3&\textbf{1.0}&2.0& \num{2.91e-11}\\
		Tourism&\textbf{1.06}&1.94&\num{1.455e-10}\\
		Wikipedia Web Traffic&1.73&\textbf{1.27}&0.028\\
		NN5&1.56&\textbf{1.44}&0.911\\
		\hline
	\end{tabular}
	\caption{Average rankings and results of the statistical testing for seasonality modelling with respect to mean SMAPE across all the datasets. On the CIF, M3 and the Tourism datasets, using STL Decomposition has the best ranking. On the Wikipedia Web Traffic and the NN5 datasets, not using STL Decomposition has the best ranking. The calculated overall $p$-values obtained from the paired Wilcoxon signed-rank test are shown for the different datasets on the Overall $p$-value column. On the CIF, M3 and the Tourism datasets, eliminating STL Decomposition performs significantly worse than applying STL Decomposition. However, on the Wikipedia Web Traffic dataset, applying STL Decomposition performs significantly worse than not applying it. On the NN5 dataset, there is no significant difference between using and not using STL Decomposition.}
	\label{tab:statistical_test_seasonality_modelling}
\end{table*}

\subsection{Performance of RNN Models Vs. Traditional Univariate Benchmarks}

\begin{figure*}[htbp!]
	\captionsetup{justification=centering}
	\vspace*{-2cm}
	\includegraphics[width=0.5\textwidth]{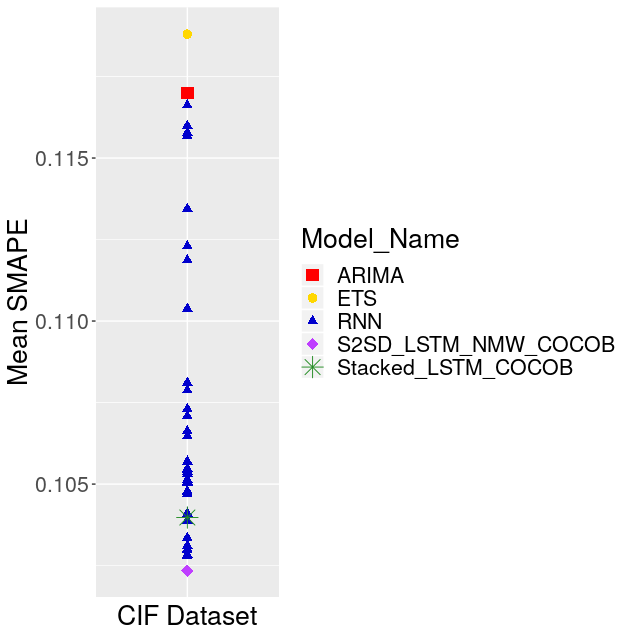}
	\includegraphics[width=0.5\textwidth]{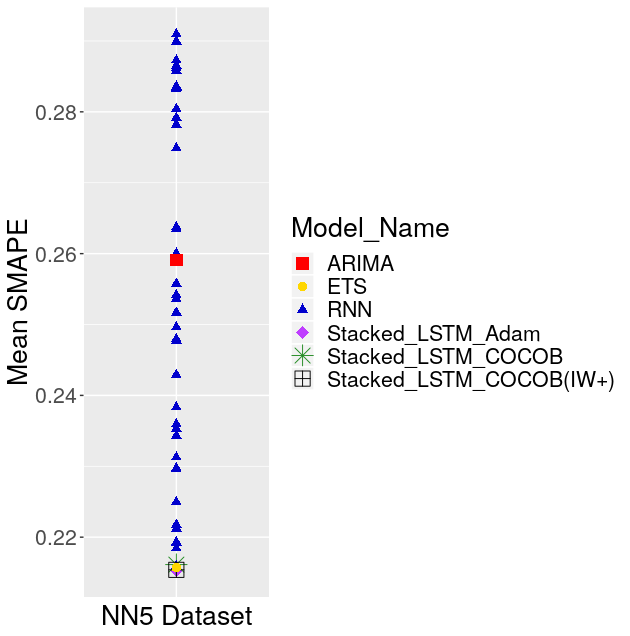}
	\includegraphics[width=0.5\textwidth]{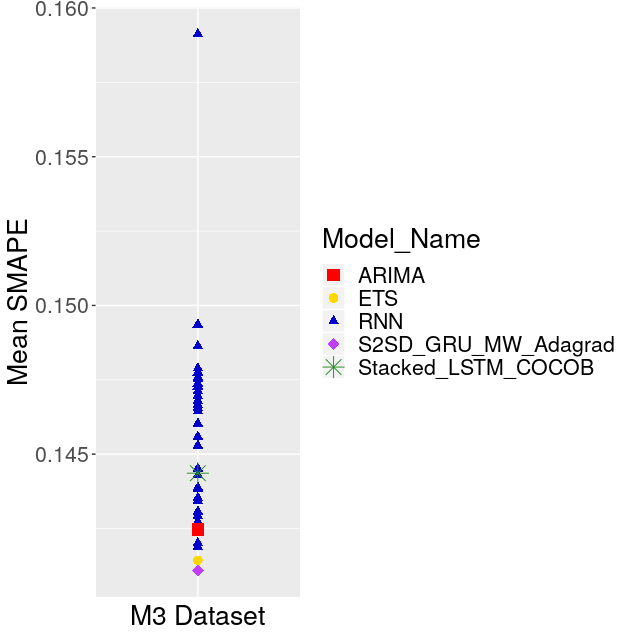}
	\includegraphics[width=0.5\textwidth]{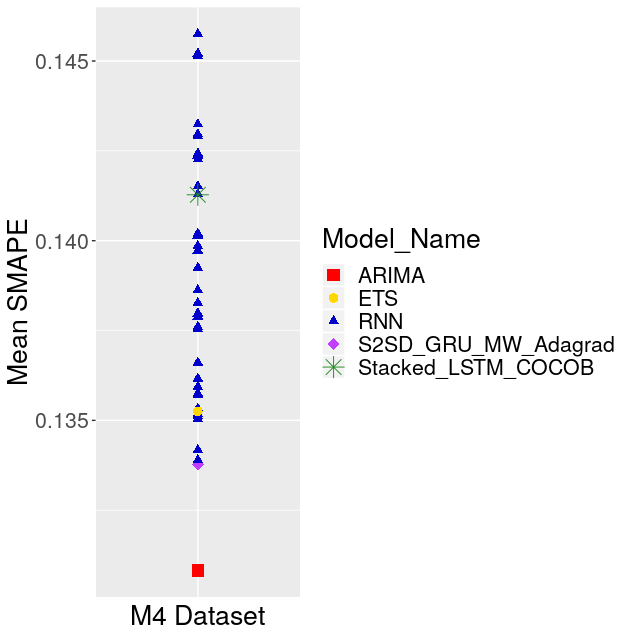}
	\includegraphics[width=0.5\textwidth]{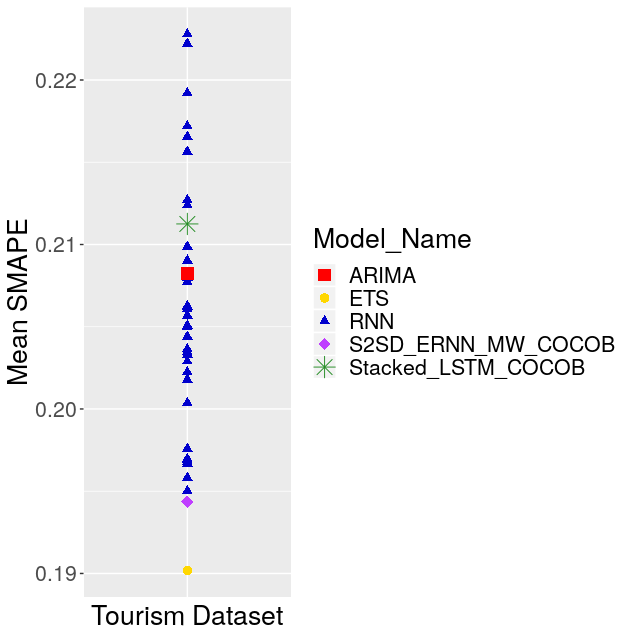}
	\includegraphics[width=0.5\textwidth]{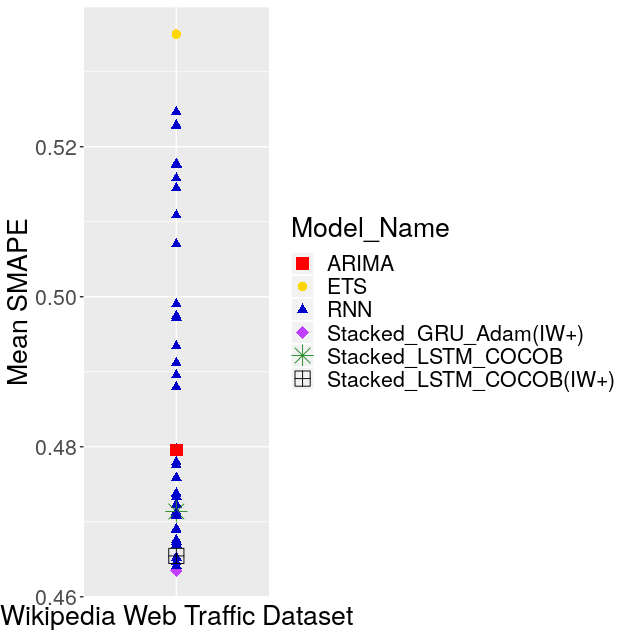}
	\caption{Performance of RNNs Compared to Traditional Univariate Techniques - Mean SMAPE}
	\label{fig:rnn_mean_smape_performance}
\end{figure*}

\begin{figure*}[htbp!]
	\captionsetup{justification=centering}
	\vspace*{-2cm}
	\includegraphics[width=0.5\textwidth]{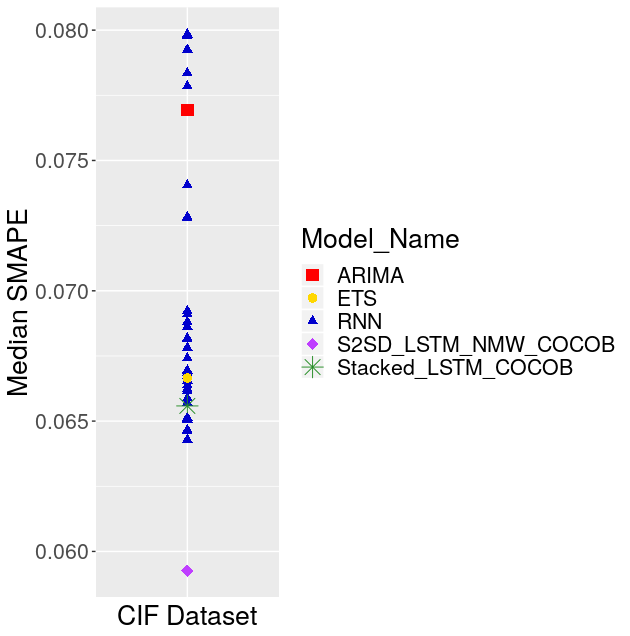}
	\includegraphics[width=0.5\textwidth]{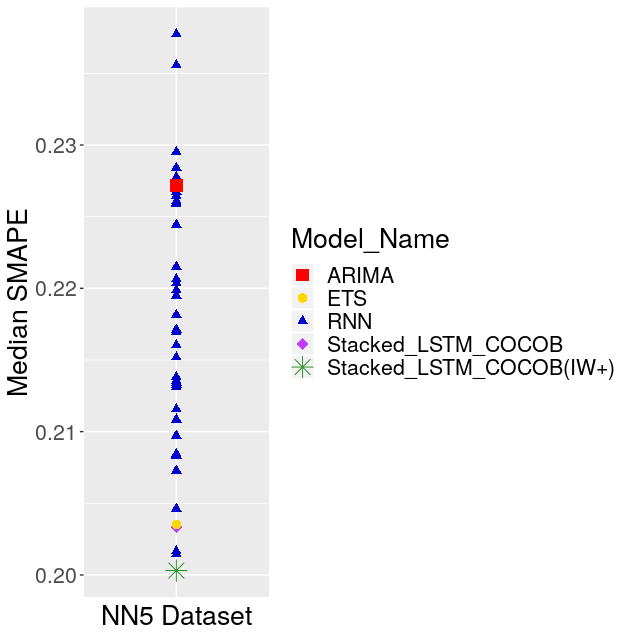}
	\includegraphics[width=0.5\textwidth]{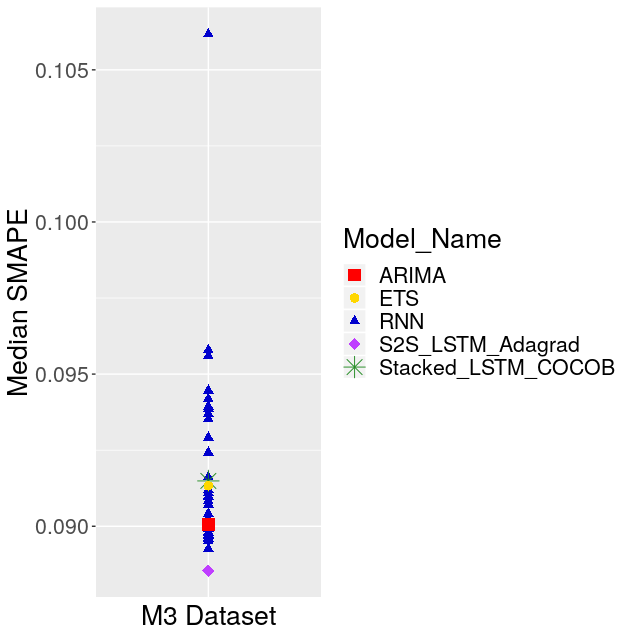}
	\includegraphics[width=0.5\textwidth]{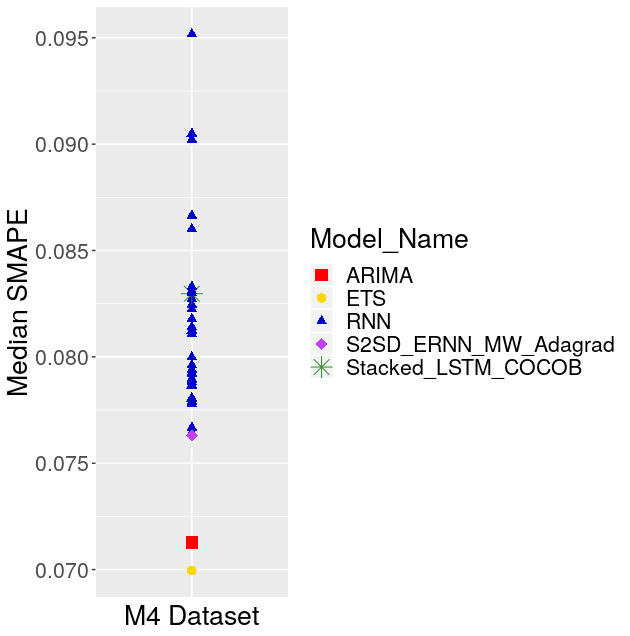}
	\includegraphics[width=0.5\textwidth]{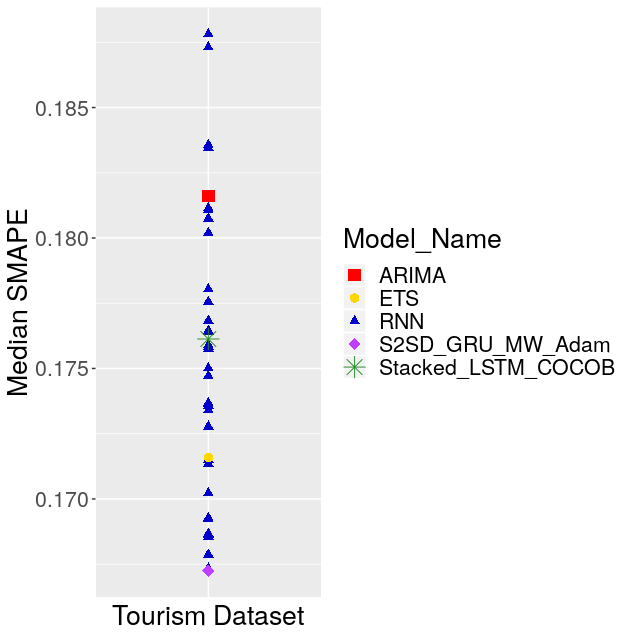}
	\includegraphics[width=0.5\textwidth]{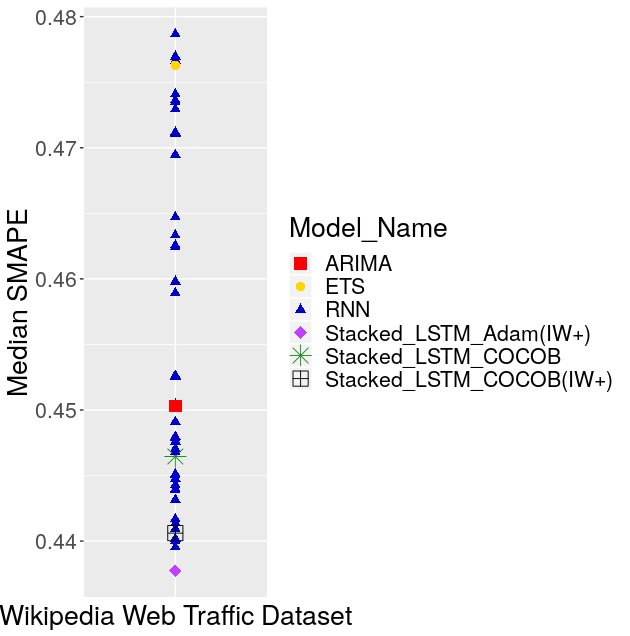}
	\caption{Performance of RNNs Compared to Traditional Univariate Techniques - Median SMAPE}
	\label{fig:rnn_median_smape_performance}
\end{figure*}

The relative performance of the model types in terms of the mean SMAPE metric and median SMAPE metric are shown in Figure \ref{fig:rnn_mean_smape_performance} and Figure \ref{fig:rnn_median_smape_performance} respectively. More comparison plots in terms of the mean MASE and median MASE metrics are available in the Online Appendix\footnote{\url{https://drive.google.com/file/d/16rdzLTFwkKs-_eG_MU0i_rDnBQBCjsGe/view?usp=sharing}}. We identify as a representative suggested model combination from the above sections the Stacked architecture with LSTM cells with peephole connections and the COCOB optimizer, and therefore indicated it in every plot as 'Stacked\_LSTM\_COCOB' (With small and large input window sizes). If the best RNN on each dataset differs from Stacked\_LSTM\_COCOB, these best models are also shown in the plots. All the other RNNs are indicated as 'RNN'.

\begin{figure*}[htbp!]
	\captionsetup{justification=centering}
	\vspace{-2cm}
	\includegraphics[width=0.5\textwidth]{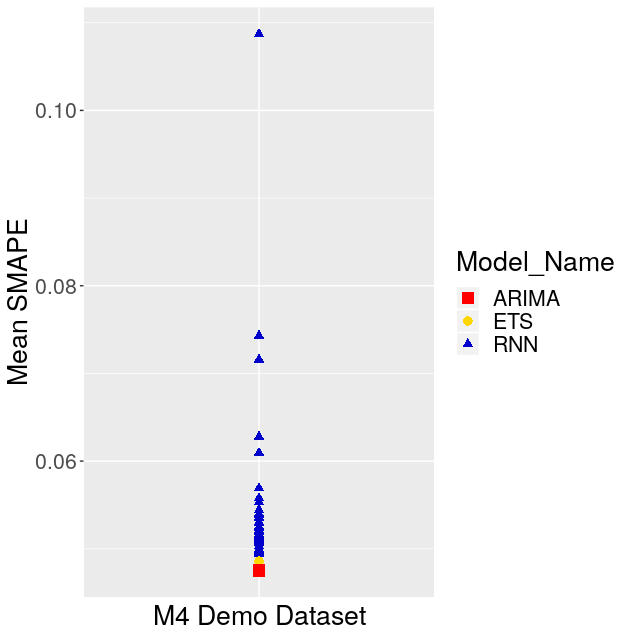}
	\includegraphics[width=0.5\textwidth]{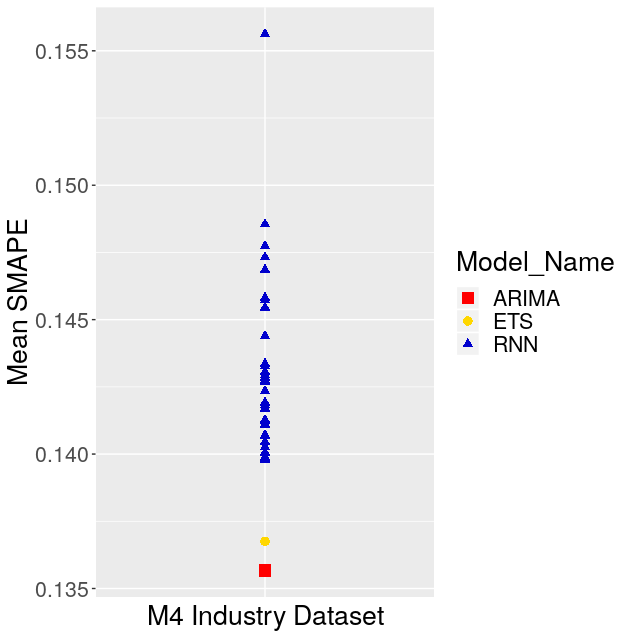}
	\includegraphics[width=0.5\textwidth]{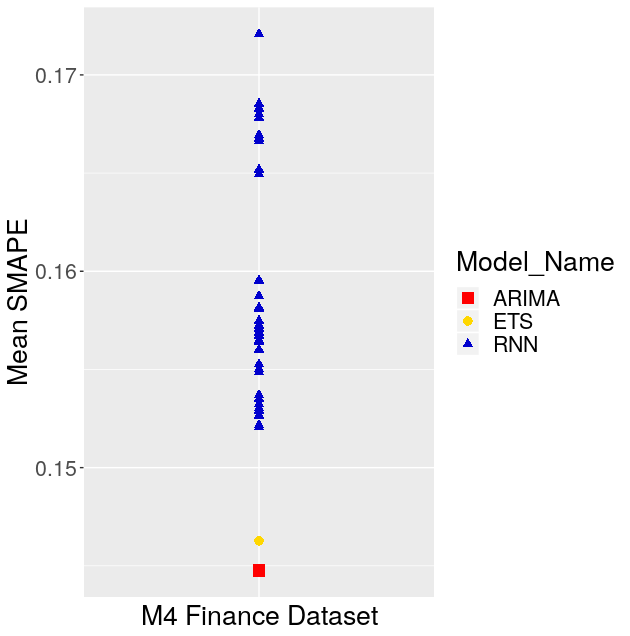}
	\includegraphics[width=0.5\textwidth]{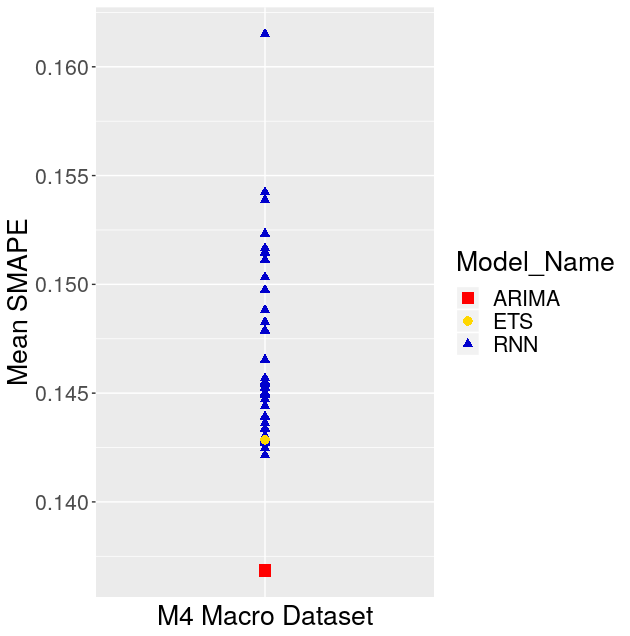}
	\includegraphics[width=0.5\textwidth]{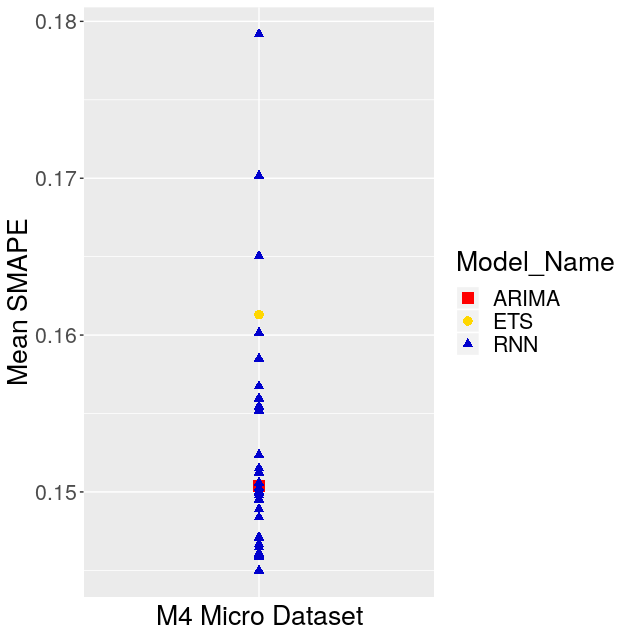}
	\includegraphics[width=0.5\textwidth]{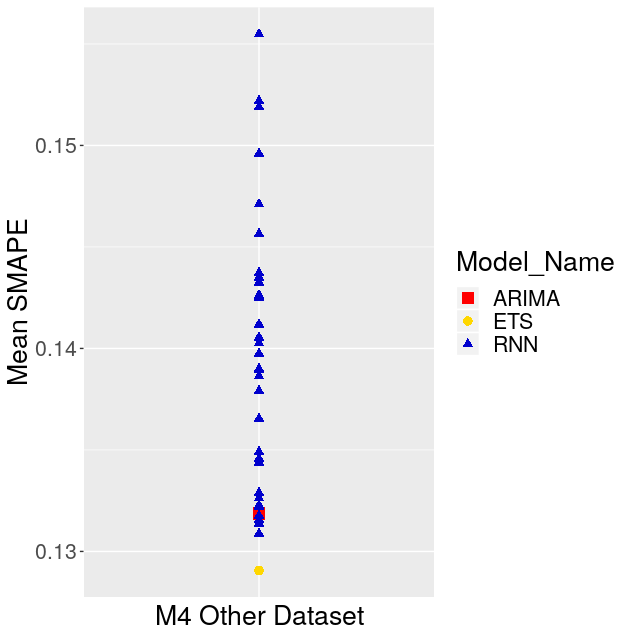}
	\caption{Performance of RNNs Compared to Traditional Univariate Techniques in Different M4 Categories - Mean SMAPE}
	\label{fig:rnn_performance_m4_categories_mean_smape}
\end{figure*}

The performance of the RNN architectures compared to traditional univariate benchmarks depend on the performance metric used.
An RNN architecture is able to outperform the benchmark techniques on all the datasets except the M4 monthly dataset, in terms of both the SMAPE and MASE error metrics. On the M4 monthly dataset, some RNNs outperform ETS, but ARIMA performs better than all of the RNNs. However, since we build the models per each category of the M4 monthly dataset, we further plot the performance in terms of the different categories in Figure \ref{fig:rnn_performance_m4_categories_mean_smape}. From these plots, we can see that RNNs outperform traditional univariate benchmarks only in the Micro category. Further plots in terms of the other error metrics are shown in the Online Appendix\footnote{\url{https://drive.google.com/file/d/16rdzLTFwkKs-_eG_MU0i_rDnBQBCjsGe/view?usp=sharing}}.

We also observe that mean error metrics are dominated by some outlier errors for certain time series. This is due to the observation that on the Tourism dataset, RNNs outperform ETS and ARIMA with respect to the median SMAPE but not the mean SMAPE. Furthermore, the 'Stacked\_LSTM\_COCOB' model combination in general performs competitively on most datasets. It outperforms the two univariate benchmarks on the CIF 2016, NN5 and the Wikipedia Web Traffic datasets. Therefore, we can state that Stacked model combined with the LSTM with peephole cells and the COCOB optimizer is a competitive model combination to perform forecasting.

\subsection{Experiments Involving the Total Number of Trainable Parameters}

The comparison of the relative performance of the recurrent units presented in Section \ref{sec:recurrent_unit_performance}, is by considering the cell dimension as a tunable hyperparameter for the RNNs. However, as mentioned before in Section \ref{sec:hyperparam}, in other research communities it is also common to tune the total number of trainable parameters as a hyperparameter instead of the cell dimension. Therefore, we also perform experiments by tuning the total number of trainable parameters as a hyperparameter to compare the relative performance of the different recurrent unit types. For this experiment we select the best configurations identified via the results analyzed thus far. Consequently, we choose the Stacked architecture along with the COCOB optimizer to run on all the three recurrent unit types. We perform this experiment on the four datasets CIF, Wikipedia Web Traffic, M3 and NN5 where the RNNs have outperformed the statistical benchmarks. Particularly on the NN5 and Wikipedia Web Traffic datasets, we run the version of the Stacked architecture without applying the STL decomposition with the increased input window size. Table \ref{tab:experiment_results_with_number_of_trainable_parameters} shows these results in terms of the mean SMAPE values. 

\begin{table*}
	\begin{center}
			\begin{tabular}{lcccc}
				\toprule
				Model Name & CIF & Kaggle & M3 & NN5\\
				\hline
				Stacked GRU cocob & 10.69 & \textbf{45.77} & 14.67 & \textbf{22.46}\\
				Stacked LSTM cocob & \textbf{10.54} & 45.93 & \textbf{14.44} & 24.04\\
				Stacked ERNN cocob & 10.66 & 129.23 & 14.99 & 24.67\\
				\hline
			\end{tabular}
		\caption{Mean SMAPE Results with the Number of Trainable Parameters as a Hyperparameter}
		\label{tab:experiment_results_with_number_of_trainable_parameters}
	\end{center}
\end{table*}

\begin{figure*}[htbp!]
	\captionsetup{justification=centering}
	\hspace{-1cm}
	\includegraphics[width=0.55\textwidth]{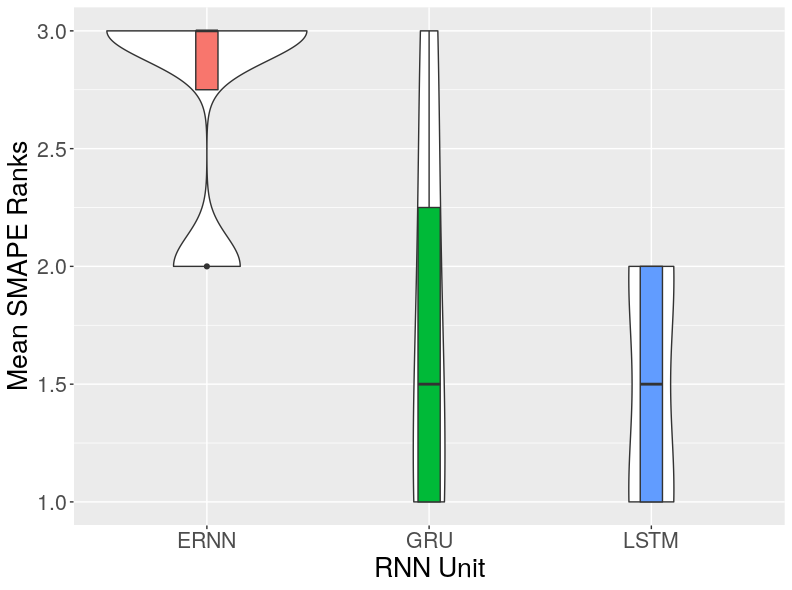}
	\includegraphics[width=0.55\textwidth]{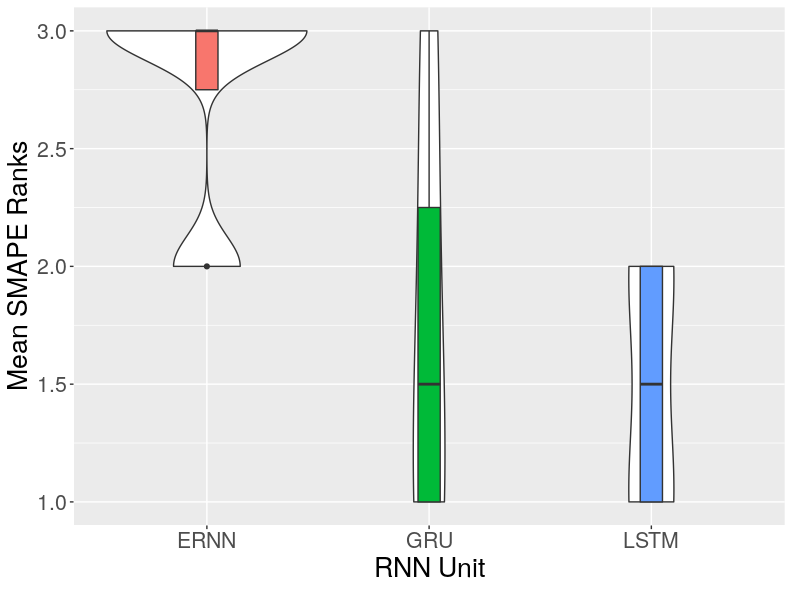}\\
	\caption{Relative Performance of Different Recurrrent Cell Types under the Same Number of Total Trainable Parameters}
	\label{fig:rnn_unit_performance_trainable_weights}
\end{figure*}

The violin plots in Figure \ref{fig:rnn_unit_performance_trainable_weights} indicate the relative performance of the three RNN units in terms of both the mean SMAPE and mean MASE ranks. In accordance with the results obtained in Section \ref{sec:recurrent_unit_performance}, Figure \ref{fig:rnn_unit_performance_trainable_weights} also indicates that the LSTM cell with peephole connections performs the best, the ERNN cell performs the worst and the performance of the GRU is between the other two. Again, the Friedman test of statistical significance performed in terms of the mean SMAPE values produces an overall $p$-value of 0.174 which means that this difference is not statistically significant.  

Overall, we see that the results obtained after tuning for the total number of trainable parameters are coherent with the results obtained by tuning for the cell dimension instead. 
As mentioned in Section \ref{sec:hyperparam}, there is a direct connection between the amount of trainable parameters and the cell dimension. 
Following \citet{smyl2020esrnn}, the cell dimension is not a very sensitive hyperparameter, which means that changes in the cell dimension, and consequently changes in the number of trainable parameters, do not have significant effects on the RNN performance. Therefore, we assume that the conclusions derived by tuning the cell dimension as a hyperparameter hold valid even after correcting for the total number of trainable parameters in the different recurrent units. Nevertheless, since tuning for the cell dimension is the viable approach due to practical limitations, we conclude that it is sufficient to tune the cell dimension across a similar range to produce a suitable number of trainable parameters in the different recurrent unit types.

\subsection{Comparison of the Computational Costs of the RNN Models Vs. the Benchmarks}

In addition to the comparisons of the prediction performance presented thus far, we perform a comparison between the computational times of the RNN models with respect to the chosen standard benchmarks, on the four datasets CIF2016, Wikipedia Web Traffic, M3 and NN5 where an RNN model outperforms the benchmarks. 

The computational times presented in Table \ref{tab:computational_costs} are for the best performing RNN models in each one of those datasets. For the purpose of comparison, both the benchmarks and the RNN models are allocated 25 CPU cores for the execution. The overall process of the RNN models has different stages in its pipeline such as preprocessing of the data, tuning the hyperparameters and the final model training with the optimal configuration for testing. Therefore, Table \ref{tab:computational_costs} shows a breakdown of the computational costs for these individual stages as well as the total time. 

From Table \ref{tab:computational_costs} we see that, compared to the standard benchmarks, the RNN models have taken a considerable amount of computational time for the overall process. For instance, on the horizon 12 category of the CIF dataset, the total time for the RNN is $3749.5\:s\;(1.0\:hr)$ whereas \texttt{auto.arima} and \texttt{ets} have taken only $85.1\:s\;(1.4\:min)$ and $41.6\:s$ respectively. Similarly on the NN5 dataset, the RNN model has taken a total of $51490.7\:s\;(14.3\:hrs)$ whereas \texttt{auto.arima} and \texttt{ets} models have taken $1067.7\:s\;(17.8\:min)$ and $53.3\:s$ only. However, it is also evident in Table \ref{tab:computational_costs}, that most of the computational time in RNNs is devoted to the hyperparameter tuning using SMAC with 50 iterations. For example, in the Micro category of the M3 dataset, although the whole pipeline takes $2523.0\:s\;(42.1\:min)$, the training of the final model with the optimal configuration and testing has accounted for only $565.6\:s\;(9.4\:min)$. On the other hand, the total running time for \texttt{auto.arima} and \texttt{ets} on the same dataset is $849.8\:s\;(14.2\:min)$ and $73.0\:s\;(1.2\:min)$ respectively. Hence, even though the overall process of RNNs is computationally costly, the training and testing of one model can be comparable to the computational costs of the statistical benchmarks. Moreover, compared to the benchmarks which build one model per every series, the final trained models from RNNs are less complex with fewer parameters than the univariate techniques, on a global scale.  

From this comparison we see that RNNs are typically computationally more expensive models compared to traditional univariate techniques. However, regardless of the computational cost they are capable of performing better than the traditional univariate benchmarks in all the cases mentioned in Table \ref{tab:computational_costs}. This finding is another aspect of the recent changes in the forecasting community now acknowledging that complex methods can have merits over simpler statistical benchmarks~\citep{Makridakis2018-nt}. Yet, our study shows how RNNs can be trained in a way to achieve such improvements. With the availability of massive amounts of computational resources nowadays, the improved accuracy brought forward by the RNNs for forecasting is certainly beneficial for forecasting practitioners.

\begin{table*}
	\begin{center}
		\resizebox{\textwidth}{!}{
			\begin{tabular}{lcSSSS}
				\toprule
				\multicolumn{1}{c}{Dataset} & \multicolumn{1}{c}{Model} & \multicolumn{1}{c}{Preprocessing} & \multicolumn{1}{c}{Hyperparameter Tuning} & \multicolumn{1}{c}{Model Training \& Testing} & \multicolumn{1}{c}{Total}\\
				\hline
				CIF(12) & S2SD LSTM NMW cocob & 1.9 & 2531.8 & 1215.7 & 3749.5\\
				CIF(12) & \texttt{auto.arima} & $\mhyphen$ &$\mhyphen$  & $\mhyphen$ & 85.1\\
				CIF(12) & \texttt{ets} & $\mhyphen$ &$\mhyphen$  &$\mhyphen$  & 41.6\\
				\hline
				CIF(6) & S2SD LSTM NMW cocob & 0.5 &  2535.3 & 304.9 & 2840.3\\
				CIF(6) & \texttt{auto.arima} & $\mhyphen$ & $\mhyphen$ & $\mhyphen$ & 4.3\\
				CIF(6) & \texttt{ets} & $\mhyphen$ & $\mhyphen$ & $\mhyphen$ & 8.3\\
				\hline
				Kaggle & NSTL Stacked GRU adam & 159.7 & 12803.2 & 5350.2 & 18313.0\\
				Kaggle & \texttt{auto.arima} & $\mhyphen$ & $\mhyphen$ & $\mhyphen$ & 834.4\\
				Kaggle & \texttt{ets} & $\mhyphen$ & $\mhyphen$ & $\mhyphen$ & 481.7\\
				\hline
				M3(Mic) & S2SD GRU MW adagrad & 30.2 & 1927.2 & 565.6 & 2523.0\\
				M3(Mic) & \texttt{auto.arima} &$\mhyphen$  & $\mhyphen$ & $\mhyphen$ & 849.8\\
				M3(Mic) & \texttt{ets} &$\mhyphen$  &$\mhyphen$  &$\mhyphen$  & 73.0\\
				\hline
				M3(Mac) & S2SD GRU MW adagrad & 28.2 & 5359.9 &  1863.0 & 7251.0\\
				M3(Mac) & \texttt{auto.arima} & $\mhyphen$ & $\mhyphen$ & $\mhyphen$ & 719.5\\
				M3(Mac) & \texttt{ets} & $\mhyphen$ & $\mhyphen$ & $\mhyphen$ & 72.7\\
				\hline
				M3(Ind) & S2SD GRU MW adagrad & 30.5 & 4084.0 & 1054.6 & 5169.1\\
				M3(Ind) & \texttt{auto.arima} &$\mhyphen$  & $\mhyphen$ & $\mhyphen$ & 1462.3\\
				M3(Ind) & \texttt{ets} & $\mhyphen$ & $\mhyphen$ & $\mhyphen$ & 218.8\\
				\hline
				M3(Dem) & S2SD GRU MW adagrad & 5.8 & 1875.9 & 422.6 & 2304.4\\
				M3(Dem) & \texttt{auto.arima} & $\mhyphen$ & $\mhyphen$ & $\mhyphen$ & 273.3\\
				M3(Dem) & \texttt{ets} & $\mhyphen$ & $\mhyphen$ & $\mhyphen$ & 73.1\\
				\hline
				M3(Fin) & S2SD GRU MW adagrad & 12.1 & 1848.9 & 493.0 & 2354.0\\
				M3(Fin) & \texttt{auto.arima} & $\mhyphen$ & $\mhyphen$ & $\mhyphen$ & 352.2\\
				M3(Fin) & \texttt{ets} & $\mhyphen$ & $\mhyphen$ & $\mhyphen$ & 90.3\\
				\hline
				M3(Oth) & S2SD GRU MW adagrad & 4.1 & 1319.8 & 147.5 & 1471.5\\
				M3(Oth) & \texttt{auto.arima} & $\mhyphen$ & $\mhyphen$ & $\mhyphen$ & 273.3\\
				M3(Oth) & \texttt{ets} & $\mhyphen$ & $\mhyphen$ & $\mhyphen$ & 31.3\\
				\hline
				NN5 & Stacked LSTM adam & 42.2 &  37660.2 &  13788.3 & 51490.7\\
			    NN5 & \texttt{auto.arima} & $\mhyphen$ & $\mhyphen$ & $\mhyphen$ & 1067.7\\
   				NN5 & \texttt{ets} & $\mhyphen$ & $\mhyphen$ & $\mhyphen$ & 53.3\\
				\hline
			\end{tabular}}
		\caption{Computational Times Comparison of the RNNs with the Benchmarks (in seconds)}
		\label{tab:computational_costs}
	\end{center}
\end{table*}
 
\subsection{Hyperparameter Configurations}

Usually, the larger the initial hyperparameter space that needs to be searched, the higher the number of iterations of the automated hyperparameter tuning technique should be. This depends on the number of hyperparameters as well as the initial hyperparameter ranges. We use 50 iterations of the SMAC algorithm for hyperparameter tuning to be suitable across all the datasets. For our experiments we choose roughly the same range across all the datasets as shown in Table \ref{tab:hyperparameter_ranges}, except for the minibatch size. The minibatch size needs to be chosen proportional to the size of the dataset. Usually, for the lower bound of the initial hyperparameter range of the minibatch size, around $1/10$th of the size of the dataset is appropriate. However, we vary the upper bound in different orders for the different datasets. For small datasets such as CIF, NN5, Tourism and the different categories of M3, the upper bound differs from the lower bound in the orders of 10s. For bigger datasets such as the Wikipedia Web Traffic, and the different categories of M4, we set the upper bound to be larger than the lower bound in the orders of 100s. For the number of hidden layers in the RNN, many recent studies suggest that a low value usually performs better~\citep{Smyl2016-rf, Flunkert2017-wp, deep_factors_wang, Bandara2017-gb}. Following this convention, we choose the number of layers between 1-2 and we observe that the RNNs perform well with such low values. On the one hand this is due to the overfitting effects resulting from the increased number of parameters with the added layers. On the other hand, the number of layers also directly relates to the computational complexity. As for the learning rates, we see that the convergence of the Adagrad optimizer usually requires higher learning rates in the range 0.01 - 0.9. For the Adam optimizer the identified range is smaller in between 0.001 - 0.1. On the other hand, variations in the cell dimension and the standard deviation of the random normal initializer barely impact the performance of the models. It is also important not to set large values for the standard deviation of the Gaussian noise and the L2 weight regularization parameters, since too high a value for them makes the model almost completely underfit the data and eliminate the NN’s effect entirely in producing the final forecasts.

%% file: sections/conclusion.tex
\section{Conclusions}
\label{sec:conclusion}

The motivation of this study is to address some of the key issues in using RNNs for forecasting and evaluate if and how they can be used by forecasting practitioners with limited knowledge of these techniques for their forecasting tasks effectively. Through a number of systematic experiments across datasets with diverse characteristics, we derive conclusions from general data preprocessing best practices to hyperparameter configurations, best RNN architectures, recurrent units and optimizers. All our models exploit cross-series information in the form of global models and thus leverage the existence of massive time series databases with many related time series.

From our experiments we conclude that the Stacked architecture combined with the LSTM cells with peephole connections and the COCOB optimizer, fed with deseasonalized data in a moving window format can be a competitive model generally across many datasets. 
In particular, though not statistically significant, the LSTM is the best unit type even when correcting for the amount of trainable parameters.
We further conclude that when all the series in the dataset follow homogeneous seasonal patterns with all of them covering the same duration in time with sufficient lengths, RNNs are capable of capturing the seasonality without prior deseasonalization. 
Otherwise, RNNs are weak in modelling seasonality on their own, and a deseasonalization step should be employed. 
From the experiments involving the pooled and unpooled versions of the regression models, we can conclude that the concept of cross-series information helps in certain types of datasets. However, even on those datasets which involve many heterogeneous series, the strong modelling capabilities of RNNs can drive them to perform competitively in terms of the forecasting accuracy. Therefore, we can conclude that leveraging cross-series information has its own benefits on sets of time series while the predictive capability provided by the RNNs can further improve the forecasting accuracy. 
With respect to the computational costs, we observe that RNNs take relatively higher computational times compared to the statistical benchmarks. However, with the cloud infrastructure nowadays commonly in place at companies such processing times are feasible. Moreover, our study has empirically proven that RNNs are good candidates for forecasting which in many cases outperform the statistical benchmarks that are currently the state-of-the-art in the community. Thus, with the extensive experiments involved with this study, we can confirm that complex methods now have benefits over simpler statistical benchmarks in many forecasting situations.  

Our procedure is (semi-)automatic as for the initial hyperparameter ranges of the SMAC algorithm, we select approximately the same range across all the datasets except for the minibatch size which we select depending on the size of each dataset. Thus, though fitting of RNNs is still not as straightforward and automatic as for the two state-of-the-art univariate forecasting benchmarks, \texttt{ets} and \texttt{auto.arima}, our paper and our released code framework are important steps in this direction.
We finally conclude that RNNs are now a good option for forecasting practitioners to obtain reliable forecasts which can outperform the benchmarks.

%% file: sections/future_directions.tex
\section{Future Directions}
\label{sec:future_directions}

The results of our study are limited to point forecasts in a univariate context. Nevertheless, modelling uncertainty of the NN predictions through probabilistic forecasting has received growing attention recently in the community. Also, when considering a complex forecasting scenario such as in retail industry, the sales of different products may be interdependent. Therefore, a sales forecasting task in such a context requires multivariate forecasting as opposed to univariate forecasting. Furthermore, although this study considers only single seasonality forecasting, in terms of higher frequency data with sufficient length, it becomes beneficial to model multiple seasonalities in a big data context. 

As seen in our study, global NN models often suffer from outlier errors for certain time series. This is probably due to the fact that the average weights of NNs found by fitting global models may not be suitable for the individual requirements of certain time series. Consequently, it becomes necessary to develop novel models which incorporate both global parameters as well as local parameters for individual time series, in the form of hierarchical models, potentially combined with ensembling where the single models are trained in different ways with the existing dataset (e.g., on different subsets). The work by \citet{Bandara2017-gb}, \citet{smyl2020esrnn}, and \citet{think_globally} are steps in this direction, but still this topic remains largely an open research question. 

In general, deep learning is a fast-paced research field, where many new architectures are introduced and discussed rapidly. However, for practitioners it often remains unclear in which situations the techniques are most useful and how difficult it is to adapt them to a given application case.
Recently, CNNs, although initially intended for image processing, have become increasingly popular for time series forecasting. The work carried out by \citet{Shih2018-tb} and \citet{Lai2018-yo} follow the argument that typical RNN-based attention schemes are not good at modelling seasonality. Therefore, they use a combination of CNN filters to capture local dependencies and a custom attention score function to model the long-term dependencies. \citet{Lai2018-yo} have also experimented with recurrent skip connections to capture seasonality patterns. On the other hand, Dilated Causal Convolutions are specifically designed to capture long-range dependencies effectively along the temporal dimension~\citep{wavenet}. Such layers stacked on top of each other can build massive hierarchical attention networks, attending points even way back in the history. They have been recently used along with CNNs for time series forecasting problems. More advanced CNNs have also been introduced such as Temporal Convolution Networks (TCN) which combine both dilated convolutions and residual skip connections. TCNs have also been used for forecasting in recent literature~\citep{Borovykh2017-ul}. Recent studies suggest that TCNs are promising NN architectures for sequence modelling tasks on top of being efficient in training~\citep{Shaojie2018}. Therefore, a competitive advantage may begin to unfold for forecasting practitioners by using CNNs instead of RNNs.